\pdfoutput=1

\documentclass[11pt]{article}

\usepackage{ACL2023}

\usepackage{times}
\usepackage{latexsym}
\usepackage{booktabs}
\usepackage{graphicx}
\usepackage{multirow}
\usepackage{array} 
\usepackage{xcolor}
\usepackage{float}
\usepackage{natbib}
\usepackage{url}
\usepackage{amsmath, amsfonts, amssymb}
\usepackage{enumitem}
\usepackage{booktabs}
\usepackage{listings}
\usepackage[most]{tcolorbox}
\usepackage{subcaption}
\usepackage{enumitem}
\usepackage{dblfloatfix} 
\usepackage{dsfont}
\usepackage{pifont}      
\newcommand{\cmark}{\ding{51}}   
\newcommand{\xmark}{\ding{55}}   

\usepackage[T1]{fontenc}

\usepackage[utf8]{inputenc}

\usepackage{microtype}

\usepackage{inconsolata}
\usepackage{amsthm}
\newtheorem*{theorem*}{Theorem}
\newtheorem{theorem}{Theorem}
\theoremstyle{remark}

\DeclareMathOperator{\Var}{Var}

\newcommand{\graytext}[1]{\textcolor{gray}{#1}}

\usepackage[commandnameprefix=ifneeded]{changes}
\definechangesauthor[name={Huanxin}, color=blue]{HS}
\definechangesauthor[name={Jian}, color=red]{JK}

\title{Analyzing Uncertainty of LLM-as-a-Judge: Interval Evaluations with Conformal Prediction}

\author{Huanxin Sheng$^1$, Xinyi Liu$^1$, Hangfeng He$^1$, Jieyu Zhao$^2$, Jian Kang$^{1,3}$ \\
$^1$ University of Rochester, $^2$ University of Southern California, $^3$ MBZUAI \\
\texttt{\{hsheng2@ur., hangfeng.he@, xinyu.liu1@simon.\}rochester.edu} \\ \texttt{jieyuz@usc.edu, jian.kang@mbzuai.ac.ae}}

\begin{document}
\maketitle

\begin{abstract}
    LLM-as-a-judge has become a promising paradigm for using large language models (LLMs) to evaluate natural language generation (NLG), but the uncertainty of its evaluation remains underexplored. This lack of reliability may limit its deployment in many applications. This work presents the first framework to analyze the uncertainty by offering a prediction interval of LLM-based scoring via conformal prediction. Conformal prediction constructs continuous prediction intervals from a single evaluation run, and we design an ordinal boundary adjustment for discrete rating tasks. We also suggest a midpoint-based score within the interval as a low-bias alternative to raw model score and weighted average. We perform extensive experiments and analysis, which show that conformal prediction can provide valid prediction interval with coverage guarantees. We also explore the usefulness of interval midpoint and judge reprompting for better judgment.
    \footnote{Our code and data are available at \\ \url{https://github.com/BruceSheng1202/Analyzing_Uncertainty_of_LLM-as-a-Judge}    } 
\end{abstract}

\section{Introduction}
Large language models (LLMs) have become powerful automatic evaluators for natural language generation (NLG) tasks, 
known as LLM-as-a-judge. Its consistency with human judgments 
results in strong performance with respect to metrics like ROUGE~\citep{Lin2004ROUGEAP}, BLEU~\citep{Papineni2002BleuAM}, and BERTScore~\citep{zhang2020bertscoreevaluatingtextgeneration}. Besides, LLM judges can flexibly adapt to diverse evaluation criteria and provide scalable, cost-effective assessments compared to expert annotation~\citep{gao2024llmbasednlgevaluationcurrent, gu2025surveyllmasajudge}. These advantages make the LLM-as-a-judge 
useful in various scenarios, such as clinical radiology~\citep{Chaves2024ACA}, rumor detection~\citep{hong2025following}, cyberattack detection~\citep{yong2025attackseqbenchbenchmarkinglargelanguage} and wildlife trafficking identification~\citep{Barbosa2025ACL}. 

However, a single evaluation from a LLM judge might be biased~\citep{wu2023stylesubstanceevaluationbiases, li2024llmsasjudgescomprehensivesurveyllmbased} and uncertain due to inherent randomness~\citep{schroeder2025trustllmjudgmentsreliability}, thus undermining its reliability in scenarios like healthcare~\citep{chung2025verifactverifyingfactsllmgenerated} and finance~\citep{kamble2025expectunexpectedfailsafelong}. Though a LLM judge can express its confidence with well-designed prompt or via fine-tuning~\citep{xu2024sayselfteachingllmsexpress, liu-etal-2024-llms-learn-uncertainty, taubenfeld2025confidenceimprovesselfconsistencyllms}, it may still suffer from overconfidence~\citep{xiong2024llmsexpressuncertaintyempirical} or dishonesty~\citep{li2024surveyhonestylargelanguage}. We ask: \textit{How can a LLM judge provide reliable evaluation given the user request?}

Conformal prediction~\citep{vovk2005algorithmic} is a promising way to quantify the uncertainty of an LLM judge~\citep{ye2024benchmarking}. It outputs a prediction interval (or set for classification) to a model output with three key advantages. First, conformal prediction is a distribution-free uncertainty quantification method, which is suitable for black-box models like LLMs due to unknown input data distribution for most (if not all) LLMs. Second, it can provide post-hoc uncertainty quantification using only a calibration step based on LLM outputs. Third, the prediction interval given by conformal prediction enjoys statistically guaranteed coverage, i.e., how likely the ground truth falls within the interval, as long as the data is exchangeable.

In this paper, we comprehensively evaluate nine\footnote{Seven regression-type and two ordinal-type methods.} conformal prediction methods in quantifying the uncertainty of a LLM judge in rating-based evaluation tasks, each of which constructs a prediction interval for a rating output by the LLM judge. For each conformal prediction method, we evaluate its efficiency (i.e., average width of prediction intervals) and coverage (i.e., the probability that ground truths fall within prediction intervals). Furthermore, to adapt to the ordinal and discrete nature in organic rating-based evaluation, we propose boundary adjustment that adjusts the endpoints of prediction interval to be aligned with the rating scales. We prove that the boundary adjustment yields an interval suitable to the ordinal setting with provable non-decreasing coverage. From our comprehensive analysis, we demonstrate that the quality of prediction interval attributes to design choices of the LLM judge (e.g., which LLM to use as the judge, which prompting strategy for evaluation) as well as the size of calibration data during calibration. Finally, we show that the midpoint of the prediction interval provides better estimate to the ground truth to further assist better decision-making, while reprompting the LLM judge with prediction interval might not improve the judgement. Our analysis advocate for a shift from direct scoring to uncertainty-aware evaluations, offering references for more reliable evaluation and better decision-making.

In summary, our contributions are
\begin{itemize}[
    align=left,
    leftmargin=1em,
    labelsep=0pt,
    labelwidth=1em,
    nosep
]
\item To our knowledge, we are the first to analyze the uncertainty of LLM-as-a-judge using conformal prediction in rating-based evaluation.

\item We design boundary adjustment to improve the efficiency empirically without compromising the coverage. The interval midpoints suggest better alignment to human evaluation.

\item We analyze factors affecting the interval quality, including the LLM-as-a-judge framework itself, the choice of LLM in the framework, and the size of calibration in conformal prediction, and offer practical insights or recommended choices.
\end{itemize}

\section{Related Work}
\paragraph{Uncertainty Quantification for LLM-as-a-Judge.}
Uncertainty quantification for LLM-as-a-judge is an important yet less explored area. 
\citet{wagner2024blackboxuncertaintyquantificationmethod} prompt the judge to justify each rating option as if it were correct and then construct a confusion matrix from token-level probabilities of these assessments to derive confidence scores.
\citet{xie2025empiricalanalysisuncertaintylarge} use token probabilities to estimate the confidence of judgments, and demonstrate that such measures exhibit bias and instability through extensive experiments.
Similar conclusions are also found when applying other two common paradigms: (1) prompting LLMs to self-report confidence~\citep{yona2024largelanguagemodelsfaithfully, xu2024sayselfteachingllmsexpress}, which can suffer from overconfidence~\citep{xiong2024llmsexpressuncertaintyempirical} or dishonesty~\citep{li2024surveyhonestylargelanguage}, and (2) consistency-based approaches that rely on multiple generations~\citep{tian2023justaskcalibrationstrategies, xiong2024llmsexpressuncertaintyempirical}, which, like the confusion matrix-based method, are computationally expensive. 
To our best knowledge, \citet{jung2024trustescalatellmjudges} is the most relevant work to us, which applies conformalized risk control~\citep{angelopoulos2022learntestcalibratingpredictive} to ensure agreement with human preferences in pairwise response comparison~\citep{zhou2024fairerpreferenceselicitimproved, li2024calibraevalcalibratingpredictiondistribution, li2024dnaevalenhancinglargelanguage, zhang2025crowdcomparativereasoningunlocking, burg2025aligningblackboxlanguagemodels}. In contrast, we focus on using conformal prediction to quantify uncertainty in rating tasks instead of pairwise preference modeling.

\paragraph{Conformal Prediction for LLMs.}
Conformal prediction~\citep{vovk2005algorithmic} has drawn interest for uncertainty quantification in LLMs~\citep{ye2024benchmarking,campos2024conformal} due to its distribution-free and post-hoc nature with provable statistical guarantee. Owing to these advantages, recent works primarily apply conformal prediction to classification tasks, such as multiple-choice question answering~\citep{kumar2023conformalpredictionlargelanguage, zhang2024calibratingconfidencelargelanguage, su2024apienoughconformalprediction, vishwakarma2025prune} and response selection for factual consistency~\citep{quach2024conformallanguagemodeling, mohri2024languagemodelsconformalfactuality, wang2024conuconformaluncertaintylarge, kladny2025conformalgenerativemodelingimproved}. These studies typically focus on ensuring that the correct answer is included in a unordered prediction set. Different from these works, we focus on providing prediction intervals that reflect the variability in LLM judgments in rating tasks, which has ordinal preference. 

\section{Analyzing Uncertainty of LLM Judges}~\label{section: framework}
\begin{figure*}[!t]           
  \centering
  \includegraphics[width=.9\textwidth]{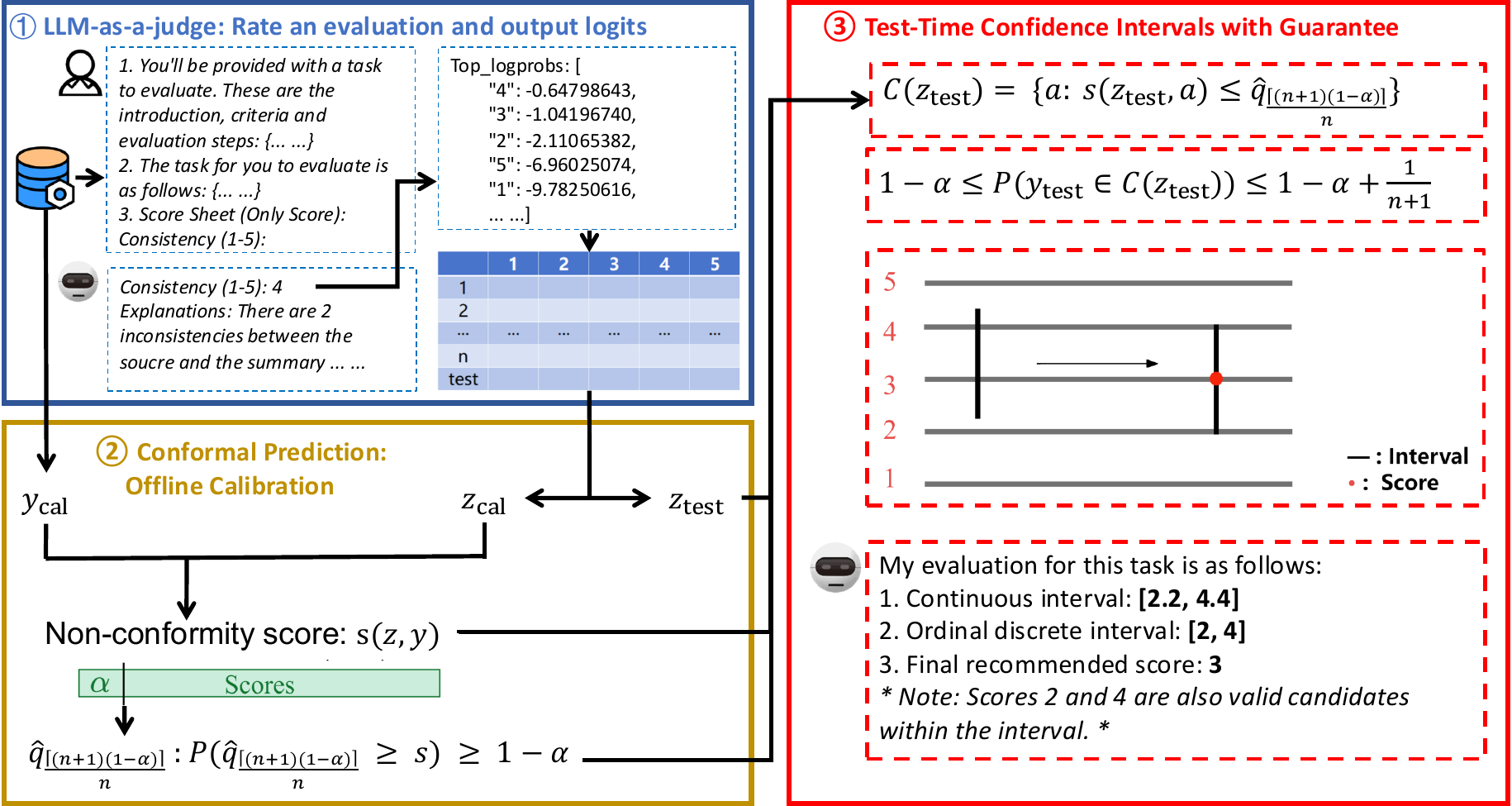}
  \caption{Overview of quantifying the uncertainty in rating-based evaluation. We apply conformal prediction to construct the prediction interval and set the width of the prediction interval as the uncertainty.}
  \label{fig:framework}
\end{figure*}

\subsection{Preliminaries}
\label{subsec: preliminaries}
\paragraph{LLM-as-a-Judge.} LLMs have been widely adopted as evaluators to score NLG tasks recently, known as LLM-as-a-judge. It commonly yields a predicted score $y_0$ on a Likert scale~\citep{nemoto2014likert}. Following G-Eval~\citep{liu2023g}, given a prompt $p$ and a generated text $x$ to be evaluated, an LLM judge $M$ is expected to produce a response 
\begin{equation}
    M(p, x) = (z, y_0),
\end{equation}
where $z$ is the token logits, and $y_0$ is a scalar score in a predefined scale, indicating the quality of generated text given by LLM judge. 
For rating-based evaluations, we extract the logits of certain tokens (e.g., $1$, $2$, $3$, $4$, $5$ if in a Likert scale) at the position of the rating token only. There are also other evaluation paradigms that a rating judge can help with~\citep{gu2025surveyllmasajudge}, such as pairwise comparison or ranking, in which candidate outputs are first scored by the LLM judge and then compared or ordered based on those scores~\citep{wang2025improvingllmasajudgeinferencejudgment, wei2025rocketevalefficientautomatedllm}.

\paragraph{Conformal Prediction.} 
Conformal prediction~\citep{vovk2005algorithmic} is a model-agnostic uncertainty quantification method. It constructs a prediction interval (or a prediction set for classification) with coverage guarantee, i.e., how likely the ground truth will fall within the prediction interval/set. Two notable advantages of conformal prediction are its post-hoc nature, i.e., free of training or prompting the judge model, and distribution-free nature, i.e., without requiring knowledge about underlying data distribution. In our work, we adopt split conformal prediction~\citep{vovk2005algorithmic}, which quantifies the uncertainty with a held-out calibration set. A non-conformity score function $s(z, y)$ is computed for each point in the calibration set, to measure how ``unusual'' a prediction $\hat{y}$ is to a ground truth $y$. For regression tasks, the non-conformity score is often defined as 
\begin{equation}
\label{equation:absolute_difference}
    s(z,y) = |\hat{y} - y|.
\end{equation}
where $\hat{y} = f(z)$ is a score predicted from the input $z$ using a function $f$. Then, given a user-desired miscoverage rate $\alpha$, the $\frac{\lceil \left(n+1\right)\left(1 - \alpha\right)\rceil}{n}$-quantile $\hat{q}$ of these scores can be used to construct 
the prediction interval $\mathcal{C}\left(z_{\rm test}, \hat{y}_{\rm test}\right)$ for the prediction $\hat{y}_{\rm test}$ of any test point $z_{\rm test}$ as
\begin{equation}\label{equation:naive_cp}
    \mathcal{C}\left(z_{\rm test}, \hat{y}_{\rm test}\right) = [\hat{y}_{\rm test} -\hat{q},\ \hat{y}_{\rm test} +\hat{q}],
\end{equation}
or equivalently
\begin{equation}
\label{equation:set_form}
    \mathcal{C}\left(z_{\rm test}, \hat{y}_{\rm test}\right) = \left\{a: s\left(z_{\rm test}, a\right) \leq \hat{q}\right\}.
\end{equation}
Statistically, it is proven that such a prediction interval enjoys the following coverage guarantee~\citep{angelopoulos2021gentle}
\begin{equation}
\label{equation:guarantee}
    1 - \alpha \leq \mathbb{P}\left(y_{\rm test} \in \mathcal{C}\left(z_{\rm test}, \hat{y}_{\rm test}\right) \right) \leq 1 - \alpha + \frac{1}{n+1},
\end{equation}
as long as the calibration set and test set are exchangeable, i.e., the joint distribution remain the same after any permutations on these two sets.

\subsection{From Logits to Intervals}
In this paper, we focus on quantifying the uncertainty of LLM-as-a-judge using conformal prediction in discrete rating-based evaluations (e.g., in Likert scale). An overview of the workflow is presented in Figure~\ref{fig:framework}.

\paragraph{Extract Logits as Feature.}
Our framework mainly targets for uncertainty estimation in a discrete, Likert-scale (1 -- 5) rating-based evaluation. An example of such rating-based evaluation is given in Figure~\ref{fig:framework}. Specifically, an LLM judge, prompted with a chain-of-thought (CoT) instruction that specifies an output format, generates a response containing its rating. Prompts that we used are listed in Appendix~\ref{appendix:prompt}) 

To obtain the $\frac{\lceil \left(n+1\right)\left(1 - \alpha\right)\rceil}{n}$-quantile of the non-conformity scores of the ratings, we extract the token logits corresponding to Likert-scale scores as features for quantile estimation. However, LLMs may not always respond in a fixed format, making extraction of the rating token logit challenging. Thus, we resort to a rule-based strategy to locate the rating token at the most frequent position of the rating. After locating the rating token (e.g., $4$ in Figure~\ref{fig:framework}), we extract the log probabilities of all potential rating tokens (e.g., 1 -- 5). Additionally, to ensure semantic consistency, we aggregate the probabilities of tokens with equivalent meanings (e.g., ``two'' vs. 2). As a result, we obtain a $K$-dimensional feature vector $z$ representing the logits associated with each candidate rating token in $\left\{1, 2, \dots, K\right\}$\footnote{We use $K = 5$ for the standard Likert scale or GPA-like settings, but $K$ can be adapted to other granularities (e.g., $10$) depending on the scale.} as the input for conformal prediction $\left(z, y\right)$. Here we assume the logits for each prompt-text sample to be exchangeable, following prior works~\citep{kumar2023conformalpredictionlargelanguage, su2024apienoughconformalprediction, quach2024conformallanguagemodeling}.

\paragraph{Prediction Interval Estimation.} To date, there has been a variety of conformal prediction methods~\citep{angelopoulos2021gentle, Fontana_2023, campos2024conformalpredictionnaturallanguage}. In our analysis, we choose a diverse set of nine conformal prediction methods that construct prediction interval rather than prediction set. Moreover, we observe that token logits exhibit strong heteroscedasticity (see more analysis in Appendix~\ref{appendix:Heteroskedasticity_testing}). Thus, we include conformalized quantile regression (CQR)~\citep{romano2019conformalized} and its variants, including asymmetric CQR, CHR~\citep{sesia2021conformalpredictionusingconditional} that uses histogram-based quantile estimator, LVD~\citep{NEURIPS2021_46c7cb50} that uses kernel regression based quantile estimator, two boosted conformal prediction methods~\citep{xie2024boostedconformalpredictionintervals} (i.e., Boosted CQR and Boosted LCP) and R2CCP~\citep{guha2024conformal}. Other than the regression-based approaches, rating-based evaluation can also be viewed as a ordinal classification task due to the ordinal nature of ratings. Thus, we also consider two conformal prediction methods designed for ordinal classification, namely Ordinal APS~\citep{lu2022improvingtrustworthinessaidisease} and Ordinal Risk Control~\citep{xu2024conformalriskcontrolordinal}. More details about these methods, including non-conformity score computation, interval construction and implementation details, are discussed in Appendix~\ref{appendix:CP Methods}. Note that, for ordinal classification-based conformal prediction methods, we view the token probabilities (i.e., token logits followed by a softmax operation) as a proxy of classification probabilities that classify the prompt and texts to be evaluated into the corresponding rating label. Consequently, we use token probabilities as input. However, token probabilities are well known to have multicollinearity, which makes them unsuitable for regression-based approaches. Thus, we only use token probabilities as the input for ordinal classification-based approaches, while use token logits directly for regression-based approaches.

\subsection{Boundary Adjustment}~\label{sec:boundaryadjustment}
Due to the ordinal and discrete nature of ratings, a continuous interval whose upper and lower bounds are continuous numeric values might not have exact meaning. Thus, we further apply a boundary adjustment strategy to transform the continuous interval to an ordinal interval whose upper and lower bounds align with the potential rating labels. 

Specifically, boundary adjustment is essentially redefining the non-conformity score function as
\begin{equation}
\label{equation:new_score_func}
s'(z,y) = s(z, y') =
\begin{cases}
s(z,\lceil y \rceil) & \text{if} \ y \leq \lfloor \hat{y}\rfloor, \\
s(z, y ) & \text{otherwise,} \\
s(z,\lfloor y \rfloor) & \text{if} \ y \geq \lceil \hat{y} \rceil. \\
\end{cases}
\end{equation}
Because all potential labels $y'$ are integers in rating evaluation, this new function ensures that the scores are consistent on calibration set. However, when constructing the prediction interval , it will transform the interval $\mathcal{C}(z_{\rm test})$ in Equation~\eqref{equation:set_form} to
\begin{equation}
\label{equation:boundary_adjustment}
     \{a: s'(z_{\rm test}, a) \leq \hat{q} \} = \left[l, u\right] \rightarrow \left[l', u'\right], 
\end{equation}
where $l' = \lceil l \rceil$ and $u' = \lfloor u \rfloor$. 

We shrink the boundaries to integer labels closest to the original continuous-valued boundaries by cutting excessive areas, because the excessive areas cover no potential labels. Thus, this shrinking adjustment will have no influence to its coverage. On the other hand, we can also expand an interval to mitigate the marginal miscoverage of the ground-truth labels or limited calibration size. For example, assuming the interval $[2.2, 3.9]$ only covers one possible rating $3$ but can be expanded to $[2, 4]$, a miscoverage can be avoided if the ground truth is either $2$ or $4$.

Theorem~\ref{theorem:adjustment_guarantee} shows the non-decreasing coverage after boundary adjustment. We defer its proof to Appendix~\ref{appendix:proofof ba coverage}. 

\begin{theorem}[Non-decreasing Coverage After Boundary Adjustment]
\label{theorem:adjustment_guarantee}
Based on coverage guarantee in Equation~\eqref{equation:guarantee}, we transform the non-conformity score function $s(z, y)$ by Equation (\ref{equation:new_score_func}) and adjust an continuous interval by Equation~\eqref{equation:boundary_adjustment}. Then, if the adjustment is performed by 
shrinking (i.e., $l' = \lceil l\rceil$, $u'=\lfloor u\rfloor$, and $\mathcal{C}'\left(z_{\rm test}\right) = [l', u']$), the coverage preserves:
\begin{equation}
    \mathbb{P}\left(Y_{\rm test} \in \mathcal{C}'\left(z_{\rm test}\right)\right) \ge 1-\alpha.
\end{equation}
If at least one boundary is expanded (i.e., $l' = \lfloor l\rfloor$ or $u' = \lceil u\rceil$), the coverage increases: 
\begin{equation}
    \mathbb{P}\left(Y_{\rm test} \in \mathcal{C}'(z_{\rm test})\right) > 1 - \alpha.
\end{equation}
\end{theorem}

Though we assume all ratings are integers in Equation~\eqref{equation:new_score_func}, ratings with discrete granularities can all be applicable. For example, GPA-scale scores can be mapped to a 1–13 Likert scale by linear transformation (e.g. $1.33 \times 3 - 2$). Then a boundary adjustment from $[4.6, 4.9]$ to $[4.67, 5]$ is equivalent to rounding $[11.8, 12.7]$ to $[12, 13]$.

\subsection{Midpoints as Calibrated Scores}
To make better use of the prediction interval for decision making, we seek to obtain a more accurate estimate of ground truth from the interval. One simple and intuitive way is to use the midpoint of the prediction interval as an alternative estimate. Due to the unknown distribution of the ground truth rating in a prediction interval, selecting the midpoint has several benefits. If the distribution can be naively assumed to be uniform or at least symmetric, the midpoint would be the best linear unbiased estimator of the true label. Even if biased, the midpoint is still the minimum-variance estimator of the ground truth rating given the interval. More intuitively, midpoint is a score evaluated in a shorter interval (i.e., the prediction interval) rather than the entire range of rating, thus should be more accurate than the primitive score given by the LLM judge. 

\section{Experiments}
\subsection{Experimental Setting}



\paragraph{Datasets.} 
We run experiments on evaluation benchmarks in text summarization, dialogue summarization and reasoning. For summarization, we use SummEval~\citep{fabbri2021summevalreevaluatingsummarizationevaluation} (1,600 samples) and DialSumm~\citep{gao-wan-2022-dialsummeval} (1,400 samples), each annotated by three human raters using Likert-scale scores across four dimensions (i.e., consistency, coherence, fluency, relevance). The average of the three ratings is used as the ground-truth label on GPA scale. For reasoning, we use the annotations of overall quality on Likert scale for CosmosQA~\citep{li2023teachingpretrainedmodelscommonsense}, DROP~\citep{dua2019dropreadingcomprehensionbenchmark}, e-SNLI~\citep{NEURIPS2018_4c7a167b} and GSM8K~\citep{cobbe2021trainingverifierssolvemath} in ROSCOE~\citep{golovneva2023roscoesuitemetricsscoring}, each with around 200 samples on Likert scale. 

\paragraph{LLM-based Evaluation.}  
We primarily adopt G-Eval~\citep{liu2023g} as our judge framework with a chain-of-thought (CoT) prompt. For reasoning tasks, we additionally use SocREval~\citep{he2024socrevallargelanguagemodels}. We provide the prompts used for experiments in Appendix~\ref{appendix:prompt}. Evaluations are mainly conducted using GPT-4o mini (2024-07-18), DeepSeek-R1-Distill-Qwen-32B~\citep{deepseekai2025deepseekr1incentivizingreasoningcapability} and Qwen2.5-72B-Instruct~\citep{qwen2025qwen25technicalreport} as LLM judges, all of which provide token-level logits. Though we choose these LLMs mainly due to budget constraints and limitations on computing resources, all selected LLMs are widely used and show strong performance in both summarization and reasoning tasks. In Appendix~\ref{sec:prompt_sense}, we provide more detailed justifications on the judge model selection and the comparison between GPT-4o mini and its larger variant GPT-4o in text summarization.

\begin{table*}[ht]
\centering
\caption{Interval width and coverage on SummEval evaluated by G-Eval and ROSCOE evaluated by SocREval before boundary adjustment. We mark coverage $< 85\%$ with \graytext{gray text}, coverage between 85\% -- 90\% with \underline{underline} and coverage with the smallest interval width $\geq 90\%$ in \textbf{bold}.}
\resizebox{\textwidth}{!}{
\begin{tabular}{lcccccccc}
\toprule
\multirow{2.5}{*}{\textbf{Method}} & \multicolumn{4}{c}{\textbf{SummEval Evaluated with G-Eval }} & \multicolumn{4}{c}{\textbf{ROSCOE Evaluated with SocREval}} \\
\cmidrule(lr){2-5} \cmidrule(lr){6-9}
    & \textbf{Consistency} 
    & \textbf{Coherence} 
    & \textbf{Fluency} 
    & \textbf{Relevance} 
    & \textbf{CosmosQA} 
    & \textbf{DROP} 
    & \textbf{e-SNLI} 
    & \textbf{GSM8K} \\
\midrule
\multicolumn{9}{c}{\textbf{GPT-4o mini}} \\[2pt]
\midrule
CQR 
    & $1.15$ / $94.16\%$ 
    & \textbf{2.87 / 93.15\%} 
    & $1.44$ / $92.92\%$ 
    & \textbf{2.09 / 90.92\%} 
    & \textbf{3.53 / 95.27\%}
    & \textbf{3.82$ / $96.70\%} 
    & $3.04$ / $96.62\%$ 
    & \textbf{3.53 / 95.67\%} \\
Asym CQR 
    & $1.25$ / $94.97\%$ 
    & $2.91$ / $93.76\%$ 
    & $1.60$ / $93.75\%$ 
    & $2.13$ / $91.42\%$ 
    & $3.90$ / $98.71\%$ 
    & $3.91$ / $98.60\%$ 
    & \textbf{2.87/ 96.67\%} 
    & $3.89$ / $98.80\%$ \\
CHR
    & \underline{$0.67$ / $88.99\%$} 
    & \graytext{$2.41$ / $82.96\%$} 
    & \underline{$0.94$ / $88.86\%$} 
    & \graytext{$1.74$ / $82.62\%$} 
    & \graytext{$2.54$ / $73.06\%$} 
    & \graytext{$1.86$ / $68.92\%$} 
    & \graytext{$1.36$ / $72.24\%$} 
    & \graytext{$1.98$ / $78.67\%$} \\
LVD
    & $1.01$ / $92.35\%$ 
    & \underline{$2.73$ / $89.76\%$} 
    & \textbf{1.11 / 90.59\%} 
    & \underline{$2.02$ / $89.55\%$} 
    & \graytext{$3.10$ / $83.95\%$} 
    & \graytext{$2.49$ / $83.05\%$} 
    & \underline{$2.17$ / $86.18\%$} 
    & \underline{$3.08$ / $89.57\%$} \\
Boosted CQR 
    & \underline{$1.01$ / $87.75\%$} 
    & \underline{$2.73$ / $87.80\%$} 
    & \underline{$1.54$ / $88.68\%$} 
    & \underline{$2.00$ / $87.42\%$} 
    & \graytext{$3.15$ / $80.07\%$} 
    & \graytext{$2.63$ / $78.57\%$} 
    & \graytext{$1.82$ / $80.26\%$} 
    & \graytext{$3.08$ / $82.50\%$} \\
Boosted LCP 
    & \underline{$0.76$ / $89.22\%$} 
    & \underline{$2.67$ / $87.34\%$} 
    & \underline{$0.92$ / $89.18\%$} 
    & \underline{$1.91$ / $87.19\%$} 
    & \graytext{$3.60$ / $83.91\%$} 
    & \underline{$2.92$ / $85.40\%$} 
    & \graytext{$1.88$ / $81.23\%$} 
    & \underline{$3.36$ / $85.93\%$} \\
R2CCP
    & \textbf{0.69 / 90.88\%} 
    & \underline{$2.62$ / $89.63\%$} 
    & \underline{$0.92$ / $89.36\%$} 
    & \underline{$1.97$ / $89.70\%$} 
    & \underline{$2.96$ / $85.85\%$} 
    & \graytext{$2.43$ / $84.73\%$} 
    & \graytext{$1.75$ / $84.02\%$} 
    & \underline{$2.15$ / $85.07\%$} \\
\midrule
\multicolumn{9}{c}{\textbf{DeepSeek-R1-Distill-Qwen-32B}} \\[2pt]
\midrule
CQR 
    & $1.16$ / $93.88\%$ 
    & $2.67$ / $92.50\%$ 
    & $1.31$ / $93.01\%$ 
    & $2.13$ / $91.05\%$ 
    & \textbf{3.48 / 96.70\%} 
    & \textbf{3.83 / 96.35\%} 
    & $2.97$ / $96.36\%$ 
    & $3.46$ / $95.60\%$ \\
Asym CQR 
    & $1.30$ / $95.13\%$ 
    & $2.72$ / $92.86\%$ 
    & $1.49$ / $94.52\%$ 
    & $2.21$ / $92.06\%$ 
    & $3.84$ / $99.08\%$ 
    & $3.95$ / $99.27\%$ 
    & $2.86$ / $96.05\%$ 
    & $3.85$ / $98.43\%$ \\
CHR 
    & $0.82$ / $91.17\%$ 
    & \underline{$2.23$ / $87.07\%$} 
    & \underline{$0.90$ / $89.24\%$} 
    & \underline{$1.87$ / $86.38\%$} 
    & \graytext{$2.66$ / $76.50\%$} 
    & \graytext{$1.95$ / $78.06\%$} 
    & \graytext{$1.38$ / $71.97\%$} 
    & \graytext{$2.01$ / $81.60\%$} \\
LVD 
    & $0.97$ / $92.93\%$ 
    & $2.43$ / $91.10\%$ 
    & $1.00$ / $91.10\%$ 
    & \textbf{2.04 / 90.14\%} 
    & \underline{$3.25$ / $88.10\%$} 
    & \underline{$2.62$ / $88.06\%$} 
    & \textbf{2.24 / 90.96\%} 
    & \textbf{3.02$ / $90.63\%} \\
Boosted CQR
    & \underline{$1.10$ / $89.30\%$} 
    & \underline{$2.36$ / $88.98\%$} 
    & \underline{$1.16$ / $89.46\%$} 
    & \underline{$2.00$ / $88.98\%$} 
    & \graytext{$3.17$ / $82.72\%$} 
    & \graytext{$2.47$ / $81.11\%$} 
    & \graytext{$1.79$ / $80.96\%$} 
    & \graytext{$2.94$ / $79.83\%$} \\
Boosted LCP 
    & \underline{$0.77$ / $89.20\%$} 
    & \underline{$2.32$ / $86.70\%$} 
    & \underline{$0.93$ / $89.10\%$} 
    & \underline{$1.91$ / $86.89\%$} 
    & \graytext{$3.48$ / $81.60\%$} 
    & \underline{$2.79$ / $85.46\%$} 
    & \graytext{$1.84$ / $80.61\%$} 
    & \underline{$3.43$ / $85.23\%$} \\
R2CCP 
    & \textbf{0.69 / 90.44\%} 
    & \textbf{2.30 / 90.12\%} 
    & \textbf{ 0.89 / 90.09\%}
    & \underline{$2.00$ / $89.84\%$} 
    & \underline{$2.94$ / $86.97\%$} 
    & \underline{$2.29$ / $86.35\%$} 
    & \underline{$1.85$ / $87.87\%$} 
    & \underline{$1.88$ / $85.33\%$} \\
\midrule
\multicolumn{9}{c}{\textbf{Qwen2.5-72B-Instruct}} \\[2pt]
\midrule
CQR 
    & $0.98$ / $93.10\%$ 
    & $2.73$ / $92.25\%$ 
    & $1.44$ / $93.73\%$ 
    & $2.11$ / $91.30\%$
    & \textbf{3.37 / 94.80\%} 
    & $3.79$ / $97.02\%$ 
    & $3.01$ / $97.37\%$ 
    & $3.35$ / $95.33\%$ \\
Asym CQR 
    & $1.11$ / $94.47\%$ 
    & $2.80$ / $93.13\%$ 
    & $1.63$ / $94.79\%$ 
    & $2.17$ / $92.21\%$
    & $3.86$ / $99.01\%$ 
    & $3.89$ / $98.67\%$ 
    & \textbf{2.77 / 96.84\%} 
    & $3.87$ / $98.97\%$ \\
CHR 
    & \underline{$0.61$ / $89.04\%$} 
    & \graytext{$2.14$ / $80.93\%$} 
    & \underline{$0.98$ / $88.93\%$} 
    & \graytext{$1.61$ / $79.61\%$}
    & \graytext{$2.44$ / $72.65\%$} 
    & \graytext{$2.08$ / $75.87\%$} 
    & \graytext{$1.22$ / $69.69\%$} 
    & \graytext{$1.81$ / $77.50\%$} \\
LVD 
    & $0.85$ / $92.82\%$ 
    & $2.55$ / $90.49\%$ 
    & $1.09$ / $90.94\%$ 
    & \underline{$1.94$ / $89.27\%$}
    & \graytext{$3.05$ / $84.29\%$} 
    & \textbf{2.67 / 90.57\%} 
    & \underline{$1.91$ / $85.96\%$} 
    & \textbf{2.83/ 90.13\%} \\
Boosted CQR 
    & \underline{$0.80$ / $88.28\%$} 
    & \underline{$2.46$ / $87.82\%$} 
    & \underline{$1.24$ / $89.22\%$} 
    & \underline{$1.88$ / $87.17\%$}
    & \graytext{$3.05$ / $79.08\%$} 
    & \graytext{$2.56$ / $81.17\%$} 
    & \graytext{$1.51$ / $77.11\%$} 
    & \graytext{$2.81$ / $80.67\%$} \\
Boosted LCP 
    & \underline{$0.67$ / $88.81\%$} 
    & \underline{$2.43$ / $86.92\%$} 
    & \underline{$0.94$ / $89.26\%$} 
    & \underline{$1.86$ / $87.51\%$}
    & \graytext{$3.46$ / $80.41\%$} 
    & \underline{$2.81$ / $85.75\%$} 
    & \graytext{$1.74$ / $77.50\%$} 
    & \underline{$3.38$ / $86.23\%$} \\
R2CCP 
    & \textbf{0.61 / 90.73\%} 
    & \underline{$2.44$ / $89.54\%$} 
    & \textbf{0.95 / 90.18\%} 
    & \textbf{1.98 / 90.45\%}
    & \underline{$2.90$ / $85.34\%$} 
    & \underline{$2.39$ / $86.25\%$} 
    & \graytext{$1.59$ / $84.50\%$} 
    & \underline{$2.00$ / $86.73\%$} \\
\bottomrule
\end{tabular}
}
\label{tab:continuous_main}
\end{table*}

\paragraph{Conformal Prediction Methods.} In our experiments, we compare seven regression-based conformal prediction methods and two ordinal classification-based methods. For regression-based approaches, we use CQR~\citep{romano2019conformalized}, Asymmetric CQR~\citep{Sesia2019ACO}, CHR~\citep{sesia2021conformalpredictionusingconditional}, LVD~\citep{NEURIPS2021_46c7cb50}, and two boosted methods~\citep{xie2024boostedconformalpredictionintervals} namely Boosted CQR and Boosted LCP; for ordinal classification-based methods, we consider Ordinal APS~\citep{lu2022improvingtrustworthinessaidisease} and Ordinal Risk Control~\citep{xu2024conformalriskcontrolordinal}. Ordinal classification-based methods are only evaluated after boundary adjustment. More introduction of each method is provided in Appendix~\ref{appendix:CP Methods}. For each method, we split of dataset into 50\% calibration set and 50\% test set with 30 random seeds (1 -- 30) and report the mean of interval width and coverage rate.

\subsection{Continuous Intervals Indicate Uncertainty}
Table~\ref{tab:continuous_main} presents the interval width and coverage rate for each conformal prediction methods on SummEval evaluated by G-Eval and on ROSCOE evaluted SocREval. Additional results on DialSumm are presented in Table~\ref{tab:dialsumm_continuous}. From these results, most conformal predictors consistently output prediction intervals with coverage close to the $90\%$ in summarization tasks, indicating that LLM judges are confident when evaluating summarization tasks, especially on fluency. However, most methods fail to achieve $90\%$ (or close to $90\%$) coverage rate in reasoning tasks, and the interval widths are high in a few cases where 90\% coverage is achieved. One possible reason is that ROSCOE has much fewer samples, resulting in poor calibration.

\begin{figure}[ht]
    \centering
    \begin{subfigure}{\linewidth}
        \centering
        \includegraphics[width=\linewidth]{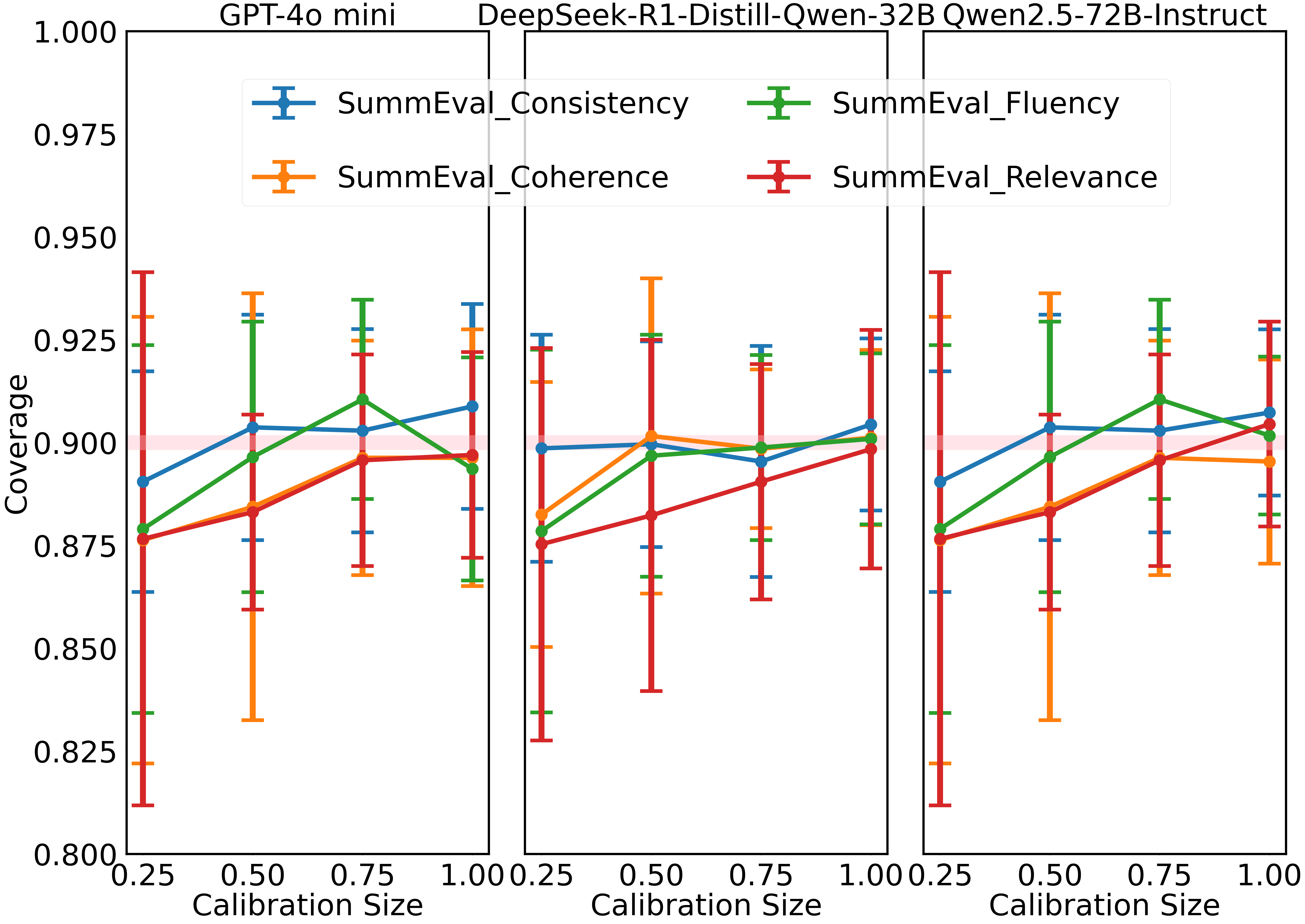}
        \label{fig:cal_size_summeval}
    \end{subfigure}
    \begin{subfigure}{\linewidth}
        \centering
        \includegraphics[width=\linewidth]{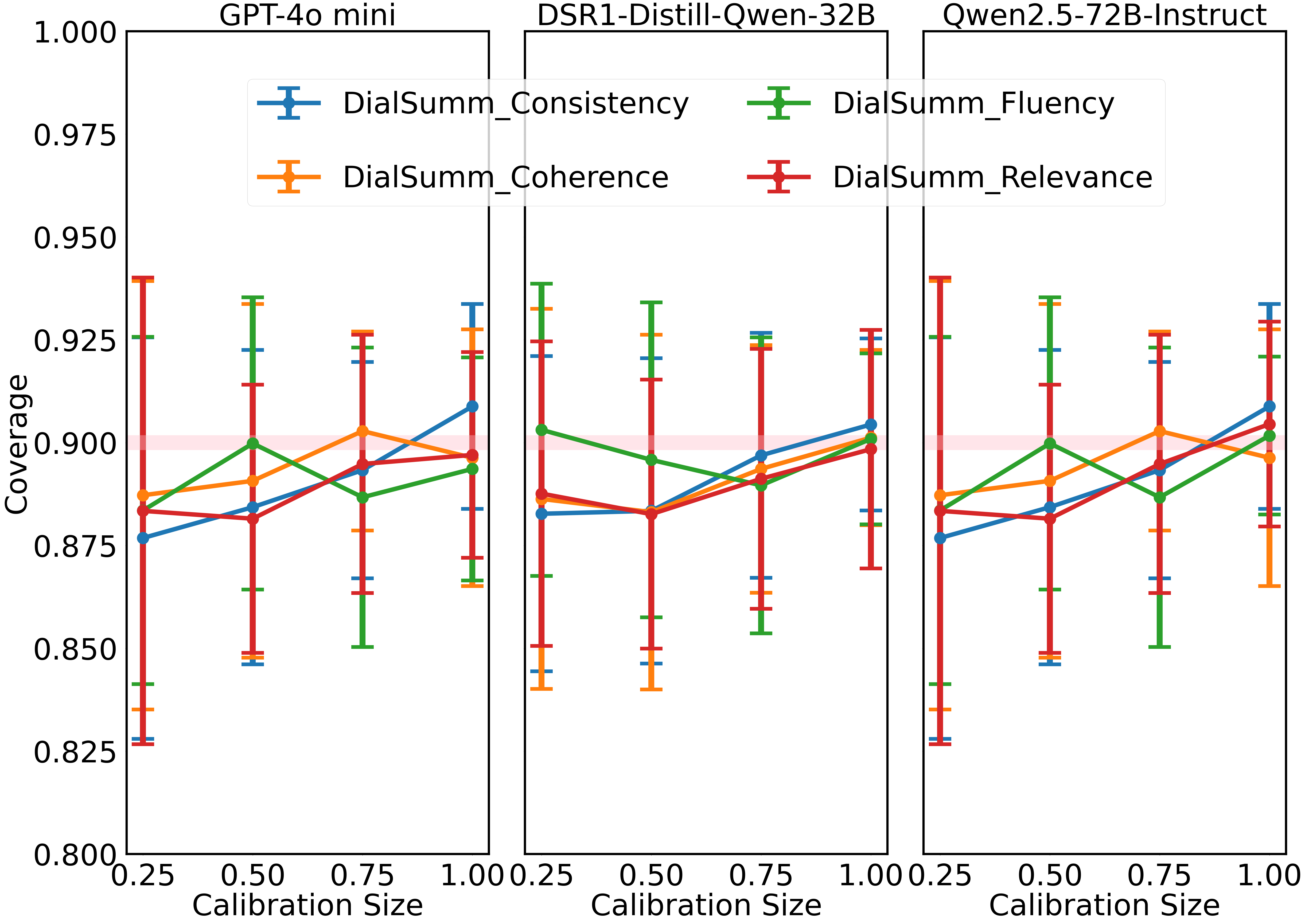}
        \label{fig:cal_size_dialsumm}
    \end{subfigure}
    \caption{Coverage vs. calibration set size. Mean coverage rates increase to 90\% and error bars shrinks, as calibration set increases.}
    \label{fig:cal_size_combined}
\end{figure}

\subsection{All Coverages Improve after Adjustment}
Table~\ref{tab:discrete_main} presents the interval width and coverage rate for each conformal prediction methods on SummEval evaluated by G-Eval and on ROSCOE evaluted SocREval. Additional results on DialSumm are presented in Table~\ref{tab:dialsumm_discrete}. From these results, we observe that all coverage rates are improved after boundary adjustment. In most settings, prediction intervals after boundary adjustment are close to or above the desired $90\%$ coverage rate across all datasets and judge frameworks. The effectiveness of boundary adjustment is brought by the fact that continuous intervals before boundary adjustment is sensitive to the variability of quantile estimation during calibration in conformal prediction.

Moreover, improvements to coverage brought by boundary adjustment are empirically robust across different datasets, conformal prediction methods, LLM judges and judge frameworks. For example, coverage rates are consistently higher than $90\%$ in SummEval and DialSumm, where coverage rates before boundary adjustment are in the range of $83\%–88\%$. A even more significant example is LVD, which, when applied to e-SNLI using Qwen2.5-72B-Instruct under SocREval, shows an increase from 85.96\% to 95.53\%.

\begin{table*}[ht]
\centering
\caption{Interval width and coverage on SummEval evaluated by G-Eval and ROSCOE evaluated by SocREval after boundary adjustment. We mark coverage $< 85\%$ with \graytext{gray text}, coverage between 85\% -- 90\% with \underline{underline} and coverage $\geq 90\%$ with the smallest interval width in \textbf{bold}.}
\resizebox{\textwidth}{!}{%
\begin{tabular}{lcccccccc}
\toprule
\multirow{2.5}{*}{\textbf{Method}} & \multicolumn{4}{c}{\textbf{SummEval Evaluated with G-Eval }} & \multicolumn{4}{c}{\textbf{ROSCOE Evaluated with SocREval}} \\
\cmidrule(lr){2-5} \cmidrule(lr){6-9}
    & \textbf{Consistency} 
    & \textbf{Coherence} 
    & \textbf{Fluency} 
    & \textbf{Relevance} 
    & \textbf{CosmosQA} 
    & \textbf{DROP} 
    & \textbf{e-SNLI} 
    & \textbf{GSM8K} \\
\midrule
\multicolumn{9}{c}{\textbf{GPT-4o mini}} \\[2pt]
\midrule
CQR 
    & $1.15$ / $95.45\%$ 
    & $2.87$ / $94.94\%$ 
    & $1.44$ / $93.80\%$ 
    & $2.09$ / $93.56\%$ 
    & $3.53$ / $95.34\%$ 
    & $3.82$ / $97.05\%$ 
    & $3.04$ / $96.89\%$ 
    & $3.53$ / $95.67\%$ \\
Asym CQR 
    & $1.25$ / $96.02\%$ 
    & $2.90$ / $95.41\%$ 
    & $1.60$ / $94.57\%$ 
    & $2.14$ / $94.14\%$ 
    & $3.90$ / $98.84\%$ 
    & $3.91$ / $98.73\%$ 
    & $2.87$ / $96.89\%$ 
    & $3.89$ / $98.80\%$ \\
CHR 
    & $0.70$ / $91.79\%$ 
    & \underline{$2.41$ / $87.78\%$} 
    & $0.94$ / $90.60\%$ 
    & \underline{$1.74$ / $88.10\%$} 
    & \graytext{$2.56$ / $82.45\%$} 
    & \graytext{$1.87$ / $78.86\%$} 
    & \graytext{$1.34$ / $83.46\%$} 
    & \graytext{$1.94$ / $83.23\%$} \\
LVD 
    & $1.01$ / $94.11\%$ 
    & $2.73$ / $93.72\%$ 
    & $1.12$ / $92.70\%$ 
    & $2.03$ / $93.82\%$ 
    & \textbf{3.13 / 91.53\%} 
    & \textbf{2.52 / 90.22\%} 
    & $2.17$ / $94.82\%$ 
    & \textbf{3.09$ / $93.37\%} \\
Boosted CQR 
    & $0.99$ / $92.81\%$ 
    & $2.73$ / $93.02\%$ 
    & $1.54$ / $94.38\%$ 
    & $2.00$ / $92.93\%$ 
    & $3.20$ / $93.40\%$ 
    & $2.63$ / $89.65\%$ 
    & $1.82$ / $92.15\%$ 
    & $3.09$ / $91.17\%$ \\
Boosted LCP 
    & $0.74$ / $91.90\%$ 
    & $2.68$ / $93.53\%$ 
    & \textbf{0.90 / 90.88\%}
    & \textbf{1.91 / 92.70\%} 
    & $3.60$ / $95.48\%$ 
    & $3.01$ / $91.27\%$ 
    & $1.90$ / $91.80\%$ 
    & $3.26$ / $92.17\%$ \\
R2CCP 
    & \textbf{0.68 / 92.15\%} 
    & \textbf{2.62 / 92.81\%} 
    & $0.91$ / $90.99\%$ 
    & $1.97$ / $93.38\%$ 
    & \underline{$2.93$ / $89.46\%$} 
    & \underline{$2.41$ / $89.21\%$} 
    & \textbf{1.71 / 90.11\%} 
    & \underline{$2.09$ / $86.93\%$} \\
OrdinalAPS  
    & \graytext{$2.28$ / $71.48\%$} 
    & \graytext{$1.88$ / $64.84\%$} 
    & \graytext{$1.78$ / $13.65\%$} 
    & \graytext{$2.36$ / $87.94\%$} 
    & \graytext{$0.73$ / $47.52\%$} 
    & \graytext{$0.83$ / $55.08\%$} 
    & \graytext{$0.72$ / $52.76\%$} 
    & \graytext{$0.58$ / $73.90\%$} \\
OrdinalRC 
    & \graytext{$2.41$ / $75.19\%$} 
    & \graytext{$2.02$ / $67.38\%$} 
    & \graytext{$1.93$ / $14.58\%$} 
    & $2.51$ / $90.30\%$ 
    & \graytext{$0.82$ / $49.46\%$} 
    & \graytext{$0.91$ / $57.11\%$} 
    & \graytext{$0.80$ / $54.61\%$} 
    & \graytext{$0.60$ / $74.43\%$} \\
\midrule
\multicolumn{9}{c}{{\textbf{DeepSeek-R1-Distill-Qwen-32B}}} \\[2pt]
\midrule
CQR 
    & $1.15$ / $95.02\%$ 
    & $2.67$ / $94.34\%$ 
    & $1.32$ / $94.44\%$ 
    & $2.13$ / $93.67\%$ 
    & $3.48$ / $96.80\%$ 
    & $3.82$ / $96.54\%$ 
    & $2.99$ / $96.80\%$ 
    & $3.46$ / $95.63\%$ \\
Asym CQR 
    & $1.31$ / $95.99\%$ 
    & $2.72$ / $94.83\%$ 
    & $1.49$ / $95.57\%$ 
    & $2.21$ / $94.53\%$ 
    & $3.84$ / $99.08\%$ 
    & $3.95$ / $99.27\%$ 
    & $2.88$ / $96.45\%$ 
    & $3.84$ / $98.47\%$ \\

CHR 
    & $0.87$ / $93.96\%$ 
    & \textbf{2.23 / 91.42\%} 
    & $0.91$ / $91.98\%$ 
    & \textbf{1.87 / 90.84\%} 
    & \underline{$2.69$ / $86.80\%$} 
    & \underline{$1.97$ / $85.90\%$} 
    & \underline{$1.39$ / $85.96\%$} 
    & \underline{$2.01$ / $86.60\%$} \\
LVD 
    & $0.97$ / $95.01\%$ 
    & $2.44$ / $94.58\%$ 
    & $1.00$ / $93.21\%$ 
    & $2.04$ / $94.12\%$ 
    & $3.28$ / $95.27\%$ 
    & $2.67$ / $93.75\%$ 
    & $2.24$ / $96.36\%$ 
    & $3.03$ / $94.40\%$ \\
Boosted CQR 
    & $1.08$ / $93.55\%$ 
    & $2.37$ / $93.96\%$ 
    & $1.15$ / $93.48\%$ 
    & $2.01$ / $93.72\%$ 
    & $3.20$ / $95.71\%$ 
    & \textbf{2.52 / 93.30\%} 
    & \textbf{1.79 / 93.25\%} 
    & $2.94$ / $92.23\%$ \\
Boosted LCP 
    & $0.76$ / $92.03\%$ 
    & $2.32$ / $92.37\%$ 
    & $0.93$ / $91.34\%$ 
    & $1.92$ / $92.81\%$ 
    & $3.46$ / $95.95\%$ 
    & $2.80$ / $91.94\%$ 
    & $1.87$ / $92.89\%$ 
    & $3.36$ / $93.63\%$ \\
R2CCP 
    & \textbf{0.68 / 91.57\%} 
    & $2.30$ / $93.22\%$ 
    & \textbf{0.89 / 91.80\%} 
    & $1.99$ / $92.96\%$ 
    & \textbf{2.91 / 90.58\%} 
    & \underline{$2.25$ / $89.97\%$} 
    & \textbf{1.80 / 92.35\%} 
    & \underline{$1.82$ / $86.93\%$} \\
OrdinalAPS 
    & $2.51$ / $90.06\%$ 
    & $2.52$ / $90.64$ 
    & $3.76$ / $91.08\%$ 
    & $2.13$ / $89.98\%$ 
    & \graytext{$1.32$ / $60.00\%$} 
    & \graytext{$1.26$ / $78.22\%$} 
    & \underline{$1.46$ / $87.85\%$} 
    & \underline{$1.50$ / $85.67\%$} \\
OrdinalRC 
    & $2.54$ / $90.11\%$ 
    & $2.56$ / $91.18\%$ 
    & \underline{$3.73$ / $89.53\%$} 
    & $2.14$ / $90.07\%$ 
    & \graytext{$1.44$ / $62.35\%$} 
    & \graytext{$1.33$ / $78.22\%$} 
    & \underline{$1.52$ / $88.33\%$} 
    & \underline{$1.55$ / $86.07\%$} \\
\midrule
\multicolumn{9}{c}{\textbf{Qwen2.5-72B-Instruct}} \\[2pt]
\midrule
CQR 
    & $0.98$ / $94.35\%$ 
    & $2.72$ / $94.18\%$ 
    & $1.45$ / $94.79\%$ 
    & $2.10$ / $94.02\%$ 
    & $3.36$ / $95.07\%$ 
    & $3.79$ / $97.08\%$ 
    & $3.01$ / $97.68\%$ 
    & $3.34$ / $95.33\%$ \\
Asym CQR 
    & $1.10$ / $95.47\%$ 
    & $2.79$ / $94.70\%$ 
    & $1.64$ / $95.63\%$ 
    & $2.17$ / $94.85\%$ 
    & $3.85$ / $99.18\%$ 
    & $3.89$ / $98.67\%$ 
    & $2.77$ / $97.06\%$ 
    & $3.87$ / $98.97\%$ \\
CHR 
    & $0.66$ / $92.21\%$ 
    & \underline{$2.14$ / $86.10\%$} 
    & $0.98$ / $91.16\%$ 
    & \underline{$1.61$ / $85.78\%$} 
    & \graytext{$2.49$ / $82.14\%$} 
    & \graytext{$2.05$ / $82.89\%$} 
    & \graytext{$1.18$ / $84.56\%$} 
    & \underline{$1.79$ / $85.27\%$} \\
LVD 
    & $0.85$ / $95.11\%$ 
    & $2.56$ / $94.05\%$ 
    & $1.09$ / $93.45\%$ 
    & $1.95$ / $93.86\%$ 
    & \textbf{3.07 / 92.01\%} 
    & $2.67$ / $93.87\%$ 
    & $1.91$ / $95.53\%$ 
    & $2.87$ / $93.43\%$ \\
Boosted CQR 
    & $0.81$ / $92.36\%$ 
    & $2.47$ / $93.06\%$ 
    & $1.25$ / $93.66\%$ 
    & $1.88$ / $92.81\%$ 
    & $3.10$ / $94.01\%$ 
    & $2.56$ / $90.79\%$ 
    & \textbf{1.49 / 92.11\%} 
    & $2.82$ / $92.03\%$ \\
Boosted LCP 
    & $0.65$ / $91.26\%$ 
    & $2.44$ / $92.26\%$ 
    & $0.93$ / $91.20\%$ 
    & \textbf{1.86 / 92.57\%} 
    & $3.40$ / $94.90\%$ 
    & $2.84$ / $92.41\%$ 
    & $1.79$ / $91.84\%$ 
    & $3.33$ / $92.90\%$ \\
R2CCP 
    & \textbf{0.59 / 91.83\%} 
    & \textbf{2.43 / 92.78\%} 
    & \textbf{0.95 / 92.12\%} 
    & $1.98$ / $93.72\%$ 
    & \underline{$2.88$ / $89.29\%$} 
    & \textbf{2.34 / 90.00\%} 
    & $1.55$ / $90.20\%$ 
    & \underline{$1.96$ / $88.57\%$} \\
OrdinalAPS 
    & $2.86$ / $90.18\%$ 
    & $3.01$ / $90.59\%$ 
    & \graytext{$3.05$ / $45.43\%$} 
    & $2.75$ / $90.29\%$ 
    & \graytext{$0.71$ / $55.99\%$} 
    & \graytext{$0.25$ / $56.83\%$} 
    & \graytext{$0.67$ / $77.68\%$} 
    & \graytext{$0.46$ / $70.87\%$} \\
OrdinalRC 
    & $2.85$ / $90.00\%$ 
    & \underline{$2.96$ / $89.35\%$} 
    & \graytext{$3.21$ / $53.31\%$} 
    & $2.75$ / $90.14\%$ 
    & \graytext{$0.75$ / $57.28\%$} 
    & \graytext{$0.29$ / $56.83\%$} 
    & \graytext{$0.80$ / $79.74\%$} 
    & \graytext{$0.49$ / $71.37\%$} \\
\bottomrule
\end{tabular}
}
\label{tab:discrete_main}
\end{table*}

\begin{table*}[ht]
\centering
\caption{Interval midpoints vs. Raw Score vs. Weighted Average (Avg.). Con R2CCP refers to midpoint of R2CCP interval before boundary adjustment, and Dis R2CCP refers to midpoint of R2CCP interval after boundary adjustment. Mean squared error (MSE), mean absolute error (MAE), mean Spearman’s $\rho$ and mean Kendall’s $\tau$ are calculated over 30 experiments. \textbf{Bold} indicates better performance than both baselines; \graytext{Gray} indicates worse performance than both baselines; Otherwise we mark the results \underline{underlined}.}
\resizebox{\textwidth}{!}{%
\begin{tabular}{l|lrrr|rrrr|rrrr|rrrr}
\toprule
\multirow{2.5}{*}{\textbf{Method}} & \multicolumn{4}{c}{\textbf{Coherence}} & \multicolumn{4}{c}{\textbf{Consistency}} & \multicolumn{4}{c}{\textbf{Fluency}} & \multicolumn{4}{c}{\textbf{Relevance}}    \\
\cmidrule(lr){2-5} \cmidrule(lr){6-9} \cmidrule(lr){10-13} \cmidrule(lr){14-17} 
    & \textbf{MSE} & \textbf{MAE} & $\rho$ & $\tau$ 
    & \textbf{MSE} & \textbf{MAE} & $\rho$ & $\tau$ 
    & \textbf{MSE} & \textbf{MAE} & $\rho$ & $\tau$ 
    & \textbf{MSE} & \textbf{MAE} & $\rho$ & $\tau$ \\
\midrule
\multicolumn{17}{c}{\textbf{GPT-4o mini}} \\
\midrule
Raw  Score 
    & 1.729 & 1.055 & 0.446 & 0.373 
    & 1.674 & 1.073 & 0.480 & 0.437 
    & 3.907 & 1.977 & 0.219 & 0.197
    & 1.009 & 0.786 & 0.512 & 0.427 \\
Weighted Avg. 
    & 1.643 & 1.037 & 0.514 & 0.379 
    & 1.548 & 1.066 & 0.478 & 0.383 
    & 3.412 & 1.733 & 0.319 & 0.250 
    & 0.865 & 0.737 & 0.567 & 0.419 \\
Con R2CCP 
    &\textbf{ 0.791} & \textbf{0.716} & \underline{0.512} & \underline{0.373} 
    & \textbf{0.510} & \textbf{0.432} & \graytext{0.455} & \graytext{}{0.371} 
    & \textbf{0.442} & \textbf{0.491} & \textbf{0.330} & \textbf{0.261} 
    & \textbf{0.418} & \textbf{0.509} & \underline{0.546} & \graytext{0.403} \\
Dis R2CCP 
    & \textbf{0.794} & \textbf{0.715} & \underline{0.508} & \textbf{0.386} 
    & \textbf{0.512} & \textbf{0.428} & \textbf{0.506} & \textbf{0.468} 
    & \textbf{0.443} & \textbf{0.488} & \textbf{0.336} & \textbf{0.300} 
    & \textbf{0.423} & \textbf{0.509} & \underline{0.540} & \underline{0.423 }\\
\midrule
\multicolumn{17}{c}{\textbf{DeepSeek-R1-Distill-Qwen-32B}} \\
\midrule
Raw Score 
    & 1.010 & 0.775 & 0.549 & 0.457 
    & 1.229 & 0.770 & 0.467 & 0.425 
    & 2.843 & 1.549 & 0.387 & 0.355 
    & 0.763 & 0.682 & 0.520 & 0.437 \\
Weighted Avg. 
    & 0.869 & 0.734 & 0.599 & 0.447 
    & 1.439 & 1.065 & 0.468 & 0.375 
    & 2.783 & 1.564 & 0.420 & 0.332 
    & 0.646 & 0.632 & 0.565 & 0.419 \\
Con R2CCP 
    & \textbf{0.599} & \textbf{0.619} & \textbf{0.663} & \textbf{0.492} 
    & \textbf{0.564} & \textbf{0.446} & \graytext{0.445} & \graytext{0.361} 
    & \textbf{0.373} & \textbf{0.455} & \underline{0.391} & \graytext{0.311} 
    & \textbf{0.431} & \textbf{0.513} & \underline{0.555} & \graytext{0.412} \\
Dis R2CCP 
    & \textbf{0.602} & \textbf{0.619} & \textbf{0.661} & \textbf{0.508} 
    & \textbf{0.566} & \textbf{0.441} & \underline{0.462} & \underline{0.423} 
    & \textbf{0.375} & \textbf{0.454} & \underline{0.393} & \underline{0.351} 
    & \textbf{0.434} & \textbf{0.512} & \underline{0.548} & \underline{0.431} \\
\midrule
\multicolumn{17}{c}{\textbf{Qwen2.5-72B-Instruct}} \\
\midrule
Raw Score 
    & 1.432 & 0.981 & 0.426 & 0.358 
    & 2.068 & 1.237 & 0.458 & 0.416 
    & 4.476 & 1.958 & 0.310 & 0.281 
    & 1.188 & 0.903 & 0.498 & 0.420 \\
Weighted Avg. 
    & 1.282 & 0.932 & 0.539 & 0.395 
    & 1.847 & 1.213 & 0.483 & 0.387 
    & 4.236 & 1.928 & 0.363 & 0.285 
    & 1.091 & 0.885 & 0.555 & 0.412 \\
Con R2CCP 
    & \textbf{0.675} & \textbf{0.659} & \textbf{0.603} & \textbf{0.444} 
    & \textbf{0.469} & \textbf{0.396} & \underline{0.465} & \underline{0.378} 
    & \textbf{0.414 }& \textbf{0.486} & \underline{0.340} & \graytext{0.269} 
    & \textbf{0.407} & \textbf{0.502} & \textbf{0.571} & \textbf{0.425} \\
Dis R2CCP 
    & \textbf{0.678} & \textbf{0.659} & \textbf{0.600} & \textbf{0.456} 
    & \textbf{0.469} & \textbf{0.387} & \textbf{0.538} & \textbf{0.498} 
    & \textbf{0.416} & \textbf{0.485} & \underline{0.342} & \textbf{0.306} 
    & \textbf{0.411} & \textbf{0.501} & \textbf{0.566} & \textbf{0.444} \\
\bottomrule
\end{tabular}
}
\label{tab:midpoints_summeval}
\end{table*}

\subsection{Recommended Choice from Reliable LLM-as-a-Judge}
Our experiments show that DeepSeek-R1-Distill-Qwen-32B provides the most consistent coverage (surpassing Qwen2.5-72B-Instruct and GPT-4o mini), while Qwen2.5-72B-Instruct typically yields the narrowest intervals. 
Under the G-Eval framework, we observe higher coverage (at the cost of slightly wider intervals) than SocREval. 
Regarding the choice of conformal prediction methods, R2CCP strikes the best balance between coverage and width; Boosted LCP performs comparably but less efficiently; And LVD delivers very tight intervals without sacrificing coverage. 
In practice, our experiments suggest that DeepSeek-R1-Distill-Qwen-32B + G-Eval + LVD might be the best option for potentially high-stake applications, whereas Qwen2.5-72B-Instruct + R2CCP + SocREval could yield the most efficient prediction interval after boundary adjustment.

\subsection{Midpoints Reduce Bias}
Here we evaluate whether the midpoint of a prediction interval could better represent human preference. Since R2CCP is considered the best conformal prediction method that balances coverage and efficiency (i.e., interval width), we evaluate two midpoints computation: midpoint of R2CCP interval before boundary adjustment (Con R2CCP) and midpoint of R2CCP interval after boundary adjustment (Dis R2CCP). We compare these two midpoints with two baselines: the raw score output by LLM judge and the weighted average of token probabilities, both of which are used in G-Eval. 

Our comparison results are shown in Table~\ref{tab:midpoints_summeval} and Table~\ref{tab:midpoints_dialsumm}. From the table, we can see that, in addition to achieving comparable or slightly better correlation with respect to ground truth, midpoint estimates achieve substantially lower mean squared error (MSE) and mean absolute error (MAE). For example, the midpoints from Con R2CCP on fluency of SummEval evaluated by GPT-4o mini reduce $88.7\%$ of MSE, from 3.907 to 0.443 (Table~\ref{tab:midpoints_summeval}). Moreover, MAE between midpoints and true ratings are consistently lower than 0.5, which shows that midpoint is a less biased estimate of human preference. Empirically, though weighted average could achieve a comparable correlation, it often fall outside of the prediction interval due to model bias, while midpoints are on average closer to ground truth.  More results on DialSumm (Table~\ref{tab:midpoints_dialsumm}) and on multimodal text-to-image consistency (Table~\ref{tab:genaibench}) yield similar observation that midpoint is consistently a more accurate alternative.

\section{Analysis}
\subsection{Proper Calibration Improves Coverage}~\label{sec:calsize}
Given the limited sample size in ROSCOE and the results shown in Table~\ref{tab:continuous_main}, we explore the relationship between the size of calibration set and the coverage. We construct a continuous prediction interval before boundary adjustment using R2CCP under four calibration regimes -- 25\%, 50\%, 75\%, and 100\% of the calibration set. Figure~\ref{fig:cal_size_combined} shows the coverage with varying calibration set size. From the figure, it is clear that the mean coverage increases and the error bar shrinks as calibration set size increases. These results highlight the importance of a sufficiently large calibration set to ensure a stable coverage rate.

\subsection{Why Boundary Adjustment is Effective}
\begin{figure}[ht]
    \centering
    \begin{subfigure}{\linewidth}
        \centering
        \includegraphics[width=1\linewidth]{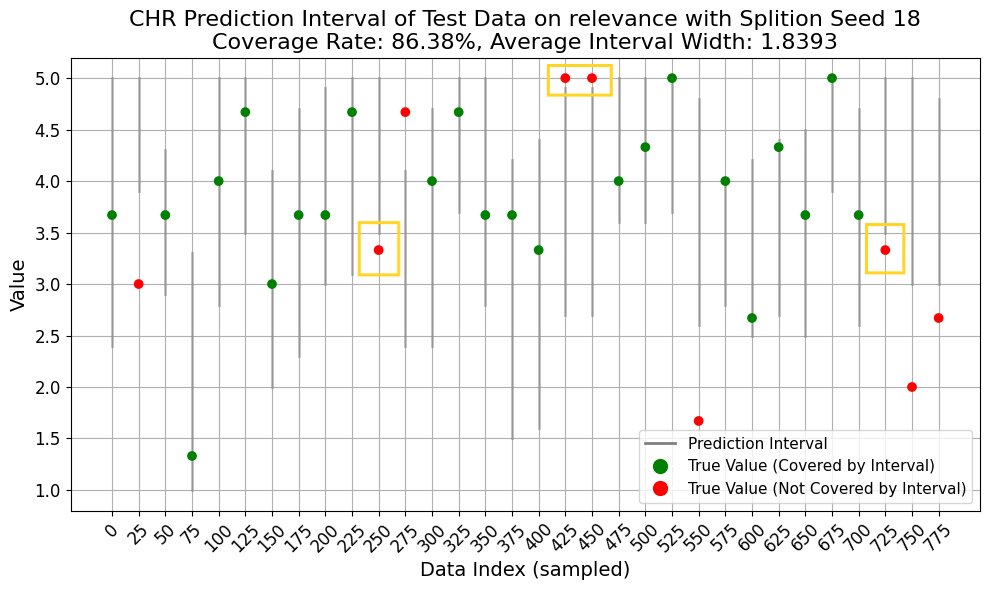}
    \end{subfigure}

    \begin{subfigure}{\linewidth}
        \centering
        \includegraphics[width=1\linewidth]{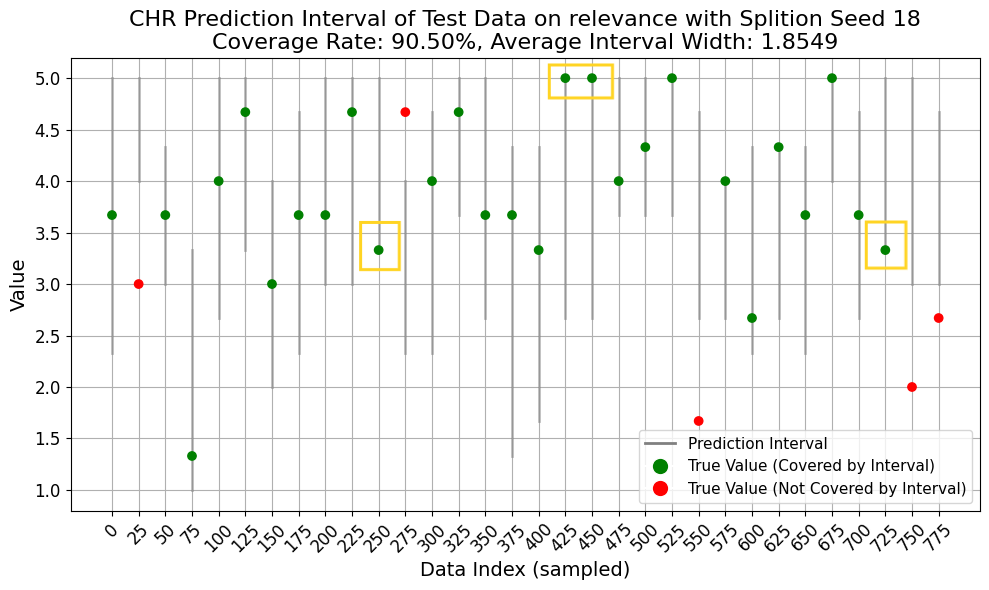}
    \end{subfigure}

    \caption{Red points mean the labels lying outside the intervals, which could turn green (inside) if the interval just extend to nearest labels (e.g. 3.33 and 5). After applying boundary adjustment, the coverage in this instance improves from 86.38\% to 90.50\%, while the average width increases slightly to 1.8549.}
    \label{fig:before_after_adjustment}
\end{figure}

We analyze why boundary adjustment could consistently improve coverage empirically in all settings. In Figure~\ref{fig:before_after_adjustment}, we plot the ground truth value (green if falling within prediction interval or red if falling outside) and the prediction interval (gray vertical line). From the figure, we can see that there are several ground-truth scores that fall outside of the interval are very close to the endpoints. Thus, a slight expansion to the discrete rating value would suffice to cover the ground truth. This demonstrates that a relatively small expansion of the prediction interval can lead to a substantial gain in calibration, successfully achieving the 90\% coverage.

\subsection{Reprompt and Regrade with Intervals}
In addition to use midpoint as an alternative judgment, we explore whether reprompting the LLM judge with information about the prediction interval could improve the judgment. We reprompt the LLM judge with the best interval among 30 runs on ROSCOE (R2CCP + DeepSeek-R1-Distill-Qwen-32B) (Table~\ref{tab:reprompt_results}). Our investigation shows that: (1) if the initial ratings fall within the prediction interval, reprompting could strengthen the judge confidence in initial ratings (Figures~\ref{fig:reprompt_example1}, ~\ref{fig:reprompt_example2} and~\ref{fig:reprompt_example3}); (2) If the initial ratings fall outside of the prediction interval, the judge might resist in changing to another rating that falls within the interval (Figures~\ref{fig:reprompt_example4} and~\ref{fig:reprompt_example5}). After deeper analysis, we found the main cause is that the model is not allowed to output intermediate scores (e.g., 4.33, 4.67), and it thinks moving to the next integer (e.g., from 4 to 5) is unreasonable. If given the option to output intermediate score, it often changes its rating to the nearest bound of the prediction interval (see an example in Figure~\ref{fig:reprompt_example8}). For example, with initial score being 4 and interval being [4.33, 5.00], the judge typically raises its rating from 4 to 4.33.

\section{Discussion}
In this work, we provide the first analysis of uncertainty in LLM-as-a-judge using conformal prediction based on the output logits in a single evaluation run. Our analysis aims to construct prediction intervals that achieve or approximate 90\% coverage rate, and includes nine conformal prediction methods, three LLM evaluators, two evaluation frameworks (i.e., G-Eval, SocREval) and two tasks (summarization and reasoning). Moreover, we design an intuitive yet theoretically grounded boundary adjustment technique that transforms continuous intervals to discrete rating scales, yielding improvements in coverage. Finally, we explore the use of interval midpoints as calibrated scores, to understand how prediction interval could help within the LLM-as-a-judge paradigm. Experimental results demonstrate that this strategy matches or slightly surpasses baselines on correlation metrics while significantly outperforming direct scoring on error metrics, thereby achieving higher accuracy. 

We believe this work take a step towards reliable LLM-as-a-judge. Our goal is to provide reference to help user determine when they can trust the judgment through uncertainty given by the width of prediction interval and sheds lights on the necessity of reliable LLM-as-a-judge. 
On the one hand, a wide prediction interval serves as a warning signal of unreliability with the LLM judge provided score, which might benefit in high-risk environments where uncertainty-induced errors must be minimized, such as in medical diagnosis~\citep{lu2022improvingtrustworthinessaidisease, tan2024finetuninglargelanguagemodel}. On the other hand, a narrow prediction interval suggests a higher degree of certainty in the score, thereby reducing the need for manual review in automated evaluation, such as in essay scoring~\citep{10520824}. We believe our framework might be helpful in example selection to avoid the model collapse when trained on LLM-generated data via active learning~\citep{Shumailov2024} or to help with reinforcement learning with a reliable AI feedback.
\section*{Acknowledgements}
This work is supported by a gift funding from Amazon Responsible AI. We also thank all reviewers and the area chair for their valuable input that led to improvements to the paper.

\section*{Limitations}
The main limitation of this work lies in the coverage of tasks that LLMs are used to judge. We primarily analyze summarization and reasoning for natural language generation through rating-based evaluations such as SummEval, DialSumm and ROSCOE. We acknowledge that there could be many other tasks that we have yet to explore, including but not limited to machine translation, multimodal generation, etc.

\section*{Ethical Considerations}
Our work analyzes how reliable LLM-as-a-judge is in rating-based evaluation using conformal prediction. Though conformal prediction can quantify uncertainty with statistical guarantee, it may impose certain ethical considerations. 

First, our framework, as well as other conformal prediction methods, relies on high-quality human annotations for calibration. If these annotations contain subjective or biased judgments, the resulting calibrated prediction intervals may reflect such subjective opinion or biases, which might further distort the evaluations. For example, a LLM judge could reduce workload in essay scoring~\citep{10520824} with calibration from teacher evaluations. But biased annotations may systematically underrate certain linguistic styles, leading to unfairly low score and wide prediction intervals for essays with these linguistic styles. Second, our interval estimates are nonparametric and should not be interpreted as classical confidence intervals in statistics. The midpoint is a heuristic and convenient choice, but not a statistical mean or mode, and does not imply symmetric or continuous uncertainty. Misinterpreting this could result in misleading conclusions.

\bibliography{main}
\bibliographystyle{acl_natbib}

\newpage
\appendix

\section{Appendix}

\subsection{Proof of Theorem 1}~\label{appendix:proofof ba coverage}
\begin{proof}
By the standard split conformal prediction procedure with the nonconformity score $s\left(z, y\right) = |\hat{y} - y|$, the prediction set 
\begin{equation}
    \mathcal{C}\left(z_{\rm test}\right) = \left\{a \in \mathbb{R} : s\left(x_{\rm test}, z\right) \le \hat{q}_{1 - \alpha}\right\}
\end{equation}
satisfies $\mathbb{P}(Y_{\rm test}\in\mathcal{C}\left(z_{\rm test}\right))\ge1-\alpha.$

In our discrete setting, every potential label is an element of a predetermined ordered set (e.g., $\left\{1, 2, 3, 4, 5\right\}$). The adjusted score $s'\left(z, y\right)$ is defined such that for each $y$,
\begin{equation}
    s'(z, y) = s\left(z, y'\right)
\end{equation}
where $y'$ is the label closest to $y$. 

In regions where the original interval $\mathcal{C}\left(z_{\rm test}\right) = \left[l, u\right]$ already contains some label, the shrinking adjustment leads to
\begin{equation}
    s'\left(z_{\rm test}, l\right) = s\left(z_{\rm test}, \lceil l\rceil\right) \le \hat{q}_{1 - \alpha}
\end{equation}
or 
\begin{equation}
    s'\left(z_{\rm test}, u\right) = s\left(z_{\rm test}, \lfloor u\rfloor\right) \le \hat{q}_{1 - \alpha}.
\end{equation}
Thus, every label that was originally covered (i.e., satisfying $s\left(z_{\rm test}, y\right) \le \hat{q}$) remains covered. Thus, the coverage remains unchanged.

Now suppose that an expansion to the interval is performed. We have 
\begin{equation}
    s'\left(z_{\rm test}, l\right) \le \hat{q}_{1 - \alpha} \le s'\left(z_{\rm test}, \lfloor l\rfloor\right) \le \hat{q}_{1 - \alpha_0} 
\end{equation}
or
\begin{equation}
    s'\left(z_{\rm test}, u\right) \le \hat{q}_{1 - \alpha} \le s'\left(z_{\rm test}, \lceil u\rceil\right) \le \hat{q}_{1 - \alpha_0}
\end{equation}
where $0 \le \alpha_0 < \alpha$. 

In this case, for any $a \notin \mathcal{C}\left(z_{\rm test}\right)$, it is possible that $z \in \mathcal{C}'\left(z_{\rm test}\right)$ if $z$ is either $\lfloor l \rfloor$ or $\lceil u \rceil$. As a consequence, if the original interval barely missed covering the label, the expansion guarantees that these outcomes are now covered.

\begin{table*}[ht!]
\centering
\caption{Breusch-Pagan (BP) and White tests detect pervasive heteroscedasticity in SummEval and DialSumm: both tests yield highly significant $p$-values ($p$ < 1e-12) across all four metrics and all evaluators. By contrast, in ROSCOE by G-Eval only DROP, e-SNLI and GSM8k exhibit significant heteroscedasticity (p < 0.05) while CosmosQA remains homoscedastic; in ROSCOE by SocREval heteroscedasticity is confined to DROP for DeepSeek-R1-Qwen-32B (DSR1-Qwen-32B) and GPT-4o mini and to CosmosQA and e-SNLI for Qwen2.5-72B-Instruct (Qwen2.5-72B).}
\resizebox{\textwidth}{!}{
\begin{tabular}{llrrrr|rrrr|rrrr|rrrr}
\toprule 
\multicolumn{18}{c}{\textbf{SummEval by G-Eval}}\\
\toprule 
\multirow{2}{*}{\textbf{Evaluator}} 
    & \multirow{2}{*}{\textbf{Test}} 
            & \multicolumn{4}{c}{\textbf{Consistency}} 
            & \multicolumn{4}{c}{\textbf{Coherence}} 
            & \multicolumn{4}{c}{\textbf{Fluency}} 
            & \multicolumn{4}{c}{\textbf{Relevance}} \\
\cmidrule(lr){3-6} \cmidrule(lr){7-10} \cmidrule(lr){11-14} \cmidrule(lr){15-18}
    &       & LM Stat. & p-value & F Stat. & p-value 
            & LM Stat. & p-value & F Stat. & p-value 
            & LM Stat. & p-value & F Stat. & p-value 
            & LM Stat. & p-value & F Stat. & p-value  \\
\midrule
\multirow{2}{*}{GPT-4o mini}
    & BP    & 372.121 & 3.02e-78 &  96.615 & 4.52e-89 
            & 147.034 & 5.71e-30 &  32.261 & 2.07e-31 
            & 144.954 & 1.58e-29 &  31.759 & 6.35e-31 
            & 102.860 & 1.32e-20 &  21.903 & 2.85e-21 \\
    & White & 446.359 & 4.68e-82 &  30.547 & 2.61e-97 
            & 204.285 & 1.60e-32 &  11.556 & 4.96e-35 
            & 187.021 & 4.08e-29 &  10.450 & 3.71e-31 
            & 132.282 & 1.45e-18 &   7.116 & 1.85e-19 \\
\midrule
\multirow{2}{*}{DSR1-Qwen-32B}
    & BP    & 332.234 & 1.17e-69 &  83.545 & 4.44e-78 
            &  64.602 & 1.36e-12 &  13.414 & 7.81e-13 
            & 209.266 & 2.95e-43 &  47.970 & 2.46e-46 
            &  78.494 & 1.73e-15 &  16.447 & 7.41e-16 \\
    & White & 406.728 & 8.21e-74 &  26.910 & 4.32e-86 
            & 142.666 & 1.58e-20 &   7.729 & 1.33e-21 
            & 242.606 & 3.52e-40 &  14.111 & 6.53e-44 
            &  92.448 & 2.76e-11 &   4.841 & 1.19e-11 \\
\midrule
\multirow{2}{*}{Qwen2.5-72B}
    & BP    & 351.775 & 7.26e-74 &  89.844 & 2.03e-83 
            &  82.248 & 2.84e-16 &  17.276 & 1.11e-16 
            & 227.917 & 2.99e-47 &  52.956 & 5.94e-51 
            &  83.830 & 1.32e-16 &  17.627 & 4.96e-17 \\
    & White & 407.695 & 5.17e-74 &  26.996 & 2.33e-86 
            & 142.134 & 1.99e-20 &   7.697 & 1.71e-21 
            & 245.423 & 9.55e-41 &  14.304 & 1.40e-44 
            & 100.688 & 9.49e-13 &   5.302 & 3.34e-13 \\
\bottomrule

\multicolumn{18}{c}{\textbf{DialSumm by G-Eval}}\\
\toprule 
\multirow{2}{*}{\textbf{Evaluator}} 
    & \multirow{2}{*}{\textbf{Test}} 
            & \multicolumn{4}{c}{\textbf{Consistency}} 
            & \multicolumn{4}{c}{\textbf{Coherence}} 
            & \multicolumn{4}{c}{\textbf{Fluency}} 
            & \multicolumn{4}{c}{\textbf{Relevance}} \\
\cmidrule(lr){3-6} \cmidrule(lr){7-10} \cmidrule(lr){11-14} \cmidrule(lr){15-18}
    &       & LM Stat. & p-value & F Stat. & p-value 
            & LM Stat. & p-value & F Stat. & p-value 
            & LM Stat. & p-value & F Stat. & p-value 
            & LM Stat. & p-value & F Stat. & p-value  \\
\midrule
\multirow{2}{*}{GPT-4o-mini}
    & BP    &  70.220 & 9.22e-14 & 14.723 & 4.30e-14 
            & 199.050 & 4.54e-41 & 46.209 & 2.85e-44 
            & 250.633 & 4.02e-52 & 60.796 & 2.06e-57 
            &  87.825 & 1.92e-17 & 18.664 & 5.49e-18 \\
    & White &  96.250 & 5.87e-12 &  5.091 & 2.01e-12 
            & 238.533 & 2.32e-39 & 14.160 & 1.56e-43 
            & 271.824 & 4.40e-46 & 16.613 & 9.80e-52 
            & 170.231 & 7.82e-26 &  9.548 & 9.86e-28 \\
\midrule
\multirow{2}{*}{DSR1-Qwen-32B}
    & BP    & 100.158 & 4.90e-20 & 21.483 & 9.27e-21 
            & 126.174 & 1.54e-25 & 27.616 & 9.87e-27 
            & 169.680 & 8.54e-35 & 38.451 & 4.63e-37 
            & 177.728 & 1.64e-36 & 40.540 & 5.12e-39 \\
    & White & 169.039 & 1.33e-25 &  9.468 & 1.83e-27 
            & 196.532 & 5.45e-31 & 11.260 & 1.21e-33 
            & 225.735 & 8.54e-37 & 13.255 & 1.83e-40 
            & 250.758 & 8.03e-42 & 15.045 & 1.65e-46 \\
\midrule
\multirow{2}{*}{Qwen2.5-72B}
    & BP    &  88.782 & 1.21e-17 & 18.877 & 3.37e-18 
            & 209.551 & 2.57e-43 & 49.076 & 6.76e-47 
            & 199.737 & 3.23e-41 & 46.395 & 1.92e-44 
            & 125.827 & 1.83e-25 & 27.532 & 1.19e-26 \\
    & White & 123.892 & 5.40e-17 &  6.694 & 7.13e-18 
            & 228.974 & 1.92e-37 & 13.482 & 3.09e-41 
            & 235.628 & 8.89e-39 & 13.953 & 7.83e-43 
            & 175.737 & 6.61e-27 &  9.897 & 6.01e-29 \\
\bottomrule

\multicolumn{18}{c}{\textbf{ROSCOE by G-Eval}} \\
\toprule 
\multirow{2}{*}{\textbf{Evaluator}} 
    & \multirow{2}{*}{\textbf{Test}} 
            & \multicolumn{4}{c}{\textbf{CosmosQA}} 
            & \multicolumn{4}{c}{\textbf{DROP}} 
            & \multicolumn{4}{c}{\textbf{e-SNLI}} 
            & \multicolumn{4}{c}{\textbf{GSM8k}} \\
\cmidrule(lr){3-6} \cmidrule(lr){7-10} \cmidrule(lr){11-14} \cmidrule(lr){15-18} 
    &       & LM Stat. & p-value & F Stat. & p-value 
            & LM Stat. & p-value & F Stat. & p-value 
            & LM Stat. & p-value & F Stat. & p-value 
            & LM Stat. & p-value & F Stat. & p-value  \\
\midrule
\multirow{2}{*}{GPT-4o-mini}
    & BP    &  5.839 & 0.3222 & 1.167 & 0.3270 
            & 11.334 & 0.0451 & 2.328 & 0.0440 
            & 26.074 & 0.0001 & 6.053 & 0.0000 
            &  7.586 & 0.1806 & 1.530 & 0.1822 \\
    & White & 17.194 & 0.6404 & 0.841 & 0.6609 
            & 23.456 & 0.2669 & 1.188 & 0.2681 
            & 35.174 & 0.0192 & 1.974 & 0.0124 
            & 26.151 & 0.1609 & 1.346 & 0.1556 \\
\midrule
\multirow{2}{*}{DSR1-Qwen-32B}
    & BP    &  8.042 & 0.1539 & 1.626 & 0.1550 
            & 20.313 & 0.0011 & 4.369 & 0.0008 
            & 24.209 & 0.0002 & 5.537 & 0.0001 
            & 15.828 & 0.0074 & 3.335 & 0.0065 \\
    & White & 17.670 & 0.6092 & 0.867 & 0.6290 
            & 40.833 & 0.0039 & 2.281 & 0.0022 
            & 58.598 & 0.0000 & 4.122 & 0.0000 
            & 33.872 & 0.0270 & 1.825 & 0.0210 \\
\midrule 
\multirow{2}{*}{Qwen2.5-72B}
    & BP    &  7.883 & 0.1628 & 1.592 & 0.1641 
            & 22.042 & 0.0005 & 4.785 & 0.0004 
            & 22.554 & 0.0004 & 5.092 & 0.0002 
            & 27.782 & 0.0000 & 6.259 & 0.0000 \\
    & White & 25.904 & 0.1690 & 1.333 & 0.1640 
            & 31.326 & 0.0510 & 1.657 & 0.0438 
            & 49.770 & 0.0002 & 3.196 & 0.0000 
            & 56.739 & 0.0000 & 3.545 & 0.0000 \\
\bottomrule 

\multicolumn{18}{c}{\textbf{ROSCOE by SocREval}}\\
\toprule 
\multirow{2}{*}{\textbf{Evaluator}} 
    & \multirow{2}{*}{\textbf{Test}} 
            & \multicolumn{4}{c}{\textbf{CosmosQA}} 
            & \multicolumn{4}{c}{\textbf{DROP}} 
            & \multicolumn{4}{c}{\textbf{e-SNLI}} 
            & \multicolumn{4}{c}{\textbf{GSM8k}} \\
\cmidrule(lr){3-6} \cmidrule(lr){7-10} \cmidrule(lr){11-14} \cmidrule(lr){15-18}
    &       & LM Stat. & p-value & F Stat. & p-value 
            & LM Stat. & p-value & F Stat. & p-value 
            & LM Stat. & p-value & F Stat. & p-value 
            & LM Stat. & p-value & F Stat. & p-value  \\
\midrule
\multirow{2}{*}{GPT-4o-mini}
    & BP    &  7.256 & 0.20231 & 1.461 & 0.20457 
            &  4.016 & 0.54705 & 0.796 & 0.55399 
            &  7.637 & 0.17742 & 1.545 & 0.17954 
            &  3.577 & 0.61180 & 0.707 & 0.61918 \\
    & White & 20.130 & 0.26762 & 1.199 & 0.26973 
            & 37.301 & 0.01077 & 2.041 & 0.00732 
            & 12.766 & 0.75172 & 0.722 & 0.77546 
            & 11.661 & 0.82022 & 0.663 & 0.83626 \\
\midrule
\multirow{2}{*}{DSR1-Qwen-32B}
    & BP    &  5.659 & 0.34085 & 1.130 & 0.34606 
            & 29.404 & 0.00002 & 6.643 & 0.00001 
            &  6.994 & 0.22105 & 1.409 & 0.22457 
            &  4.244 & 0.51487 & 0.841 & 0.52199 \\
    & White & 13.283 & 0.86492 & 0.636 & 0.88162 
            & 38.105 & 0.00860 & 2.095 & 0.00561 
            & 18.955 & 0.52477 & 0.933 & 0.54694 
            &  8.983 & 0.98311 & 0.421 & 0.98678 \\
\midrule 
\multirow{2}{*}{Qwen2.5-72B}
    & BP    & 13.470 & 0.01935 & 2.805 & 0.01810 
            &  8.464 & 0.13245 & 1.714 & 0.13293 
            & 16.545 & 0.00545 & 3.569 & 0.00450 
            &  2.321 & 0.80313 & 0.456 & 0.80886 \\
    & White & 34.356 & 0.00755 & 2.227 & 0.00499 
            & 22.780 & 0.19917 & 1.291 & 0.19706 
            & 29.926 & 0.03818 & 1.813 & 0.02981 
            & 12.387 & 0.77613 & 0.707 & 0.79345 \\
\bottomrule 
\end{tabular}
}
\label{tab:het_ROSCOE}
\end{table*}

Hence, we have
\begin{equation}
    \left\{Y_{\rm test} \in \mathcal{C}\left(z_{\rm test}\right)\right\} \subseteq \left\{Y_{\rm test} \in \mathcal{C}'\left(z_{\rm test}\right)\right\}
\end{equation}
which implies
\begin{equation}
    \mathbb{P}\left(Y_{\rm test} \in \mathcal{C}'\left(z_{\rm test}\right)\right) \ge  \mathbb{P}\left(Y_{\rm test} \in \mathcal{C}\left(z_{\rm test}\right)\right) \ge 1-\alpha.
\end{equation}
Moreover, we have 
\begin{equation}
\begin{aligned}
& \mathbb{P}\left(Y_{\text{test}} \in \mathcal{C}'\left(z_{\text{test}}\right)\right) - \mathbb{P}\left(Y_{\text{test}} \in \mathcal{C}\left(z_{\text{test}}\right)\right)  \\
&= \mathbb{P}\left(q_{1 - \alpha} \leq s'\left(z_{\rm test}, \lfloor l \rfloor \text{ or } \lceil u \rceil \right) \leq q_{1 - \alpha_0}\right) \\
&= \left(1 - \alpha_0\right) - \left(1 - \alpha\right) = \alpha - \alpha_0 > 0.
\end{aligned}
\end{equation}
Thus 
\begin{equation}
    \mathbb{P}\left(Y_{\rm test} \in \mathcal{C}'\left(z_{\rm test}\right)\right) > 1 - \alpha
\end{equation}
which completes the proof.
\end{proof}

\subsection{Hypothesis Testing of Heteroscedasticity}~\label{appendix:Heteroskedasticity_testing}
Heteroscedasticity indicates a non-constant residual variance across all observation, which could cause deviation in terms of coverage rates and interval widths for regression-based conformal prediction. It further motivates several conformal prediction methods, including CQR~\citep{romano2019conformalized}, LCP~\citep{guan2022localizedconformalpredictiongeneralized}, and R2CCP~\citep{guha2024conformal}. Here, we perform two classic tests for heteroscedasticity in our datasets: the Breusch–Pagan test and the White test.

\paragraph{Breusch–Pagan Test.}
The Breusch–Pagan (BP) test~\citep{59931fb7-1472-3118-95f0-a6dba150b325} regresses the squared ordinary least square residuals $\hat e_i^2$ on the original covariates $Z_i \in R^k$ of the $i$-th observation. Formally, it tests
\begin{equation}
    H_0: \Var\left(\varepsilon_i\right) = \sigma^2~\text{vs.}~H_1: \Var\left(\varepsilon_i\right) = \sigma^2 h\left(Z_i\right),
\end{equation}
where $h(\cdot)$ is an unknown positive-valued function to capture how the variance might change with $Z_i$. Note that we do not need to estimate $h$ since it is only used to define heteroscedasticity in alternative hypothesis. The BP test uses it to infer the simplified Langrange Multiplier (LM) test statistic as
\begin{equation}
    \mathrm{LM}_{\rm BP} = n \cdot R^2_{\hat e^2 \sim Z}~\dot\sim~\chi^2_{k} ,\quad m = \dim(Z_i)
\end{equation}
from the auxiliary regression \(\hat{e}^2 = \delta0 + Z_i'\delta+v_i\), where $n$ is the sample size. A small $p$‐value indicates rejection of homoscedasticity.

\paragraph{White Test.}
The White test~\citep{f457a0f7-5c1e-3d02-9161-7806a34faad9} extends BP by including not only $Z$ but also their squares and cross products, so the variance function \emph{h(·)} is not required. Let $Z_i=(Z_{i1}, \ldots, Z_{ik})'$ and define 
\begin{equation}
    V_i = (Z_i', Z_{i1}^2, \ldots,  Z_{ik}^2,  Z_{i1}Z_{i2}, \ldots, Z_{i\left(k - 1\right)}Z_{ik})'.
\end{equation}
Then under null hypotheses,
\begin{equation}
    \mathrm{LM}_{\rm White} = n \cdot R^2_{\hat e^2 \sim V}~\dot\sim~\chi^2_{m},\quad m = \dim\left(V_i\right).
\end{equation}

\paragraph{Results.} We report $p$-values for both BP Test and White Test in Table~\ref{tab:het_ROSCOE}. From the table, we have the following observations: First, for SummEval and DialSumm by G-Eval, all four metrics and all models exhibit highly significant heteroskedasticity (\(p<10^{-12}\)); Second, regarding ROSCOE by G-Eval, CosmosQA remains homoscedastic, whereas DROP, e-SNLI and GSM8k show $p < 0.05$; Third, regarding ROSCOE by SocREval, heteroscedasticity is confined to DROP for DSR1-Qwen-32B and GPT-4o mini and to CosmosQA and e-SNLI for Qwen2.5-72B.

\subsection{Compared Conformal Prediction Methods}~\label{appendix:CP Methods}
We test a total of seven regression-based conformal prediction (CP) methods to generate continuous prediction intervals (e.g., $[3.2, 4.1]$), as well as two ordinal classification-based CP methods to produce ordered discrete intervals (e.g., $[3, 4]$). Here we provide a detailed discussion of these approaches, including the motivation behind our choice to focus on regression and ordinal formulations rather than methods based on risk control. We further elaborate on how each method computes non-conformity scores and constructs predictive intervals accordingly.

\paragraph{Why Not Classification-based Methods.} Prior works primarily apply conformal prediction to classification-style tasks, which produces non-ordered prediction set, e.g. $\left\{A, C\right\}$ in multiple choice question answering. Admittedly, the rating scale $\left\{1, 2, 3, 4, 5\right\}$ can be cast as a multiple-choice problem. However, it is hard to interpret a prediction set such as $\left\{1, 5\right\}$ (i.e., a set with both the lowest and highest scores to be both plausible but nothing in between). Moreover, \citet{wang2025improvingllmasajudgeinferencejudgment} show that the judgment distributions from LLMs can be irregular or even bimodal, making such fragmented prediction sets not only difficult to interpret, but also problematic for downstream decision-making. In contrast, regression-based and ordinal conformal predictors generate ordered prediction intervals, offering a more coherent and interpretable descriptions of score variability. These intervals not only shows inclusion, but also a range of plausible scores, which could be crucial in high-stakes applications such as medical diagnosis. For example, if an LLM judge assigns a rating of $3$ (i.e., ``neutral''), a prediction set like $\{1, 5\}$ could be confusing as it shows the condition can be either very mild or very severe. However, a prediction interval such as $\left[3, 5\right]$ would be more intuitive to medical experts and patients.

\paragraph{Continuous Conformal Prediction Methods.} We present a brief description of each continuous conformal prediction method used in our experiments, including the definition of non-conformity score, procedure to construct prediction interval, and hyperparameter settings.

\begin{itemize}[
    align=left,
    leftmargin=1em,
    itemindent=0pt,
    labelsep=0pt,
    labelwidth=1em,
]
\item \textit{Conformalized Quantile Regression (CQR)~\citep{romano2019conformalized}.} CQR calibrates the prediction interval using a conditional quantile regression instead of conditional mean regression in naive conformal prediction. The non-conformity scores of the $i$-th calibration data in CQR is defined as 
\begin{equation}
    s_i = \max\left\{\hat{q}_{\alpha / 2}\left(z_i\right) - y_i,~y_i - \hat{q}_{1 - \alpha / 2}\left(z_i\right)\right\}
\end{equation}
where $\hat{q}_\tau$ is the $\tau$–quantile regression estimator. The quantile regression estimator is obtained by optimizing the pinball loss. To construct the prediction interval, we compute $s_i, \forall~i$ on the calibration set and let $Q_{1 - \alpha}$ be the $\left(1 - \alpha\right)$-quantile of $\left\{s_i | \forall~i \in \textrm{calibration~set}\right\}$. Then for any test data $z_{\rm test}$, we obtain its prediction interval by
\begin{equation}
    \left[\hat{q}_{\alpha / 2}\left(z_{test}\right) - Q_{1 - \alpha},~\hat{q}_{1 - \alpha / 2}\left(z_{test}\right) + Q_{1 - \alpha}\right].
\end{equation} 
In our experiments, we implement CQR using the \texttt{MapieQuantileRegressor} from the \texttt{mapie} package~\citep{taquet2022mapieopensourcelibrarydistributionfree}. We use gradient boosting regressor as the base quantile regression estimator.

\item \textit{Asymmetric CQR (CQR)~\citep{Sesia2019ACO}.} It is an asymmetric variant of CQR. In asymmetric CQR, we define two non-conformity scores $s_i^{\ell}$ and $s_i^{u}$ for the $i$-th data in the calibration set. These two scores are defined as  
\begin{equation}
    s_i^{\ell} = \hat{q}_{\alpha}\left(z_i\right) - y_i,
    \quad
    s_i^{u} = y_i - \hat{q}_{1-\alpha}\left(z_i\right)
\end{equation}
Then, let $Q_{\ell}$ and $Q_{u}$ be the $\left(1 - \alpha\right)$-quantiles of $\left\{s_i^{\ell} | \forall~i \in \textrm{calibration~set}\right\}$ and $\left\{s_i^{u} | \forall~i \in \textrm{calibration~set}\right\}$, respectively. we construct the interval by
\begin{equation}
    \left[\hat{q}_{\alpha}\left(z_{\rm test}\right) - Q_{\ell},~\hat{q}_{1-\alpha}\left(z_{\rm test}\right) + Q_{u}\right]
\end{equation}
In our experiments, we also implement asymmetric CQR using the \texttt{MapieQuantileRegressor} from the \texttt{mapie} package with the same quantile regressor as CQR.

\item \textit{Conditional Histogram Regression (CHR)~\citep{sesia2021conformalpredictionusingconditional}.} CHR aims to predict a shortest interval that expected to meet the confidence level by estimating a histogram of conditional probability. It partitions the range of label into $J$ bins and try to merge them into a valid interval, which is similar to ordinal predictors. To be specific, it firstly runs a model (e.g., quantile regression) to estimate the conditional probability $\mathbb{P}(Y \in \text{bin}_j \mid Z=z),~\forall~j$ on each bin to form a probability histogram. Then for any given confidence level $1 - \alpha_t = \frac{t}{T}$ for $t \in \{0,1,...,T\}$, it constructs the shortest continuous interval $\mathcal{C}_t$ containing $J_t$ bins and satisfying 
\begin{equation}
\Sigma_{j=1}^{J_t} \mathbb{P}(y_i\in \text{bin}_j \mid z_i) \geq 1 - \alpha_t,
\end{equation} 
where $t$ is an index representing the confidence level. For each confidence level $1 - \alpha_t$, CHR constructs a series of nested intervals $\{\mathcal{C}_t\}^T_{t = 0}$, where $\mathcal{C}_0 \subseteq \mathcal{C}_1 \subseteq ... \subseteq \mathcal{C}_T$ and $\mathcal{C}_T$ is the interval that cover the real label with 100\% confidence. Then for each data point $z_i$ we can calculate a score as
\begin{equation}
    s_i = \min \{t: y_i \in \mathcal{C}_{t}(z_i)\},
\end{equation}
which is the smallest index to contain $y_i$. Then a quantile $\hat{q}$ is estimated to construct the interval for the test point $z_{\rm test}$ as 
\begin{equation}
    \mathcal{C}(z_{\rm test}) = \mathcal{C}_{\hat{q}}(z_{\rm test}) \in \{\mathcal{C}_t(z_{\rm test})\}^T_{t=0}.
\end{equation}

\item \textit{Locally Valid and Discriminative (LVD)~\citep{NEURIPS2021_46c7cb50}.}
LVD follows the naive split conformal prediction with the same non-conformity score calculation (Equation~\ref{equation:absolute_difference}) and interval construction (Equation~\ref{equation:naive_cp}). It then uses similarity to determine $\hat{q}$. Specifically, it first compute the similarity $K_f\left(z_i, z_{\rm test}\right)$ between the test point $z_{\rm test}$ and any $i$-th calibration point $z_i$ using a learned kernel function $K_f$, and then normalizes the similarities into a weight
\begin{equation}
    w_i(z_{\rm test}) \propto K_f\left(z_i, z_{\rm test}\right).
\end{equation}
Then points closer to $z_{\rm test}$ would have higher weights to estimate the quantile for a test point $z_{\rm test}$. Mathematically, the quantile for a test point $z_{\rm test}$ can be estimated by
\begin{equation}
    \hat{q}(z_{\rm test}) = \inf \{s\geq 0: \Sigma_{i=1}^n w_i \mathds{1}\{s_i\leq s\} \geq 1-\alpha \}
\end{equation}
Finally, the prediction interval is given by:
\begin{equation}
    \left[\hat{y}_{\rm test} - \hat{q}(z_{\rm test}),~\hat{y}_{\rm test} + \hat{q}(z_{\rm test})\right].
\end{equation}

\item \textit{Boosted Conformal Prediction~\citep{xie2024boostedconformalpredictionintervals}.} Boosted conformal prediction is a post-hoc method that applies gradient boosting on top of classic conformal prediction methods such as CQR~\citep{romano2019conformalized} and  LCP~\citep{guan2022localizedconformalpredictiongeneralized}. The boosted non-conformity score for $i$-th calibration data $z_i$ is defined as 
\begin{equation}
s_i^T = s_i^0 - g_T(z_i),
\end{equation}
where $s_i^0$ is the baseline conformity score, and $g_T(z_i)$ is a correction that given by a gradient boosting model after $T$ rounds. The loss function is designed to minimize interval width while guarantee the coverage. Following this paradigm, we analyze two variants: Boosted CQR which uses LCP as the conformal prediction method and Boosted LCP which uses LCP as the conformal prediction method.

\item \textit{R2CCP (Regression-to-Classification Conformal Prediction)~\citep{guha2024conformal}.} R2CCP is similar to CHR that splits the whole prediction range ($[1, 5]$ in Likert scale) into bins and calculate conditional distribution $\mathbb{P}(Y \in \text{bin}_j \mid Z = z), \forall~j$. But it predicts the bins that the true label might falls into by a softmax‐output neural network and combines the predicted bins into an interval. \\To be specific, it predicts the probability of each bin and runs a linear interpolation between adjacent bin midpoints. Thus, for each calibration point $z_i$, we have a probability density function $\hat{f}_{z_i}(·)$ on the prediction range. Then we calculate the a score 
\begin{equation}
    s_i = \hat{f}_{z_i}(y_i),
\end{equation}
which is the probability density at the real label $y_i$, on the calibration set. Then a $\hat{q}$ is calculated to infer a prediction set 
\begin{equation}
    \mathcal{C}(z_{\rm test}) = \{a: \hat{f}_{z_{\rm test}}\left(a\right) \geq \hat{q}\}
\end{equation} 
for a test point $z_{\rm test}$, which implies the coverage guarantee. \\ However, such a set may consist of multiple disjoint intervals. For the sake of explanability and to ensure fairness when comparing with other methods, we merge the disjoint intervals into a single one. For example, 
\{[1, 1.35], [2.02, 3.1]\} is merged into [1,3.1]. This procedure often leads to wider prediction intervals for R2CCP. Nevertheless, our experiments demonstrate that it remains the best predictor in terms of coverage, interval width, and computational cost.
\end{itemize}

\paragraph{Ordinal Conformal Prediction Methods.}
Ordinal conformal prediction methods is naturally suited for rating-based tasks. It generate intervals using the probabilities of ordinal labels (e.g., different ratings), which can be derived from judgment logits after softmax. We present two ordinal conformal prediction methods below.

\begin{itemize}
\item \textit{Ordinal Adaptiive Prediction Set (OrdinalAPS)~\citep{lu2022improvingtrustworthinessaidisease}.} Ordinal APS is the ordinal version of Adaptive Prediction Set (APS)~\citep{romano2020classificationvalidadaptivecoverage, angelopoulos2022uncertaintysetsimageclassifiers}. Its non-conformity score equals to $1$ if the true label falls within in the interval and $0$ otherwise 
\begin{equation}
    s_i = \mathds{1}_{y_i\in \mathcal{C}(z_i)}.
\end{equation} 
The empirical quantile $\hat{q}$ is defined as $\frac{\lceil \left(n+1\right)\left(1 - \alpha\right)\rceil}{n}$-quantile of non-conformity scores in calibration set. Then for a test point $z_{\rm test}$, OrdinalAPS begins from the label with highest probability and then extend the prediction set in both directions until the accumulated probability mass reaches $\hat{q}$. Then the prediction interval can be transformed from the prediction set by solving the following problem
\begin{equation}
    \underset{\left(l, u\right) \in \mathcal{Y}^2,~l \leq u}{\rm argmin} \left\{u-l: \Sigma_{j=l}^u \hat{f}(j | z_{\rm test}) \geq \hat{q}\right\}
\end{equation}
where $l$ and $u$ denote the lower bound and upper bound, respectively, and $\hat{f}(j)$ is the probability of $j$ in classification.

\item \textit{Ordinal Risk Control (OrdinalRC)~\citep{xu2024conformalriskcontrolordinal}.} Ordinal Risk Control is similar to OrdinalAPS that they both start from the label with highest probability and then expand the set in both directions until a threshold is met. Specifically, OrdinalRC designs two types of risk: weight-based risk and divergence-based risk. In our analysis, we select the weight-based risk since predictions with another risk often only output a single score. Regarding interval construction with weight-based risk, similar to OrdinalAPS, $C(z_{\rm test})$ can be obtained by solving the following optimization problem
\begin{equation}
    \underset{\left(l, u\right) \in \mathcal{Y}^2, l \leq u}{\rm argmin} \left\{u - l: \Sigma_{j = l}^u h\left(j\right)\hat{f}(j | z_{\rm test}) \geq \hat{q}\right\},
\end{equation}
where $h(j)$ is a weight specifically assigned to label $j$ and $\hat{q}$ here is a weight-adjusted counterpart compared to the quantile in OrdinalAPS.
\end{itemize}

\begin{table*}[ht]
\centering
\caption{Detailed hyperparameter settings for each conformal prediction method.}
\begin{tabular}{l>{\raggedright\arraybackslash}p{12cm}}
\hline
\textbf{Method} 
    & \textbf{Hyperparameter} \\ \hline
CQR 
    & MAPIE, gradient boosting regressor with quantile loss \\
Asymmetric CQR 
    & Same as CQR \\
CHR 
    & QNet estimator, batch\_size=32, hidden\_dim=256, lr=5e-4, epoch=1000 \\
LVD 
    & DNN\_model, readout\_layer = pretrain\_general(seed=0, quiet=True, model\_setting=0), kernel\_model = KernelMLKR(d=10, seed=0, n\_iters=500, norm=True, lr=1e-3) \\
Boosted LCP 
    & len\_local\_boost: n\_rounds\_cv=500, learning\_rate=0.02, store=True, verbose=False \\
Boosted CQR 
    & len\_cqr\_boost: same with Boosted LCP \\
R2CCP 
    & Default setting in original implementation but max\_epochs=100 \\
OrdinalAPS 
    & Default setting in original implementation \\
OrdinalRC 
    & Default setting in original implementation with WeightedCRPredictor \\ \hline
\end{tabular}
\label{table:hyperparameters}
\end{table*}

\begin{table*}[ht!]
\centering
\caption{Running time and memory cost of each conformal prediction method.}
\begin{tabular}{lcccc}
\hline
\textbf{Method} & \textbf{Time Mean (s)} & \textbf{Time Std. (s)} & \textbf{Memory Mean (MB)} & \textbf{Memory Std. (MB)} \\ \hline
CQR             &  0.83 & 0.03 &  0.35 & 0.00 \\
Asymmetric CQR  &  0.82 & 0.03 &  0.35 & 0.00 \\
CHR             &  9.54 & 0.25 &  0.62 & 0.01 \\
LVD             & 93.67 & 2.82 &  0.55 & 0.01 \\
Boosted CQR     & 91.43 & 3.24 &  0.89 & 0.05 \\
Boosted LCP     & 87.27 & 2.57 &  2.05 & 0.01 \\
R2CCP           &  9.25 & 0.53 &  1.35 & 0.00 \\
OrdinalAPS      &  0.01 & 0.00 &  0.20 & 0.00 \\
OrdinalRC       &  0.03 & 0.00 &  0.19 & 0.00 \\ \hline
\end{tabular}
\label{table:runtime_comparison}
\end{table*}

A detailed hyperparameter settings for each aforementioned conformal prediction methods are listed in Table~\ref{table:hyperparameters}.

\subsection{Running Time and Memory Cost} 
Table~\ref{table:runtime_comparison} shows the runtime and memory analysis of each method. Boosted CQR and Boosted LCP are having higher cost potentially due to boosting. For LVD, the main reason for higher cost might be due to computing the kernel matrix for hundreds of iterations to quantify pairwise similarities.

\subsection{LLM Judge Selection and Prompt Sensitivity}~\label{sec:prompt_sense}
In our experiments, we choose GPT-4o mini as a budget-friendly evaluator. In Table~\ref{tab:gpt4ocomparewith4omini}, we compare GPT-4o mini vs. its larger variant GPT-4o on five rephrased G-Eval chain-of-thought (CoT) prompts. We use R2CCP to construct prediction interval without boundary adjustment. Our preliminary results show that they tend to produce prediction interval with similar interval width and coverage rate. We can also observe that interval width and coverage are similar across all rephrases, which shows that GPT-4o mini, though a budget-friendly model, is insensitive to prompt variants.

Regarding the selection of Qwen2.5-72B-Instruct and DeepSeek-R1-Distill-Qwen-32B, we choose them because they are all widely used open-source models with strong performance and could also fit with our GPU limitations. Besides, we specifically choose DeepSeek-R1-Distill-Qwen-32B to investigate how reasoning models would impact the LLM-based evaluation and interval quality in both summarization and reasoning tasks.

\begin{table*}[ht]
\centering
\caption{GPT-4o vs. GPT-4o mini as LLM judge with five rephrased CoT prompts over 30 different runs. }
\resizebox{\textwidth}{!}{%
\begin{tabular}{lcccc}
\toprule
\textbf{Prompt} 
    & \textbf{Consistency} 
    & \textbf{Coherence} 
    & \textbf{Fluency} 
    & \textbf{Relevance} \\ 
\midrule
\multicolumn{5}{c}{\textbf{GPT-4o}} \\[2pt]
\midrule
CoT 0 
    & $0.84 \pm 0.22 / 90.06\% \pm 2.03\%$ 
    & $3.09 \pm 0.15 / 89.61\% \pm 2.92\%$ 
    & $1.09 \pm 0.22 / 89.94\% \pm 2.33\%$ 
    & $2.41 \pm 0.15 / 89.52\% \pm 2.94\%$ \\ 
CoT 1 
    & $0.80 \pm 0.25 / 89.59\% \pm 2.06\%$ 
    & $2.98 \pm 0.16 / 88.56\% \pm 2.89\%$ 
    & $1.05 \pm 0.23 / 89.95\% \pm 2.00\%$ 
    & $2.37 \pm 0.11 / 89.19\% \pm 2.37\%$ \\ 
CoT 2 
    & $0.87 \pm 0.29 / 89.75\% \pm 2.11\%$ 
    & $2.98 \pm 0.14 / 89.35\% \pm 2.81\%$ 
    & $1.07 \pm 0.23 / 90.00\% \pm 2.24\%$ 
    & $2.40 \pm 0.16 / 89.34\% \pm 2.99\%$ \\ 
CoT 3 
    & $0.88 \pm 0.34 / 90.05\% \pm 2.13\%$ 
    & $3.02 \pm 0.13 / 89.53\% \pm 2.31\%$ 
    & $1.05 \pm 0.27 / 89.95\% \pm 2.27\%$ 
    & $2.38 \pm 0.14 / 89.65\% \pm 2.81\%$ \\ 
CoT 4 
    & $0.81 \pm 0.27 / 89.74\% \pm 2.26\%$ 
    & $3.05 \pm 0.13 / 89.65\% \pm 2.65\%$ 
    & $1.05 \pm 0.24 / 89.80\% \pm 2.10\%$ 
    & $2.39 \pm 0.15 / 89.87\% \pm 3.20\%$ \\ \midrule
\multicolumn{5}{c}{\textbf{GPT-4o mini}} \\ [2pt]
\midrule
CoT 0 
    & $0.83 \pm 0.26 / 89.65\% \pm 2.44\%$ 
    & $2.97 \pm 0.14 / 88.79\% \pm 2.65\%$ 
    & $1.06 \pm 0.24 / 89.62\% \pm 2.09\%$ 
    & $2.36 \pm 0.13 / 89.76\% \pm 2.48\%$ \\ 
CoT 1 
    & $0.81 \pm 0.21 / 89.70\% \pm 2.05\%$ 
    & $3.02 \pm 0.16 / 89.30\% \pm 2.95\%$ 
    & $1.07 \pm 0.27 / 89.87\% \pm 2.26\%$ 
    & $2.37 \pm 0.14 / 89.55\% \pm 2.86\%$ \\ 
CoT 2 
    & $0.81 \pm 0.24 / 89.66\% \pm 2.25\%$ 
    & $2.98 \pm 0.12 / 89.13\% \pm 2.55\%$ 
    & $1.05 \pm 0.26 / 89.62\% \pm 2.07\%$ 
    & $2.39 \pm 0.13 / 89.65\% \pm 2.71\%$ \\ 
CoT 3 
    & $0.88 \pm 0.31 / 90.09\% \pm 2.18\%$ 
    & $3.03 \pm 0.13 / 89.46\% \pm 2.69\%$ 
    & $1.07 \pm 0.26 / 89.85\% \pm 1.89\%$ 
    & $2.42 \pm 0.15 / 89.83\% \pm 2.67\%$ \\ 
CoT 4 
    & $0.85 \pm 0.28 / 89.73\% \pm 2.12\%$ 
    & $3.02 \pm 0.12 / 89.09\% \pm 2.49\%$ 
    & $1.09 \pm 0.27 / 89.77\% \pm 2.24\%$ 
    & $2.42 \pm 0.14 / 89.98\% \pm 2.62\%$ \\ 
\bottomrule
\end{tabular}
}
\label{tab:gpt4ocomparewith4omini}
\end{table*}

\begin{table*}[ht]
\centering
\caption{Impact of temperature (Tmp.) for different LLM judge over 30 runs with different random seeds. Values in parentheses are $p$-values for testing whether the coverage equals 0.9. Only DeepSeek-R1-Distill-Qwen-32B exhibits noticeable changes when temperature changes, while the other two are relatively stable.}
\resizebox{\textwidth}{!}{%
\begin{tabular}{lcccc}
\toprule
\textbf{Tmp.} & \textbf{Consistency} 
    & \textbf{Coherence} 
    & \textbf{Fluency} 
    & \textbf{Relevance} \\ 
\midrule
\multicolumn{5}{c}{\textbf{GPT-4o mini}} \\[2pt]
\midrule
0   & $0.66\pm0.17/90.58\%\pm2.47\%$ (9e-05) 
    & $2.62\pm0.16/89.28\%\pm3.03\%$ (7e-11) 
    & $0.98\pm0.16/90.24\%\pm2.10\%$ (3e-07) 
    & $2.00\pm0.11/89.65\%\pm2.17\%$ (3e-10) \\ 
1   & $0.69\pm0.19/90.88\%\pm2.49\%$ (5e-04) 
    & $2.62\pm0.15/89.63\%\pm3.12\%$ (9e-10) 
    & $0.92\pm0.16/89.36\%\pm2.71\%$ (3e-10) 
    & $1.97\pm0.12/89.70\%\pm2.50\%$ (4e-10) \\ 
\midrule
\multicolumn{5}{c}{\textbf{DeepSeek-R1-Distill-Qwen-32B}} \\[2pt]
\midrule
0   & $0.69\pm0.13/90.44\%\pm2.10\%$ (6e-05) 
    & $2.30\pm0.12/90.12\%\pm2.13\%$ (2e-08) 
    & $0.89\pm0.15/90.09\%\pm2.08\%$ (5e-04) 
    & $2.00\pm0.15/89.84\%\pm2.90\%$ (4e-08) \\ 
1   & $0.79\pm0.19/90.17\%\pm2.09\%$ (1e-06) 
    & $2.66\pm0.11/90.04\%\pm1.92\%$ (1e-06) 
    & $1.11\pm0.26/90.06\%\pm1.98\%$ (2e-05) 
    & $2.19\pm0.11/90.06\%\pm2.29\%$ (4e-09) \\ 
\midrule
\multicolumn{5}{c}{\textbf{Qwen2.5-72B-Instruct}} \\[2pt]
\midrule
0   & $0.61\pm0.13/90.73\%\pm2.02\%$ (3e-04) 
    & $2.44\pm0.14/89.54\%\pm2.48\%$ (2e-10) 
    & $0.95\pm0.12/90.18\%\pm1.92\%$ (4e-08) 
    & $1.98\pm0.12/90.45\%\pm2.50\%$ (9e-05) \\ 
1   & $0.61\pm0.18/90.29\%\pm2.09\%$ (3e-04) 
    & $2.69\pm0.12/90.56\%\pm2.16\%$ (2e-02) 
    & $0.93\pm0.13/90.65\%\pm2.29\%$ (5e-04) 
    & $2.01\pm0.11/89.91\%\pm2.73\%$ (2e-06) \\ 
\bottomrule
\end{tabular}
}
\label{tab:tmpcompare}
\end{table*}

\subsection{On Exchangeability, Randomness, and Distribution Shift of Logits}~\label{appendix:discussion_logits}
We discuss the feasibility of using logits as input for conformal prediction. 

In conformal prediction, a basic assumption is the exchangeability of data, which means that the joint distribution of test set and calibration set should remain invariant to any permutation. In our work, we assume logits from different (prompt and evaluation task) pair to satisfy exchangeability. The main intuition is that different pairs should not interfere with each other in evaluations.

There could also be randomness in LLM-as-a-judge. Temperature is commonly used to control such randomness in LLM decoding. However, we could view such randomness as adding a noise to the logits when temperature is $0$. Then \citet{zargarbashi2025samplemakeconformalprediction} show that coverage could still be theoretically guaranteed. In Table~\ref{tab:tmpcompare}, we show the interval width and coverage given by R2CCP when temperature is set to $0$ and $1$, respectively. From the table, we could see that the interval width could slightly increase when temperature is $1$, which means that more randomness might introduce higher uncertainty. Besides, we show in Table~\ref{tab:example_same_evaluation} that, even in the case where the model gives different raw score to the same prompt, the prediction intervals remain close to each other, which demonstrate the reliability of using conformal prediction to quantify uncertainty. In our main results, we set temperture to $1$ for GPT-4o mini and set temperature to $0$ for Qwen2.5-72B-Instruct and DeepSeek-R1-Distill-Qwen-32B.

Note that our paper does not focus on distribution shift, which remains an open question in conformal prediction. To better understand the impact of distribution shift, we use SummEval and DialSumm to calibrate each other, since they share common dimensions. Table~\ref{tab:distribution_shift} presents the results under such distribution shift. From the table, we observe that, even though most settings fail to achieve $90\%$ coverage, boundary adjustment could still help improve the coverage with a large margin. Besides, by digging deeper, we find that the coverage is often higher when the calibrated labels are roughly balanced or not too skewed.

\begin{table*}[ht!]
\centering
\caption{Two evaluations with the same summarization but different judge scores yield similar evaluations under our framework. In this table, Raw  is the raw score in the judgment, weighted is the weighted average based on logits, Con.Interval is the original interval predicted by R2CCP while Dis.Interval is the one after boundary adjustment.}
\resizebox{\textwidth}{!}{%
\begin{tabular}{lccccc}
\toprule
\textbf{custom\_id} & \textbf{Raw} & \textbf{Weighted} & \textbf{Logits (1–5)} & \textbf{Con.\ Interval} & \textbf{Dis.\ Interval} \\
\midrule
93\_10\_COT2 & 5 & 4.64 & $-$12.69, $-$9.06, $-$5.06, $-$1.06, $-$0.44 & [4.612, 5] & [4.67, 5] \\
93\_11\_COT2 & 4 & 4.37 & $-$11.67, $-$7.67, $-$3.67, $-$0.55, $-$0.92 & [4.626, 5] & [4.67, 5] \\
\bottomrule
\end{tabular}
}
\label{tab:example_same_evaluation}
\end{table*}

\begin{table*}[ht]
\centering
\caption{Impact of distribution shift on interval width and coverage. SummEval and DialSumm are used to calibrate each other. We mark coverage $< 85\%$ with \graytext{gray text}, coverage between 85\% -- 90\% with \underline{underline} and coverage $\geq 90\%$ with the smallest interval width in \textbf{bold}.}
\resizebox{\textwidth}{!}{%
\begin{tabular}{lcccc}
\toprule
\textbf{Model} 
    & \textbf{Fluency} 
    & \textbf{Consistency} 
    & \textbf{Coherence} 
    & \textbf{Relevance} \\
\midrule
\multicolumn{5}{c}{From SummEval to DialSumm: Before boundary adjustment} \\[2pt]
\midrule
GPT-4o mini 
    & \graytext{1.349 / 79.07\%} 
    & \graytext{0.5431 / 18.45\%} 
    & \underline{2.275 / 86.07\%} 
    & \graytext{1.912 / 84.81\%} \\
DeepSeek-R1-Distill-Qwen-32B  
    & \graytext{0.6925 / 46.57\%} 
    & \graytext{0.5956 / 24.00\%} 
    & \textbf{1.924 / 92.71\%} 
    & \graytext{1.793 / 81.86\%} \\
Qwen2.5-72B-Instruct 
    & \graytext{0.8463 / 47.43\%} 
    & \graytext{0.5864 / 20.36\%} 
    & \graytext{2.067 / 72.00\%} 
    & \underline{1.978 / 87.43\%} \\
\midrule
\multicolumn{5}{c}{From SummEval to DialSumm: After boundary adjustment} \\[2pt]
\midrule
GPT-4o mini 
    & \underline{1.352 / 86.50\%} 
    & \graytext{0.5233 / 31.26\%} 
    & \underline{2.271 / 89.57\%} 
    & \textbf{1.895 / 91.55\%} \\
DeepSeek-R1-Distill-Qwen-32B 
    & \graytext{0.6531 / 57.00\%} 
    & \graytext{0.5563 / 30.71\%} 
    & \textbf{1.922 / 94.36\%} 
    & \underline{1.773 / 87.07\%} \\
Qwen2.5-72B-Instruct 
    & \graytext{0.8376 / 62.93\%} 
    & \graytext{0.5537 / 30.64\%} 
    & \graytext{2.064 / 76.57\%} 
    & 1.980 / 90.29\% \\
\midrule
\multicolumn{5}{c}{From DialSumm to SummEval: Before boundary adjustment} \\[2pt]
\midrule
GPT-4o mini 
    & \graytext{1.316 / 53.69\%} 
    & \graytext{1.837 / 66.50\%} 
    & \graytext{1.764 / 56.50\%} 
    & \graytext{1.892 / 82.81\%} \\
DeepSeek-R1-Distill-Qwen-32B 
    & \graytext{1.121 / 35.94\%} 
    & \graytext{1.880 / 50.38\%} 
    & \graytext{1.344 / 52.38\%} 
    & \graytext{1.727 / 80.19\%} \\
Qwen2.5-72B-Instruct 
    & \graytext{1.164 / 27.19\%} 
    & \graytext{1.596 / 37.75\%} 
    & \graytext{1.491 / 48.31\%} 
    & \graytext{1.588 / 77.63\%} \\
\midrule
\multicolumn{5}{c}{From DialSumm to SummEval: After boundary adjustment} \\[2pt]
\midrule
GPT-4o mini 
    & 1.334 / 92.06\% 
    & \graytext{1.849 / 80.50\%} 
    & \graytext{1.773 / 63.25\%} 
    & 1.894 / 90.88\% \\
DeepSeek-R1-Distill-Qwen-32B 
    & \graytext{1.160 / 82.38\%} 
    & \graytext{1.878 / 69.94\%} 
    & \graytext{1.332 / 60.25\%} 
    & \underline{1.723 / 86.25\%} \\
Qwen2.5-72B-Instruct 
    & \textbf{1.219 / 90.00\%} 
    & \graytext{1.604 / 59.25\%} 
    & \graytext{1.521 / 57.38\%} 
    & \graytext{1.562 / 84.50\%} \\
\bottomrule
\end{tabular}
}
\label{tab:distribution_shift}
\end{table*}

\subsection{Human-based Baseline in Summarization Tasks}~\label{appendix:human performances}
In our experiments, we also consider a human-based baseline method. Following naive split conformal prediction (Equations~\eqref{equation:absolute_difference} and~\eqref{equation:naive_cp}), we compute the non-conformity score as
\begin{equation}
    s(y_{\rm random},y) = |y_{\rm random}-y|,
\end{equation}
where $y_{random}$ is an annotation randomly chosen from three annotations while the ground truth $y$ is their average. And then we estimate the quantile $\hat{q}$ to construct confidence interval by 
\begin{equation}
    [y_{\rm random} -\hat{q},\ y_{\rm random} +\hat{q}].
\end{equation} 
A conformal prediction method is considered better than human performance if it achieves shorter intervals with a comparable or better coverage rate. From Table~\ref{tab:baselines}, we can see that R2CCP could consistently matches or outperforms the human baselines for both SummEval and DialSumm.

\begin{table*}[ht]
\centering
\caption{Comparison of human-based baseline and R2CCP (seed = 42) on SummEval and DialSumm}
\resizebox{\textwidth}{!}{
\begin{tabular}{lccccc}
\toprule
\textbf{Evaluator} & \textbf{Method}
    & \textbf{Consistency} 
    & \textbf{Coherence} 
    & \textbf{Fluency} 
    & \textbf{Relevance} \\
\midrule
\multicolumn{6}{c}{SummEval} \\
\midrule
Human-based & Baseline 
    & 0.667 (91.4\%) 
    & 2.000 (95.6\%) 
    & 1.333 (96.3\%) 
    & 2.000 (92.8\%) \\
GPT-4o-mini & R2CCP 
    & 0.621 (90.1\%) 
    & 2.652 (89.9\%) 
    & 1.135 (93.4\%) 
    & 2.076 (91.5\%) \\
DSR1-Qwen-32B & R2CCP 
    & 0.598 (89.3\%) 
    & 2.168 (85.8\%) 
    & 0.850 (90.1\%) 
    & 2.142 (93.3\%) \\
Qwen-2.5-72B & R2CCP 
    & 0.491 (88.9\%) 
    & 2.429 (88.0\%) 
    & 0.812 (88.0\%) 
    & 1.969 (91.4\%) \\
\midrule
\multicolumn{6}{c}{DialSumm}\\
\midrule
Human-based & Baseline 
    & 2.667 (95.9\%) 
    & 2.000 (96.9\%) 
    & 2.000 (95.1\%) 
    & 2.667 (95.6\%) \\
GPT-4o-mini & R2CCP 
    & 1.799 (91.99\%) 
    & 1.701 (91.00\%) 
    & 1.215 (89.71\%) 
    & 1.580 (85.2\%) \\
DSR1-Qwen-32B & R2CCP 
    & 1.912 (88.7\%) 
    & 1.283 (89.3\%) 
    & 0.812 (88.0\%) 
    & 1.805 (89.9\%) \\
Qwen-2.5-72B & R2CCP 
    & 1.591 (87.0\%) 
    & 1.494 (90.3\%) 
    & 1.136 (90.3\%) 
    & 1.653 (91.9\%) \\
\bottomrule
\end{tabular}
}
\label{tab:baselines}
\end{table*}

\begin{table*}[ht]
\centering
\caption{Interval width and coverage on DialSumm and ROSCOE evaluated with G-Eval before boundary adjustment. We mark coverage $< 85\%$ with \graytext{gray text}, coverage between 85\% -- 90\% with \underline{underline} and coverage $\geq 90\%$ with the smallest interval width in \textbf{bold}.}
\resizebox{\textwidth}{!}{%
\begin{tabular}{lcccccccc}
\toprule
\multirow{2.5}{*}{\textbf{Method}} & \multicolumn{4}{c}{\textbf{DialSumm Evaluated with G-Eval }} & \multicolumn{4}{c}{\textbf{ROSCOE Evaluated with G-Eval}} \\
\cmidrule(lr){2-5} \cmidrule(lr){6-9}
 & \textbf{Consistency} & \textbf{Coherence} & \textbf{Fluency} & \textbf{Relevance} & \textbf{CosmosQA} & \textbf{DROP} & \textbf{e-SNLI} & \textbf{GSM8K} \\
\midrule
\multicolumn{9}{c}{\textbf{GPT-4o mini}} \\[2pt]
\midrule
CQR       & 2.41 / 91.99\% & 1.77 / 92.41\% & \textbf{1.08 / 93.38\%} & 2.06 / 91.57\% & \textbf{3.60 / 96.43\%} & \textbf{3.77 / 96.54\%} & 3.35 / 95.31\% & \textbf{3.58 / 94.83\%} \\
Asym CQR      & 2.43 / 92.30\% & 1.87 / 94.00\% & 1.18 / 94.40\% & 2.09 / 92.38\% & 3.95 / 99.56\% & 3.89 / 98.19\% & \textbf{2.98 / 96.67\%} & 3.94 / 99.27\% \\
CHR           & \graytext{1.54 / 80.01\%} & \graytext{1.48 / 83.03\%} & \graytext{0.99 / 84.01\%} & \graytext{1.40 / 80.53\%} & \graytext{2.47 / 70.78\%} & \graytext{1.82 / 68.44\%} & \graytext{1.28 / 55.66\%} & \graytext{2.27 / 70.27\%} \\
LVD           & \underline{1.90 / 89.20\%} & 1.75 / 90.67\% & \underline{1.20 / 88.40\%} & \underline{1.79 / 89.23\%} & \graytext{3.18 / 83.44\%} & \graytext{2.33 / 79.11\%} & 3.00 / 91.18\% & \graytext{3.10 / 84.50\%} \\
Boosted CQR   & \underline{1.85 / 86.81\%} & \underline{1.61 / 87.26\%} & \underline{1.03 / 86.33\%} & \underline{1.65 / 87.06\%} & \graytext{3.12 / 77.99\%} & \graytext{2.58 / 78.32\%} & \graytext{2.13 / 75.79\%} & \graytext{3.20 / 80.03\%} \\
Boosted LCP   & \underline{1.83 / 87.45\%} & \underline{1.59 / 88.30\%} & \underline{1.00 / 87.53\%} & \underline{1.76 / 87.20\%} & \graytext{3.45 / 79.66\%} & \underline{2.94 / 86.41\%} & \graytext{1.94 / 80.26\%} & \underline{3.42 / 83.53\%} \\
R2CCP         & \textbf{1.84 / 90.13\%} & \textbf{1.63 / 90.15\%} & \underline{1.14 / 89.64\%} & \textbf{1.72 / 90.11\%} & \underline{3.09 / 86.77\%} & \underline{2.54 / 86.70\%} & \underline{2.20 / 88.01\%} & \graytext{2.43 / 84.67\%} \\
\midrule
\multicolumn{9}{c}{\textbf{DeepSeek-R1-Distill-Qwen-32B}} \\[2pt]
\midrule
CQR       & 2.50 / 91.30\% & 1.51 / 91.85\% & \textbf{1.11 / 92.83\%} & 2.33 / 91.69\% & \textbf{3.62 / 96.29\%} & \textbf{3.82 / 96.22\%} & 3.33 / 97.85\% & 3.54 / 95.27\% \\
Asym CQR      & 2.52 / 91.92\% & 1.58 / 93.20\% & 1.22 / 94.04\% & 2.42 / 92.51\% & 3.89 / 98.95\% & 3.88 / 97.78\% & 2.95 / 97.15\% & 3.90 / 99.27\% \\
CHR           & \underline{1.76 / 86.09\%} & \underline{1.26 / 87.80\%} & \underline{1.06 / 87.79\%} & \underline{1.53 / 85.22\%} & \graytext{2.64 / 78.10\%} & \graytext{2.19 / 80.13\%} & \graytext{1.80 / 77.24\%} & \graytext{2.77 / 81.90\%} \\
LVD           & \textbf{2.03 / 90.19\%} & \textbf{1.41 / 90.29\%} & 1.22 / 90.41\% & \textbf{1.87 / 90.01\%} & \underline{3.31 / 89.52\%} & \underline{2.81 / 88.98\%} & \textbf{2.86 / 94.82\%} & \textbf{3.39 / 90.07\%} \\
Boosted CQR   & \underline{1.89 / 87.48\%} & \underline{1.31 / 88.61\%} & \underline{1.11 / 88.07\%} & \underline{1.71 / 87.39\%} & \graytext{3.40 / 80.92\%} & \underline{2.84 / 85.02\%} & \graytext{2.23 / 83.68\%} & \graytext{3.27 / 80.53\%} \\
Boosted LCP   & \underline{1.88 / 86.05\%} & \underline{1.32 / 86.77\%} & \underline{1.02 / 87.28\%} & \underline{1.82 / 87.24\%} & \graytext{3.49 / 81.73\%} & \underline{2.94 / 86.06\%} & \graytext{1.99 / 83.33\%} & \graytext{3.38 / 81.13\%} \\
R2CCP         & \underline{1.86 / 89.22\%} & \underline{1.31 / 89.92\%} & 1.19 / 90.57\% & \underline{1.70 / 89.39\%} & \underline{3.05 / 86.84\%} & \underline{2.44 / 85.87\%} & \underline{1.96 / 85.43\%} & \underline{2.51 / 86.77\%} \\
 \midrule
\multicolumn{9}{c}{\textbf{Qwen2.5-72B-Instruct}} \\[2pt]
\midrule
CQR       & 2.37 / 91.51\% & \textbf{1.52 / 91.57\%} & \textbf{1.06 / 93.11\%} & \textbf{2.02 / 91.55\%} & \textbf{3.62 / 96.67\%} & 3.78 / 96.86\% & 3.37 / 98.29\% & \textbf{3.58 / 95.30\%} \\
Asym CQR      & 2.40 / 91.92\% & 1.61 / 92.88\% & 1.14 / 93.87\% & 2.04 / 92.38\% & 3.93 / 99.12\% & 3.92 / 99.17\% & 2.96 / 97.41\% & 3.90 / 99.00\% \\
CHR           & \graytext{1.48 / 81.50\%} & \graytext{1.25 / 81.88\%} & \underline{1.00 / 85.97\%} & \graytext{1.33 / 81.15\%} & \graytext{2.59 / 74.97\%} & \graytext{1.76 / 68.38\%} & \graytext{1.39 / 66.84\%} & \graytext{1.76 / 72.57\%} \\
LVD           & \textbf{1.84 / 90.43\%} & \underline{1.47 / 89.91\%} & 1.20 / 90.26\% & \underline{1.74 / 89.99\%} & \underline{3.34 / 89.01\%} & \graytext{2.41 / 84.57\%} & \textbf{2.65 / 92.41\%} & \underline{2.83 / 88.50\%} \\
Boosted CQR   & \underline{1.69 / 86.57\%} & \underline{1.35 / 87.06\%} & \underline{1.05 / 87.38\%} & \underline{1.52 / 87.08\%} & \graytext{3.35 / 81.02\%} & \graytext{2.55 / 83.52\%} & \graytext{1.90 / 81.84\%} & \graytext{3.18 / 82.70\%} \\
Boosted LCP   & \underline{1.75 / 86.08\%} & \underline{1.35 / 86.29\%} & \underline{0.96 / 87.77\%} & \underline{1.63 / 86.88\%} & \graytext{3.45 / 80.41\%} & \graytext{2.79 / 83.05\%} & \graytext{1.85 / 80.79\%} & \graytext{3.42 / 83.47\%} \\
R2CCP         & \underline{1.74 / 89.97\%} & \underline{1.41 / 89.67\%} & \underline{1.14 / 89.70\%} & \underline{1.61 / 89.80\%} & \underline{3.07 / 87.11\%} & \textbf{3.10 / 93.40\%} & \graytext{1.68 / 84.80\%} & \underline{2.38 / 87.53\%} \\

\bottomrule
\end{tabular}%
}
\label{tab:dialsumm_continuous}
\end{table*}

Adding more samples leads to varying effects on correlation across dimensions. For coherence and fluency, the impact is minimal or slightly negative. In contrast, consistency and relevance benefit, particularly under the quantile method. Among evaluation methods, quantile performs best in relevance, while stratified excels in other three dimensions.

\subsection{Additional Results on Continuous Intervals Before Boundary Adjustment}
Table~\ref{tab:dialsumm_continuous} presents additional experiment results for DialSumm and ROSCOE evaluated with G-Eval before boundary adjustment. From the table, Boosted CQR and Boosted LCP consistently fail to achieve $90\%$ coverage rate on the DialSumm dataset. In contrast, R2CCP maintains coverage in the 89\%–91\% range while yielding the narrowest intervals among methods with comparable performance, thus offering an optimal trade-off between coverage and efficiency. LVD achieves slightly higher coverage but at the cost of wider intervals, making it suitable for scenarios that prioritize coverage over interval compactness. Though CQR and Asymmetric CQR guarantee 90 \% coverage, their interval widths are much larger than other methods (width $> 3$ on ROSCOE). In terms of LLM judge, DeepSeek-R1-Distill-Qwen-32B achieve slightly higher average coverage rates, and those generated by Qwen2.5-72B-Instruct are generally shorter with lower coverage.

\begin{figure*}
    \centering
    \includegraphics[width=\linewidth]{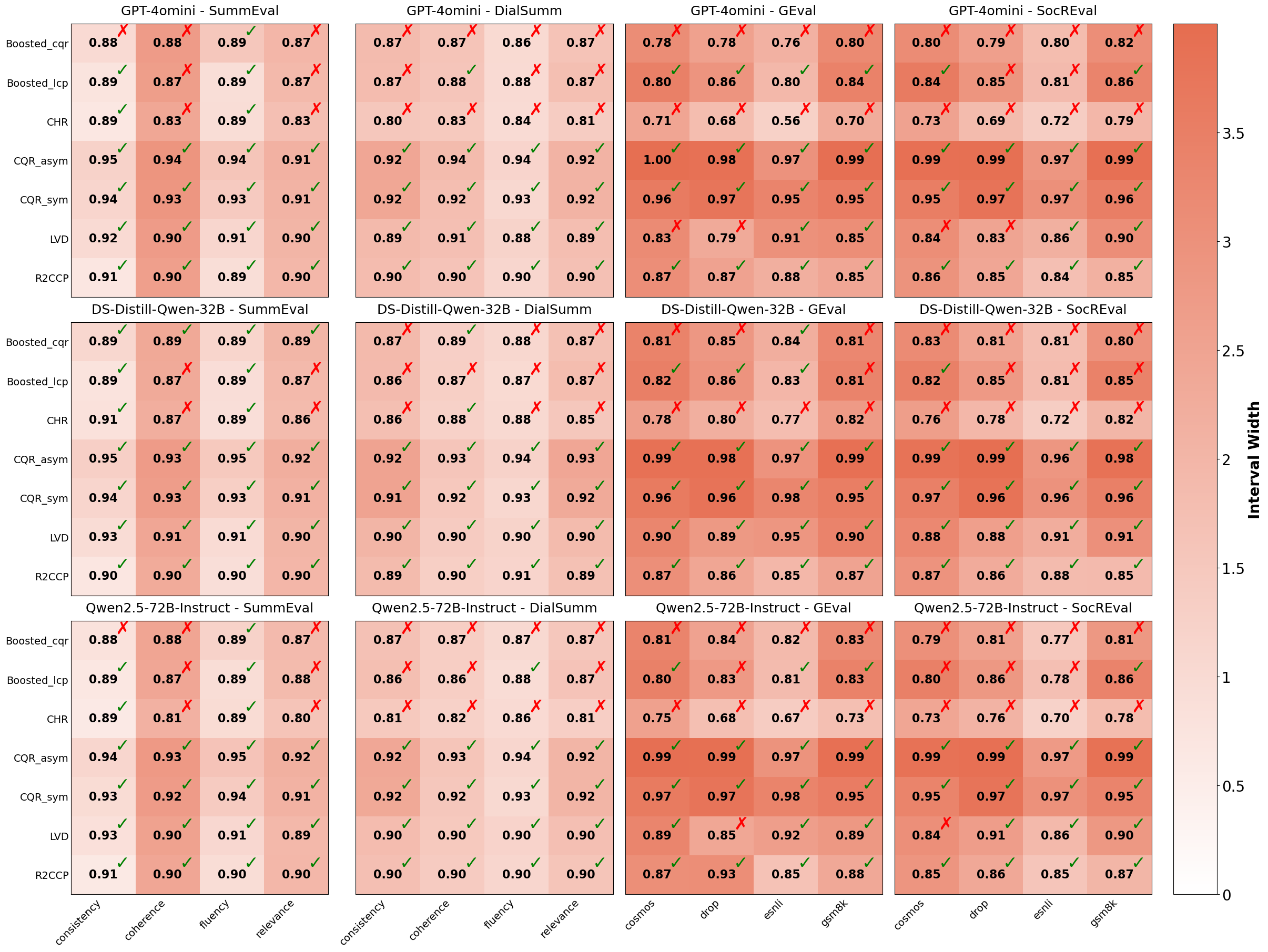}
    \caption{Summary of all experiments regarding continuous interval constructions. Each cell displays the mean coverage of the corresponding conformal prediction method over 30 runs on the given dataset. Cell shading encodes the average interval width, with lighter hues denoting narrower intervals. A \cmark~or \xmark~in the upper-right corner of each cell denotes whether the coverage criterion is met: \cmark~for met; otherwise \xmark.}
    \label{fig:summary_continuous}
\end{figure*}

\subsection{Additional Results on Discrete Intervals After Boundary Adjustment}
Table~\ref{tab:dialsumm_discrete} presents additional experiment results for DialSumm and ROSCOE evaluated with G-Eval after boundary adjustment. Compared to the results in Table~\ref{tab:dialsumm_continuous}, boundary adjustment could improve coverage without sacrificing too much or even shorten the interval width, thereby improving the trade-off between coverage and interval width. We also notice that Boosted CQR and Boosted LCP benefit the most from boundary adjustment. After boundary adjustment, prediction intervals of Boosted CQR and Boosted LCP are comparable to that of R2CCP while achieving 90\% coverage. However, it should be noted that there is no conformal prediction methods that achieve the best trade-off across different LLM judges.

\begin{table*}[ht]
\centering
\caption{Interval width and coverage on DialSumm and ROSCOE evaluated with G-Eval after boundary adjustment. We mark coverage $< 85\%$ with \graytext{gray text}, coverage between 85\% -- 90\% with \underline{underline} and coverage $\geq 90\%$ with the smallest interval width in \textbf{bold}.}
\resizebox{\textwidth}{!}{%
\begin{tabular}{lcccccccc}
\toprule
\multirow{2.5}{*}{\textbf{Method}} & \multicolumn{4}{c}{\textbf{DialSumm Evaluated with G-Eval }} & \multicolumn{4}{c}{\textbf{ROSCOE Evaluated with G-Eval}} \\
\cmidrule(lr){2-5} \cmidrule(lr){6-9}
 & \textbf{Consistency} & \textbf{Coherence} & \textbf{Fluency} & \textbf{Relevance} & \textbf{CosmosQA} & \textbf{DROP} & \textbf{e-SNLI} & \textbf{GSM8K} \\
\midrule
\multicolumn{9}{c}{\textbf{GPT-4o mini}} \\[2pt]
\midrule
CQR       & 2.40 / 94.09\% & 1.76 / 94.50\% & 1.07 / 95.01\% & 2.05 / 94.33\% & 3.60 / 96.60\% & 3.77 / 96.86\% & 3.42 / 97.50\% & 3.57 / 95.10\% \\
Asym CQR      & 2.43 / 94.34\% & 1.86 / 95.87\% & 1.18 / 95.97\% & 2.08 / 94.68\% & 3.95 / 99.69\% & 3.89 / 98.54\% & 2.99 / 97.54\% & 3.94 / 99.27\% \\
CHR           & \underline{1.54 / 86.65\%} & \textbf{1.47 / 90.28\%} & \underline{0.96 / 89.86\%} & \underline{1.40 / 87.49\%} & \graytext{2.48 / 80.31\%} & \graytext{1.82 / 78.06\%} & \graytext{1.30 / 72.41\%} & \graytext{2.25 / 78.90\%} \\
LVD           & 1.90 / 93.81\% & 1.75 / 94.43\% & 1.21 / 93.81\% & 1.80 / 93.77\% & 3.20 / 91.70\% & \underline{2.33 / 86.63\%} & 3.01 / 96.89\% & \underline{3.11 / 89.53\%} \\
Boosted CQR   & 1.85 / 93.33\% & 1.60 / 93.95\% & 1.00 / 93.32\% & \textbf{1.66 / 92.66\%} & 3.16 / 93.40\% & 2.60 / 90.51\% & \underline{2.16 / 89.39\%} & \textbf{3.22 / 90.50\%} \\
Boosted LCP   & \textbf{1.83 / 92.94\%} & 1.60 / 93.01\% & \textbf{0.96 / 93.50\%} & 1.76 / 91.85\% & 3.39 / 94.46\% & 2.97 / 91.71\% & \textbf{1.96 / 92.32\%} & 3.32 / 92.87\% \\
R2CCP         & 1.84 / 93.32\% & 1.63 / 93.38\% & 1.15 / 93.28\% & 1.72 / 93.65\% & \textbf{3.06 / 90.31\%} & \textbf{2.52 / 90.48\%} & 2.16 / 92.35\% & \underline{2.42 / 86.77\%} \\
OrdinalAPS    & 2.24 / 90.39\% & 2.03 / 35.69\% & \graytext{1.87} / 60.61\% & 2.07 / 79.40\% & \graytext{1.79 / 70.61\%} & \graytext{1.44 / 78.57\%} & \graytext{1.75 / 70.13\%} & \graytext{1.36 / 75.03\%} \\
OrdinalRC     & 2.33 / 91.49\% & \graytext{3.17 / 39.46\%} & \graytext{2.01 / 64.66\%} & \graytext{2.21 / 83.29\%} & \graytext{1.94 / 73.16\%} & \graytext{1.52 / 80.73\%} & \graytext{1.84 / 72.32\%} & \graytext{1.44 / 75.47\%} \\
\midrule
\multicolumn{9}{c}{\textbf{DeepSeek-R1-Distill-Qwen-32B}} \\[2pt]
\midrule
CQR       & 2.49 / 93.67\% & 1.51 / 94.80\% & 1.11 / 94.58\% & 2.31 / 93.54\% & 3.61 / 96.43\% & 3.83 / 96.95\% & 3.32 / 97.98\% & 3.54 / 95.80\% \\
Asym CQR      & 2.51 / 93.90\% & 1.58 / 95.41\% & 1.22 / 95.61\% & 2.43 / 94.35\% & 3.89 / 98.95\% & 3.88 / 97.87\% & 2.94 / 97.28\% & 3.90 / 99.37\% \\
CHR           & \textbf{1.76 / 90.78\%} & \textbf{1.25 / 93.09\%} & 1.03 / 92.48\% & \textbf{1.53 / 90.65\%} & \underline{2.66 / 85.95\%} & \underline{2.21 / 87.94\%} & 1.83 / 90.57\% & \underline{2.75 / 89.17\%} \\
LVD           & 2.04 / 93.87\% & 1.41 / 95.07\% & 1.23 / 94.74\% & 1.87 / 93.86\% & 3.34 / 94.69\% & 2.82 / 93.40\% & 2.87 / 98.55\% & 3.42 / 95.47\% \\
Boosted CQR   & {1.89 / 92.91\%} & {1.31 / 94.36\%} & {1.08 / 94.16\%} & {1.71 / 92.95\%} & 3.44 / 94.69\% & 2.88 / 94.41\% & 2.27 / 95.44\% & 3.29 / 93.60\% \\
Boosted LCP   & 1.87 / 91.83\% & 1.32 / 93.31\% & \textbf{0.98 / 93.27\%} & 1.82 / 91.52\% & 3.49 / 94.76\% & 3.03 / 91.27\% & 2.01 / 92.59\% & 3.31 / 91.70\% \\
R2CCP         & 1.85 / 92.87\% & 1.31 / 93.84\% & 1.19 / 93.79\% & 1.70 / 93.23\% & 3.04 / 91.29\% & \underline{2.40 / 89.84\%} & \textbf{1.90 / 90.79\%} & \underline{2.49 / 88.87\%} \\
OrdinalAPS    & 2.05 / 90.05\% & 3.17 / 90.40\% & 3.43 / 90.25\% & \underline{2.17} / \underline{89.90\%} & \textbf{2.90 / 90.99\%} & \textbf{2.27 / 91.24\%} & {3.20 / 91.93\%} & {2.98 / 91.93\%} \\
OrdinalRC     & 2.05 / 90.04\% & {3.17 / 90.30\%} & \graytext{3.42 / 89.93\%} & \underline{2.17 / 89.80\%} & \underline{2.79 / 89.59\%} & 2.22 / 90.73\% & 3.15 / 90.48\% & \textbf{2.86 / 90.83\%} \\
 \midrule
\multicolumn{9}{c}{\textbf{Qwen2.5-72B-Instruct}} \\[2pt]
\midrule
CQR       & 2.37 / 94.62\% & 1.51 / 94.42\% & 1.06 / 94.40\% & 2.05 / 94.91\% & 3.63 / 97.14\% & 3.78 / 97.14\% & 3.38 / 98.42\% & 3.58 / 95.73\% \\
Asym CQR      & 2.41 / 95.10\% & 1.61 / 95.21\% & 1.14 / 95.08\% & 2.04 / 95.00\% & 3.93 / 99.42\% & 3.92 / 99.24\% & 2.95 / 97.50\% & 3.89 / 99.07\% \\
CHR           & \underline{1.48 / 87.99\%} & \underline{1.25 / 89.46\%} & {0.97 / 91.59\%} & \underline{1.34 / 88.06\%} & \graytext{2.62 / 83.74\%} & \graytext{1.77 / 82.92\%} & \underline{1.43 / 85.31\%} & \graytext{1.78 / 82.10\%} \\
LVD           & 1.84 / 94.48\% & 1.48 / 94.77\% & 1.20 / 95.10\% & 1.73 / 94.04\% & 3.36 / 94.90\% & \textbf{2.41 / 92.44\%} & 2.65 / 98.25\% & 2.83 / 92.47\% \\
Boosted CQR   & 1.70 / 92.85\% & \textbf{1.35 / 93.90\%} & 1.05 / 93.82\% & 1.52 / 92.95\% & 3.40 / 94.56\% & 2.57 / 93.30\% & 1.92 / 94.47\% & 3.22 / 92.03\% \\
Boosted LCP   & 1.76 / 92.50\% & \textbf{1.35 / 93.38\%} & \textbf{0.90 / 92.39\%} & 1.62 / 92.52\% & 3.45 / 95.24\% & 2.85 / 90.95\% & \textbf{1.91 / 92.02\%} & 3.38 / 92.73\% \\
R2CCP         & \textbf{1.73 / 93.55\%} & 1.41 / 93.72\% & 1.15 / 93.83\% & 1.60 / 93.17\% & 3.05 / 90.71\% & 3.08 / 95.65\% & \underline{1.59 / 89.67\%} & \underline{2.39 / 89.60\%} \\
OrdinalAPS    & \underline{2.57} / \underline{89.84\%} & \graytext{2.88} / \graytext{63.10\%} & \graytext{2.95} / \graytext{75.48\%} & 2.83 / 90.10\% & \underline{2.78 / 89.42\%} & \textbf{2.02 / 90.95\%} & {2.79 / 93.25\%} & \textbf{2.46 / 90.30\%} \\
OrdinalRC     & \underline{2.57 / 89.85\%} & \graytext{3.01 / 68.81\%} & \graytext{3.09 / 78.40\%} & \underline{2.81 / 89.81\%} & \textbf{2.85 / 90.95\%} & \underline{1.93 / 89.62\%} & {2.71 / 91.40\%} & {2.60 / 91.63\%} \\

\bottomrule
\end{tabular}%
}
\label{tab:dialsumm_discrete}
\end{table*}

\subsection{Partial Boundary Adjustment is Effective to Mitigate Miscoverage}~\label{appendix:0.1_adjustment}
Owing to the heteroscedastic, and correlated nature of LLM‐generated judgments and to the limited calibration set size, prediction intervals before boundary adjustment often fail to achieve 90\% coverage. We propose boundary adjustment as a remedy to overcome miscoverage caused by the ordinal, discrete nature of ratings. A standard boundary adjustment could mitigate miscoverage, but it might also introduce bias or fail to satisfy potential user preference for continuous outputs. Here, we explore partial boundary adjustment that adjust the endpoints of a prediction interval by a pre-defined threshold $\lambda$. More specifically, only those interval endpoints within absolute distance $\lambda$ to an integer will be rounded (e.g., $[3.2, 4.9]$ to $[3.2, 5.0]$ for $\lambda = 0.1$). 

Theoretically, Theorem~\ref{theorem:adjustment_guarantee} suggests that such an outward adjustment shifts the quantile level to include more potential labels, thus improving coverage. Empirically, we conduct experiments with vary $\lambda$ and show the results in Tables~\ref{tab:BoundaryComparison_summarization} and~\ref{tab:BoundaryComparison_reasoning}. Our results suggest that larger $\lambda$ often yield greater coverage gains without much increase or even with decrease on average interval width.

\begin{table*}[ht]
\centering
\caption{R2CCP interval width and coverage for summarization tasks under partial boundary adjustment with $\lambda = 0.5,~0.1,~0$ (no boundary adjustment). Width$\pm$std / coverage\%$\pm$std are computed based on 30 different runs.}
\resizebox{\textwidth}{!}{%
\begin{tabular}{ll|ccc}
\toprule
\textbf{Dataset}   & \textbf{Dimension} & \textbf{0.167 (Full Adjustment)}                          & \textbf{0.1}                          & \textbf{0}                            \\
\toprule
\multicolumn{5}{c}{\textbf{GPT-4o mini}}\\[2pt]
\midrule
  \multirow{4}{*}{SummEval}
   & Consistency & $0.6753\pm0.2026$ / $92.15\%\pm2.25\%$ & $0.6800\pm0.1951$ / $91.68\%\pm2.33\%$ & $0.6858\pm0.1859$ / $90.88\%\pm2.49\%$ \\
                  & Coherence   & $2.6186\pm0.1522$ / $92.81\%\pm2.37\%$ & $2.6201\pm0.1497$ / $91.54\%\pm2.69\%$ & $2.6243\pm0.1466$ / $89.63\%\pm3.12\%$ \\
                  & Fluency     & $0.9116\pm0.1673$ / $90.99\%\pm2.06\%$ & $0.9166\pm0.1657$ / $90.49\%\pm2.29\%$ & $0.9213\pm0.1641$ / $89.36\%\pm2.71\%$ \\
                  & Relevance   & $1.9688\pm0.1288$ / $93.38\%\pm1.96\%$ & $1.9693\pm0.1244$ / $91.90\%\pm2.19\%$ & $1.9705\pm0.1215$ / $89.70\%\pm2.50\%$ \\
\midrule
  \multirow{4}{*}{DialSumm}
   & Consistency & $1.8443\pm0.1299$ / $93.32\%\pm1.85\%$ & $1.8425\pm0.1298$ / $92.03\%\pm1.96\%$ & $1.8436\pm0.1287$ / $90.13\%\pm2.39\%$ \\
                  & Coherence   & $1.6264\pm0.1363$ / $93.38\%\pm2.68\%$ & $1.6274\pm0.1354$ / $92.10\%\pm3.08\%$ & $1.6256\pm0.1337$ / $90.15\%\pm3.38\%$ \\
                  & Fluency     & $1.1504\pm0.1187$ / $93.28\%\pm2.03\%$ & $1.1484\pm0.1237$ / $91.87\%\pm2.56\%$ & $1.1357\pm0.1226$ / $89.64\%\pm2.92\%$ \\
                  & Relevance   & $1.7161\pm0.1398$ / $93.65\%\pm2.06\%$ & $1.7178\pm0.1391$ / $92.15\%\pm2.32\%$ & $1.7209\pm0.1395$ / $90.11\%\pm2.75\%$ \\
\toprule
\multicolumn{5}{c}{\textbf{DeepSeek-R1-Distill-Qwen-32B}} \\[2pt]
\midrule
  \multirow{4}{*}{SummEval}
   & Consistency & $0.6804\pm0.1521$ / $91.57\%\pm2.17\%$ & $0.6876\pm0.1437$ / $91.02\%\pm2.11\%$ & $0.6941\pm0.1343$ / $90.44\%\pm2.09\%$ \\
                  & Coherence   & $2.2972\pm0.1161$ / $93.22\%\pm1.65\%$ & $2.2994\pm0.1169$ / $91.91\%\pm1.88\%$ & $2.3042\pm0.1172$ / $90.12\%\pm2.13\%$ \\
                  & Fluency     & $0.8886\pm0.1605$ / $91.80\%\pm1.82\%$ & $0.8907\pm0.1561$ / $91.09\%\pm2.00\%$ & $0.8926\pm0.1512$ / $90.09\%\pm2.08\%$ \\
                  & Relevance   & $1.9935\pm0.1557$ / $92.96\%\pm2.09\%$ & $1.9951\pm0.1514$ / $91.72\%\pm2.35\%$ & $1.9984\pm0.1482$ / $89.84\%\pm2.90\%$ \\
\midrule
  \multirow{4}{*}{DialSumm}
   & Consistency & $1.8534\pm0.1426$ / $92.87\%\pm2.01\%$ & $1.8574\pm0.1397$ / $91.43\%\pm2.10\%$ & $1.8601\pm0.1371$ / $89.22\%\pm2.62\%$ \\
                  & Coherence   & $1.3113\pm0.1082$ / $93.84\%\pm1.85\%$ & $1.3126\pm0.1077$ / $92.28\%\pm2.14\%$ & $1.3138\pm0.1068$ / $89.92\%\pm2.76\%$ \\
                  & Fluency     & $1.1903\pm0.1368$ / $93.79\%\pm1.78\%$ & $1.1915\pm0.1349$ / $92.58\%\pm2.03\%$ & $1.1859\pm0.1348$ / $90.57\%\pm2.35\%$ \\
                  & Relevance   & $1.6952\pm0.1660$ / $93.23\%\pm1.89\%$ & $1.6982\pm0.1639$ / $91.70\%\pm2.16\%$ & $1.7043\pm0.1601$ / $89.39\%\pm2.57\%$ \\
\toprule
\multicolumn{5}{c}{\textbf{Qwen2.5-72B-Instruct}} \\[2pt]
\midrule
  \multirow{4}{*}{SummEval}
   & Consistency & $0.5876\pm0.1520$ / $91.83\%\pm1.92\%$ & $0.5973\pm0.1447$ / $91.47\%\pm1.88\%$ & $0.6122\pm0.1341$ / $90.73\%\pm2.02\%$ \\
                  & Coherence   & $2.4308\pm0.1457$ / $92.78\%\pm2.07\%$ & $2.4331\pm0.1444$ / $91.53\%\pm2.19\%$ & $2.4367\pm0.1426$ / $89.54\%\pm2.48\%$ \\
                  & Fluency     & $0.9494\pm0.1180$ / $92.12\%\pm1.46\%$ & $0.9500\pm0.1216$ / $91.38\%\pm1.76\%$ & $0.9527\pm0.1218$ / $90.17\%\pm1.92\%$ \\
                  & Relevance   & $1.9765\pm0.1257$ / $93.72\%\pm1.78\%$ & $1.9776\pm0.1253$ / $92.50\%\pm2.06\%$ & $1.9789\pm0.1237$ / $90.45\%\pm2.49\%$ \\
\midrule
  \multirow{4}{*}{DialSumm} &
 Consistency & $1.7319\pm0.1106$ / $93.55\%\pm1.59\%$ & $1.7350\pm0.1083$ / $92.16\%\pm1.91\%$ & $1.7368\pm0.1050$ / $89.97\%\pm2.18\%$ \\
                 & Coherence   & $1.4060\pm0.1115$ / $93.72\%\pm1.97\%$ & $1.4079\pm0.1086$ / $92.07\%\pm2.36\%$ & $1.4094\pm0.1076$ / $89.67\%\pm2.81\%$ \\
                 & Fluency     & $1.1518\pm0.1265$ / $93.83\%\pm2.29\%$ & $1.1475\pm0.1345$ / $92.37\%\pm2.74\%$ & $1.1376\pm0.1398$ / $89.70\%\pm3.42\%$ \\
                 & Relevance   & $1.5966\pm0.1742$ / $93.17\%\pm2.02\%$ & $1.6015\pm0.1714$ / $91.82\%\pm2.43\%$ & $1.6071\pm0.1682$ / $89.80\%\pm2.85\%$ \\
\bottomrule
\end{tabular}%
}
\label{tab:BoundaryComparison_summarization}
\end{table*}

\begin{table*}[ht]
\centering
\caption{R2CCP interval width and coverage for reasoning tasks under partial boundary adjustment with $\lambda = 0.5,~0.1,~0$ (no boundary adjustment). Width$\pm$std / coverage\%$\pm$std are computed based on 30 different runs.}
\resizebox{\textwidth}{!}{%
\begin{tabular}{ll|ccc}
\toprule
\textbf{Judge} & \textbf{Dataset} & \textbf{0.5} & \textbf{0.1} & \textbf{0} \\
\toprule
\multicolumn{5}{c}{\textbf{GPT-4o mini}}\\[2pt]
\midrule
 \multirow{4}{*}{G-Eval}
   & Cosmos & $3.0612\pm0.5594$ / $90.31\%\pm7.07\%$ & $3.0847\pm0.5371$ / $87.31\%\pm8.17\%$ & $3.0864\pm0.5335$ / $86.77\%\pm7.98\%$ \\
  & DROP   & $2.5230\pm0.4804$ / $90.48\%\pm5.53\%$ & $2.5410\pm0.4402$ / $87.30\%\pm6.03\%$ & $2.5431\pm0.4363$ / $86.70\%\pm5.87\%$ \\
  & e-SNLI & $2.1562\pm0.4732$ / $92.35\%\pm6.88\%$ & $2.1932\pm0.4282$ / $88.30\%\pm7.15\%$ & $2.1953\pm0.4264$ / $88.01\%\pm7.17\%$ \\
  & GSM8K  & $2.4205\pm0.7782$ / $86.77\%\pm7.63\%$ & $2.4283\pm0.7639$ / $85.10\%\pm7.50\%$ & $2.4298\pm0.7626$ / $84.67\%\pm8.00\%$ \\
\midrule
  \multirow{4}{*}{SocREval}
   & Cosmos & $2.9294\pm0.4597$ / $89.46\%\pm7.01\%$ & $2.9586\pm0.4396$ / $86.73\%\pm7.80\%$ & $2.9618\pm0.4350$ / $85.85\%\pm7.79\%$ \\
  & DROP   & $2.4125\pm0.8208$ / $89.21\%\pm9.21\%$ & $2.4271\pm0.7631$ / $85.40\%\pm10.04\%$ & $2.4300\pm0.7600$ / $84.73\%\pm9.97\%$ \\
  & e-SNLI & $1.7076\pm0.5804$ / $90.11\%\pm8.41\%$ & $1.7467\pm0.4878$ / $84.99\%\pm8.23\%$ & $1.7480\pm0.4842$ / $84.02\%\pm8.62\%$ \\
  & GSM8K  & $2.0943\pm1.1782$ / $86.93\%\pm8.15\%$ & $2.1452\pm1.0893$ / $85.70\%\pm7.95\%$ & $2.1452\pm1.0866$ / $85.07\%\pm7.87\%$ \\
\toprule
\multicolumn{5}{c}{\textbf{DeepSeek-R1-Distill-Qwen-32B}} \\[2pt]
\midrule
  \multirow{4}{*}{G-Eval}
   & Cosmos & $3.0357\pm0.4873$ / $91.29\%\pm5.71\%$ & $3.0474\pm0.4629$ / $87.35\%\pm6.04\%$ & $3.0489\pm0.4601$ / $86.84\%\pm5.87\%$ \\
  & DROP   & $2.4000\pm0.6406$ / $89.84\%\pm7.80\%$ & $2.4360\pm0.5746$ / $86.48\%\pm7.94\%$ & $2.4385\pm0.5714$ / $85.87\%\pm8.05\%$ \\
  & e-SNLI & $1.8952\pm0.4948$ / $90.79\%\pm7.39\%$ & $1.9532\pm0.4343$ / $86.06\%\pm6.95\%$ & $1.9585\pm0.4315$ / $85.43\%\pm7.02\%$ \\
  & GSM8K  & $2.4865\pm0.8454$ / $88.87\%\pm9.31\%$ & $2.5067\pm0.8065$ / $87.07\%\pm9.04\%$ & $2.5078\pm0.8053$ / $86.77\%\pm8.89\%$ \\
\midrule
 \multirow{4}{*}{SocREval}
   & Cosmos & $2.9094\pm0.6323$ / $90.58\%\pm7.92\%$ & $2.9365\pm0.5718$ / $87.76\%\pm8.45\%$ & $2.9378\pm0.5689$ / $86.97\%\pm8.37\%$ \\
  & DROP   & $2.2457\pm0.6021$ / $89.97\%\pm8.61\%$ & $2.2906\pm0.5369$ / $86.92\%\pm9.06\%$ & $2.2931\pm0.5342$ / $86.35\%\pm9.08\%$ \\
  & e-SNLI & $1.7965\pm0.5139$ / $92.35\%\pm7.77\%$ & $1.8413\pm0.4481$ / $88.45\%\pm8.00\%$ & $1.8450\pm0.4443$ / $87.87\%\pm7.92\%$ \\
  & GSM8K  & $1.8238\pm1.2189$ / $86.93\%\pm7.45\%$ & $1.8767\pm1.1507$ / $85.67\%\pm7.30\%$ & $1.8796\pm1.1480$ / $85.33\%\pm7.02\%$ \\
\toprule
\multicolumn{5}{c}{\textbf{Qwen2.5-72B-Instruct}} \\[2pt]
\midrule
  \multirow{4}{*}{G-Eval}
   & Cosmos & $3.0529\pm0.5262$ / $90.71\%\pm6.80\%$ & $3.0624\pm0.5089$ / $87.82\%\pm8.11\%$ & $3.0652\pm0.5059$ / $87.11\%\pm8.10\%$ \\
  & DROP   & $3.0765\pm0.9169$ / $95.65\%\pm5.33\%$ & $3.0954\pm0.8907$ / $93.68\%\pm7.50\%$ & $3.0964\pm0.8894$ / $93.40\%\pm7.74\%$ \\
  & e-SNLI & $1.5885\pm0.4282$ / $89.67\%\pm6.42\%$ & $1.6737\pm0.3734$ / $85.28\%\pm6.56\%$ & $1.6792\pm0.3694$ / $84.80\%\pm6.89\%$ \\
  & GSM8K  & $2.3922\pm0.6387$ / $89.60\%\pm4.34\%$ & $2.3778\pm0.6328$ / $87.93\%\pm5.71\%$ & $2.3782\pm0.6323$ / $87.53\%\pm5.95\%$ \\
\midrule
  \multirow{4}{*}{SocREval}
   & Cosmos & $2.8786\pm0.5572$ / $89.29\%\pm7.43\%$ & $2.8999\pm0.5210$ / $86.16\%\pm8.76\%$ & $2.8996\pm0.5176$ / $85.34\%\pm8.46\%$ \\
  & DROP   & $2.3446\pm0.6459$ / $90.00\%\pm7.90\%$ & $2.3832\pm0.5883$ / $86.92\%\pm8.14\%$ & $2.3852\pm0.5854$ / $86.25\%\pm8.22\%$ \\
  & e-SNLI & $1.5461\pm0.6282$ / $90.20\%\pm8.42\%$ & $1.5854\pm0.5467$ / $85.43\%\pm8.44\%$ & $1.5897\pm0.5397$ / $84.50\%\pm8.75\%$ \\
  & GSM8K  & $1.9602\pm1.0970$ / $88.57\%\pm7.01\%$ & $2.0026\pm1.0422$ / $87.07\%\pm7.14\%$ & $2.0015\pm1.0391$ / $86.73\%\pm7.09\%$ \\
\bottomrule
\end{tabular}%
}
\label{tab:BoundaryComparison_reasoning}
\end{table*}

\subsection{Additional Results on Midpoints Effectiveness}
Tables~\ref{tab:midpoints_dialsumm},~\ref{tab:midpoints_geval_roscoe} and~\ref{tab:midpoints_socreval_roscoe} include additional results on evaluating the effectiveness of interval midpoint under different settings, including on DialSumm, and on ROSCOE evaluated with G-Eval and SocREval. From the tables, we observe similar findings as to our findings in Table~\ref{tab:midpoints_summeval} that midpoints are consistently less biased evaluation compared to other baseline methods.

\begin{table*}[ht]
\centering
\caption{Midpoints experiment on DialSumm. 
\textbf{Bold} indicates better performance than baselines, \underline{underlined} denotes comparable performance, and \graytext{gray} indicates worse performance.}
\resizebox{\textwidth}{!}{
\begin{tabular}{l|lrrr|rrrr|rrrr|rrrr}
\toprule
\multirow{2.5}{*}{\textbf{Method}} & \multicolumn{4}{c}{\textbf{Coherence}} & \multicolumn{4}{c}{\textbf{Consistency}} & \multicolumn{4}{c}{\textbf{Fluency}} & \multicolumn{4}{c}{\textbf{Relevance}}    \\
\cmidrule(lr){2-5} \cmidrule(lr){6-9} \cmidrule(lr){10-13} \cmidrule(lr){14-17} 
    & \textbf{MSE} & \textbf{MAE} & $\rho$ & $\tau$ 
    & \textbf{MSE} & \textbf{MAE} & $\rho$ & $\tau$ 
    & \textbf{MSE} & \textbf{MAE} & $\rho$ & $\tau$ 
    & \textbf{MSE} & \textbf{MAE} & $\rho$ & $\tau$ \\
\midrule
\multicolumn{17}{c}{\textbf{GPT-4o mini}}{GPT-4o-mini} \\
\midrule
Raw Score 
    & 3.787 & 1.711 & 0.205 & 0.172 
    & 1.000 & 0.772 & 0.656 & 0.547 
    & 2.111 & 1.171 & 0.400 & 0.344 
    & 1.278 & 0.874 & 0.668 & 0.564 \\
Weighted Avg. 
    & 3.701 & 1.699 & 0.218 & 0.162 
    & 0.825 & 0.704 & 0.702 & 0.546 
    & 1.688 & 1.066 & 0.434 & 0.338 
    & 1.175 & 0.855 & 0.703 & 0.549 \\
Con R2CCP 
    & \textbf{0.344} & \textbf{0.454} & \textbf{0.396} & \textbf{0.300}
    & \textbf{0.391} & \textbf{0.489} & \underline{0.688} & \graytext{0.532}
    & \textbf{0.173} & \textbf{0.309} & \underline{0.433} & \underline{0.340}
    & \textbf{0.338} & \textbf{0.445} & \textbf{0.716} & \underline{0.563} \\
Dis R2CCP 
    & \textbf{0.348} & \textbf{0.453} & \textbf{0.385} & \textbf{0.313}
    & \textbf{0.395} & \textbf{0.489} & \underline{0.684} & \textbf{0.553}
    & \textbf{0.178} & \textbf{0.309} & \underline{0.418} & \textbf{0.364}
    & \textbf{0.342} & \textbf{0.446} & \textbf{0.714} & \textbf{0.584} \\
\midrule
\multicolumn{17}{c}{\textbf{DeepSeek-R1-Distill-Qwen-32B}} \\
\midrule
Raw Score 
    & 2.908 & 1.412 & 0.396 & 0.329 
    & 1.422 & 0.952 & 0.589 & 0.497 
    & 2.454 & 1.383 & 0.414 & 0.356 
    & 1.214 & 0.829 & 0.555 & 0.461 \\
Weighted Avg. 
    & 2.149 & 1.241 & 0.456 & 0.343 
    & 0.652 & 0.614 & 0.642 & 0.491 
    & 2.115 & 1.287 & 0.452 & 0.347 
    & 0.674 & 0.625 & 0.621 & 0.476 \\
Con R2CCP 
    & \textbf{0.211} & \textbf{0.348} & \textbf{0.627} & \textbf{0.488}
    & \textbf{0.451} & \textbf{0.509} & \textbf{0.668} & \textbf{0.512}
    & \textbf{0.185} & \textbf{0.315} & \textbf{0.460} & \underline{0.356}
    & \textbf{0.348} & \textbf{0.450} & \textbf{0.721} & \textbf{0.563} \\
Dis R2CCP 
    & \textbf{0.215} & \textbf{0.347} & \textbf{0.615} & \textbf{0.511}
    & \textbf{0.455} & \textbf{0.508} & \textbf{0.665} & \textbf{0.534}
    & \textbf{0.188} & \textbf{0.314} & \textbf{0.455} & \textbf{0.389}
    & \textbf{0.352} & \textbf{0.450} & \textbf{0.716} & \textbf{0.581} \\
\midrule
\multicolumn{17}{c}{\textbf{Qwen2.5-72B-Instruct}} \\
\midrule
Raw Score 
    & 3.934 & 1.775 & 0.321 & 0.267 
    & 1.344 & 0.897 & 0.704 & 0.599 
    & 2.796 & 1.420 & 0.478 & 0.406 
    & 1.812 & 1.070 & 0.609 & 0.521 \\
Weighted Avg. 
    & 3.693 & 1.746 & 0.358 & 0.266 
    & 1.076 & 0.819 & 0.737 & 0.577 
    & 2.575 & 1.335 & 0.499 & 0.386 
    & 1.552 & 1.014 & 0.660 & 0.516 \\
Con R2CCP 
    & \textbf{0.241} & \textbf{0.381} & \textbf{0.583} & \textbf{0.450}
    & \textbf{0.370} & \textbf{0.467} & \underline{0.737} & \underline{0.577}
    & \textbf{0.169} & \textbf{0.306} & \underline{0.489} & \graytext{0.380}
    & \textbf{0.311} & \textbf{0.424} & \textbf{0.727} & \textbf{0.578} \\
Dis R2CCP 
    & \textbf{0.245} & \textbf{0.380} & \textbf{0.574} & \textbf{0.471}
    & \textbf{0.373} & \textbf{0.467} & \underline{0.734} & \underline{0.594}
    & \textbf{0.174} & \textbf{0.306} & \underline{0.485} & \textbf{0.419}
    & \textbf{0.315} & \textbf{0.424} & \textbf{0.722} & \textbf{0.594} \\
\bottomrule
\end{tabular}
}
\label{tab:midpoints_dialsumm}
\end{table*}

\begin{table*}[ht]
\centering
\caption{Midpoints experiment on ROSCOE evaluated by G-Eval. \textbf{Bold} indicates better performance than baselines, \underline{underlined} denotes comparable performance, and \graytext{gray} indicates worse performance. }
\resizebox{\textwidth}{!}{
\begin{tabular}{l|lrrr|rrrr|rrrr|rrrr}
\toprule
\multirow{2.5}{*}{\textbf{Method}} & \multicolumn{4}{c}{\textbf{CosmosQA}} & \multicolumn{4}{c}{\textbf{DROP}} & \multicolumn{4}{c}{\textbf{e-SNLI}} & \multicolumn{4}{c}{\textbf{GSM8k}} \\
\cmidrule(lr){2-5} \cmidrule(lr){6-9} \cmidrule(lr){10-13} \cmidrule(lr){14-17} 
    & \textbf{MSE} & \textbf{MAE} & $\rho$ & $\tau$ 
    & \textbf{MSE} & \textbf{MAE} & $\rho$ & $\tau$ 
    & \textbf{MSE} & \textbf{MAE} & $\rho$ & $\tau$ 
    & \textbf{MSE} & \textbf{MAE} & $\rho$ & $\tau$ \\
\midrule
\multicolumn{17}{c}{\textbf{GPT-4o mini}} \\
\midrule
Raw Score
    & 1.780 & 1.044 & 0.483 & 0.406 
    & 1.843 & 0.951 & 0.490 & 0.411 
    & 2.719 & 1.210 & 0.340 & 0.288 
    & 2.216 & 0.909 & 0.586 & 0.516 \\
Weighted Avg. 
    & 1.704 & 1.065 & 0.490 & 0.371 
    & 1.651 & 0.894 & 0.516 & 0.391 
    & 2.610 & 1.221 & 0.357 & 0.273 
    & 2.169 & 0.936 & 0.577 & 0.458 \\
Con R2CCP 
    & \graytext{2.035} & \graytext{1.223} & \graytext{0.366} & \graytext{0.282}
    & \textbf{1.509} & \graytext{1.034} & \graytext{0.458} & \graytext{0.353}
    & \textbf{1.045} & \textbf{0.865} & \graytext{0.239} & \graytext{0.189}
    & \graytext{2.307} & \graytext{1.282} & \graytext{0.493} & \graytext{0.396} \\
Dis R2CCP 
    & \graytext{2.044} & \graytext{1.220} & \graytext{0.348} & \graytext{0.293}
    & \textbf{1.526} & \graytext{1.024} & \graytext{0.469} & \underline{0.402}
    & \textbf{1.061} & \textbf{0.854} & \graytext{0.231} & \graytext{0.206}
    & \graytext{2.317} & \graytext{1.277} & \graytext{0.501} & \graytext{0.434} \\
\midrule
\multicolumn{17}{c}{\textbf{DeepSeek-R1-Distill-Qwen-32B}} \\
\midrule
Raw Score 
    & 2.353 & 1.166 & 0.396 & 0.335 
    & 2.156 & 0.977 & 0.478 & 0.419 
    & 3.090 & 1.466 & 0.225 & 0.199 
    & 2.300 & 0.906 & 0.596 & 0.538 \\
Weighted Avg. 
    & 1.805 & 1.157 & 0.462 & 0.348 
    & 1.281 & 0.913 & 0.551 & 0.422 
    & 2.144 & 1.286 & 0.279 & 0.214 
    & 1.907 & 1.045 & 0.602 & 0.476 \\
Con R2CCP 
    & \underline{1.931} & \graytext{1.172} & \underline{0.440} & \underline{0.344}
    & \underline{1.485} & \graytext{1.003} & \underline{0.491} & \graytext{0.380}
    & \textbf{0.904} & \textbf{0.802} & \textbf{0.423} & \textbf{0.334}
    & \underline{2.232} & \graytext{1.283} & \graytext{0.540} & \graytext{0.432} \\
Dis R2CCP 
    & \underline{1.936} & \textbf{0.999} & \underline{0.407} & \underline{0.345}
    & \underline{1.518} & \graytext{0.999} & \underline{0.478} & \graytext{0.406}
    & \textbf{0.916} & \textbf{0.792} & \textbf{0.405} & \textbf{0.355}
    & \underline{2.256} & \graytext{1.281} & \graytext{0.539} & \graytext{0.472} \\
\midrule
\multicolumn{17}{c}{\textbf{Qwen2.5-72B-Instruct}} \\
\midrule
Raw Score 
    & 1.964 & 1.179 & 0.420 & 0.364 
    & 1.797 & 0.928 & 0.498 & 0.421 
    & 1.920 & 1.173 & 0.359 & 0.304 
    & 1.911 & 0.820 & 0.653 & 0.589 \\
Weighted Avg. 
    & 1.840 & 1.166 & 0.484 & 0.367 
    & 1.381 & 0.867 & 0.569 & 0.437 
    & 1.615 & 1.101 & 0.388 & 0.292 
    & 1.767 & 0.857 & 0.662 & 0.529 \\
Con R2CCP 
    & \graytext{1.992} & \graytext{1.207} & \underline{0.429} & \graytext{0.329}
    & \underline{1.789} & \graytext{1.124} & \textbf{0.584} & \textbf{0.453}
    & \textbf{0.796} & \textbf{0.727} & \textbf{0.471} & \textbf{0.372}
    & \graytext{2.021} & \graytext{1.174} & \graytext{0.566} & \graytext{0.460} \\
Dis R2CCP 
    & \graytext{2.004} & \graytext{1.200} & \graytext{0.390} & \graytext{0.327}
    & \graytext{1.801} & \graytext{1.121} & \textbf{0.573} & \textbf{0.487}
    & \textbf{0.816} & \textbf{0.716} & \textbf{0.455} & \textbf{0.400}
    & \graytext{2.044} & \graytext{1.173} & \graytext{0.578} & \graytext{0.511} \\
\bottomrule
\end{tabular}
}
\label{tab:midpoints_geval_roscoe}
\end{table*}

\begin{table*}[ht]
\centering
\caption{Midpoints experiment on ROSCOE evaluated by SocREval. 
\textbf{Bold} indicates better performance than baselines, \underline{underlined} denotes comparable performance, and \graytext{gray} indicates worse performance.}
\resizebox{\textwidth}{!}{%
\begin{tabular}{l|lrrr|rrrr|rrrr|rrrr}
\toprule
\multirow{2.5}{*}{\textbf{Method}} & \multicolumn{4}{c}{\textbf{CosmosQA}} & \multicolumn{4}{c}{\textbf{DROP}} & \multicolumn{4}{c}{\textbf{e-SNLI}} & \multicolumn{4}{c}{\textbf{GSM8k}} \\
\cmidrule(lr){2-5} \cmidrule(lr){6-9} \cmidrule(lr){10-13} \cmidrule(lr){14-17} 
    & \textbf{MSE} & \textbf{MAE} & $\rho$ & $\tau$ 
    & \textbf{MSE} & \textbf{MAE} & $\rho$ & $\tau$ 
    & \textbf{MSE} & \textbf{MAE} & $\rho$ & $\tau$ 
    & \textbf{MSE} & \textbf{MAE} & $\rho$ & $\tau$ \\
\midrule
\multicolumn{17}{c}{\textbf{GPT-4o mini}} \\
\midrule
Raw Score 
    & 1.780 & 1.044 & 0.483 & 0.406 
    & 2.969 & 1.284 & 0.202 & 0.168 
    & 1.096 & 0.841 & 0.551 & 0.496 
    & 4.103 & 1.613 & 0.173 & 0.148 \\
Weighted Avg. 
    & 1.704 & 1.065 & 0.490 & 0.371 
    & 1.408 & 0.905 & 0.563 & 0.429 
    & 1.054 & 0.849 & 0.574 & 0.460 
    & 1.612 & 0.771 & 0.649 & 0.523 \\
Con R2CCP 
    & \graytext{1.904} & \graytext{1.170} & \graytext{0.430} & \graytext{0.330}
    & \underline{1.560} & \graytext{1.017} & \graytext{0.495} & \underline{0.386}
    & \textbf{0.725} & \textbf{0.716} & \graytext{0.509} & \graytext{0.408}
    & \underline{2.061} & \underline{1.154} & \underline{0.569} & \underline{0.470} \\
Dis R2CCP 
    & \graytext{1.917} & \graytext{1.165} & \graytext{0.415} & \graytext{0.348}
    & \underline{1.578} & \underline{1.013} & \underline{0.493} & \underline{0.421}
    & \textbf{0.753} & \textbf{0.711} & \graytext{0.505} & \graytext{0.453}
    & \underline{2.095} & \underline{1.144} & \underline{0.589} & \underline{0.527} \\
\midrule
\multicolumn{17}{c}{\textbf{DeepSeek-R1-Distill-Qwen-32B}} \\
\midrule
Raw Score
    & 2.130 & 1.128 & 0.500 & 0.432 
    & 1.443 & 0.803 & 0.630 & 0.564 
    & 0.693 & 0.629 & 0.581 & 0.531 
    & 1.445 & 0.628 & 0.707 & 0.640 \\
Weighted Avg. 
    & 2.016 & 1.107 & 0.525 & 0.398 
    & 1.446 & 0.825 & 0.639 & 0.503 
    & 0.668 & 0.632 & 0.622 & 0.496 
    & 1.425 & 0.645 & 0.664 & 0.522 \\
Con R2CCP 
    & \textbf{1.853} & \graytext{1.151} & \graytext{0.468} & \graytext{0.362}
    & \textbf{1.264} & \graytext{0.914} & \graytext{0.602} & \graytext{0.476}
    & \graytext{0.717} & \graytext{0.708} & \underline{0.615} & \graytext{0.490}
    & \graytext{1.849} & \graytext{1.048} & \graytext{0.595} & \graytext{0.477} \\
Dis R2CCP 
    & \textbf{1.875} & \graytext{1.146} & \graytext{0.595} & \graytext{0.515}
    & \textbf{1.290} & \graytext{0.907} & \graytext{0.595} & \textbf{0.515}
    & \graytext{0.734} & \graytext{0.695} & \graytext{0.580} & \graytext{0.517}
    & \graytext{1.891} & \graytext{1.045} & \graytext{0.637} & \graytext{0.577} \\
\midrule
\multicolumn{17}{c}{\textbf{Qwen2.5-72B-Instruct}} \\
\midrule
Raw Score
    & 1.737 & 0.975 & 0.533 & 0.444 
    & 1.313 & 0.730 & 0.610 & 0.536 
    & 0.590 & 0.488 & 0.651 & 0.591 
    & 1.387 & 0.653 & 0.730 & 0.663 \\
Weighted Avg. 
    & 1.688 & 0.986 & 0.527 & 0.407 
    & 1.290 & 0.735 & 0.603 & 0.475 
    & 0.558 & 0.499 & 0.665 & 0.540 
    & 1.388 & 0.659 & 0.681 & 0.557 \\
Con R2CCP 
    & \graytext{1.897} & \graytext{1.162} & \graytext{0.446} & \graytext{0.348}
    & \graytext{1.378} & \graytext{0.968} & \graytext{0.556} & \graytext{0.456}
    & \underline{0.609} & \underline{0.656} & \underline{0.597} & \underline{0.482}
    & \graytext{1.823} & \graytext{1.063} & \graytext{0.648} & \graytext{0.556} \\
Dis R2CCP
    & \graytext{1.910} & \graytext{1.156} & \graytext{0.442} & \graytext{0.375}
    & \graytext{1.403} & \graytext{0.965} & \graytext{0.541} & \graytext{0.487}
    & \underline{0.632} & \underline{0.652} & \underline{0.595} & \underline{0.533}
    & \graytext{1.849} & \graytext{1.057} & \underline{0.653} & \underline{0.596} \\
\bottomrule
\end{tabular}
}
\label{tab:midpoints_socreval_roscoe}
\end{table*}

\begin{table*}[ht]
\centering
\caption{Reprompting the LLM with prediction intervals reinforces its original judgments since initial scores already lie within those intervals, the model makes trivial adjustments, revealing that it might be hard for interval alone to correct inherent bias since there is no significant difference in each metric for comparison.}
\resizebox{\textwidth}{!}{%
\begin{tabular}{llcccccc}
\toprule
\textbf{Dataset} & \textbf{Width / Coverage} & \textbf{Method} & \textbf{MSE} & \textbf{MAE} & \textbf{$\rho$} & \textbf{$\tau$} \\
\midrule
\multirow{4}{*}{CosmosQA}   & \multirow{5}{*}{2.60 / 89.80\%} & Initial Raw       & 2.204082 & 1.163265 & 0.480293 & 0.419364 \\
\multirow{4}{*}{(Seed = 1)} &                                    & Reprompt Raw      & 2.193877 & 1.153061 & 0.476310 & 0.417798 \\
                            &                                    & Initial Weighted  & 2.052884 & 1.133203 & 0.508314 & 0.390947 \\
                            &                                    & Reprompt Weighted & 2.111918 & 1.167847 & 0.499106 & 0.377264 \\
                            &                                    & Majority Vote & 2.06124 & 1.142857 & 0.453829 & 0.389735 \\
\addlinespace
\multirow{4}{*}{DROP}       & \multirow{5}{*}{1.67 / 89.52\%} & Initial Raw       & 1.371429 & 0.800000 & 0.603949 & 0.551028 \\
\multirow{4}{*}{(Seed = 18)}&                                    & Reprompt Raw      & 1.380952 & 0.809524 & 0.603821 & 0.550921 \\
                            &                                    & Initial Weighted  & 1.333399 & 0.800079 & 0.612075 & 0.485937 \\
                            &                                    & Reprompt Weighted & 1.345206 & 0.814889 & 0.634605 & 0.503989 \\
                            &                                    & Majority Vote & 1.161904 & 1.077917 & 0.636556 & 0.583316 \\
\addlinespace
\multirow{4}{*}{e-SNLI}     & \multirow{5}{*}{1.26 / 89.47\%} & Initial Raw       & 0.684211 & 0.631579 & 0.561363 & 0.517585 \\
\multirow{4}{*}{(Seed = 9)} &                                    & Reprompt Raw      & 0.657895 & 0.631579 & 0.595460 & 0.548320 \\
                            &                                    & Initial Weighted  & 0.610842 & 0.623154 & 0.639802 & 0.512257 \\
                            &                                    & Reprompt Weighted & 0.605095 & 0.638822 & 0.646223 & 0.517462 \\
                            &                                    & Majority Vote & 0.763158 & 0.873589 & 0.548492 & 0.503363 \\
\addlinespace
\multirow{4}{*}{GSM8K}      & \multirow{5}{*}{1.14 / 92.00\%} & Initial Raw       & 0.860000 & 0.420000 & 0.816251 & 0.747567 \\
\multirow{4}{*}{(Seed = 30)}&                                    & Reprompt Raw      & 0.850000 & 0.410000 & 0.819313 & 0.755605 \\
                            &                                    & Initial Weighted  & 0.840141 & 0.437920 & 0.763763 & 0.599531 \\
                            &                                    & Reprompt Weighted & 0.833816 & 0.463369 & 0.738347 & 0.590121 \\
                            &                                    & Majority Vote & 0.91 & 0.45 & 0.817539 & 0.748547 \\
\bottomrule
\end{tabular}%
}
\label{tab:reprompt_results}
\end{table*}

\subsection{Potential Extension to Multimodal Evaluation}
\begin{table*}[ht]
\centering
\caption{An example of extending our framework to text-image consistency evaluation. This table shows the comparison of correlation and error metrics between model original evaluations and midpoints of intervals.}
\begin{tabular}{lcccccc}
\toprule
Method
    & Pearson & Spearman & Kendall & MSE    & MAE    & RMSE \\
\midrule
Std - Raw Score
    & 0.5311  & 0.5650   & 0.4722  & 0.8978 & 0.7718 & 0.9475 \\
Std - Weighted Avg. 
    & 0.5820  & 0.6523   & 0.4790  & 0.8204 & 0.7414 & 0.9058 \\
Std - Con R2CCP 
    & 0.6250  & 0.6296   & 0.4574  & 0.7396 & 0.6828 & 0.8600 \\
Std - Dis R2CCP 
    & 0.6246  & 0.6286   & 0.4747  & 0.7430 & 0.6816 & 0.8620 \\
\midrule
CoT - Raw Score 
    & 0.4932  & 0.5100   & 0.4266  & 1.1778 & 0.8746 & 1.0853 \\
CoT - Weighted Avg.  
    & 0.5571  & 0.6017   & 0.4385  & 1.1177 & 0.8490 & 1.0572 \\
CoT - Con R2CCP         
    & 0.6053  & 0.6096   & 0.4421  & 0.7703 & 0.7083 & 0.8777 \\
CoT - Dis R2CCP    
    & 0.6036  & 0.6067.  & 0.4631  & 0.7752 & 0.7082 & 0.8805 \\
\bottomrule
\end{tabular}
\label{tab:genaibench}
\end{table*}

To explore the broader application of our framework, we conduct experiments on GenAI-Bench~\citep{li2024genaibenchevaluatingimprovingcompositional}, a benchmark designed for vision-language tasks. GenAI-Bench contains 1600 prompt texts, each paired with six model generated images, and each output is rated by three human experts. We prompt Qwen2.5-VL-32B-Instruct~\citep{bai2025qwen25vltechnicalreport} to assess image-text consistency and apply our framework to construct prediction intervals for evaluations.

Two prompting strategies are considered in this new experiment. The first is a standard prompt (Figure~\ref{fig:Std_prompt}), which simply asks the model to assign a quality score. The second is a chain-of-thought (Figure~\ref{fig:CoT_prompt}), followed a chain-of-thought design that asks the model to reason through a more detailed evaluation process and produce its score. For evaluation, we partition the dataset into two splits (800:800) that serves alternately as calibration and test sets, and employ R2CCP to generate intervals.  

For the evaluations by Std\_prompt, the original average interval width is $2.43$ with a coverage of $0.884$, which increase to $2.45$ and $0.919$ after adjustment. For CoT\_prompt, the original interval average width is $2.45$ with a coverage of $0.897$, improving to $2.47$ and $0.929$ after adjustment. As shown in Table~\ref{tab:genaibench}, the midpoints of the intervals yield scores that achieve higher correlations with human ratings and lower error compared to raw or Weighted Avg., demonstrating the effectiveness of our interval-based framework.

For other potential applications, one might follow \citet{wang2025improvingllmasajudgeinferencejudgment, wei2025rocketevalefficientautomatedllm} to obtain pairwise preference based on our intervals as we mention in Section~\ref{subsec: preliminaries}. For example, there are 1600 prompt texts for six models to generate images. For each image, we can obtain an interval by our framework. One possible preference learning method is to compute a preference score that measures the overlapping of a pair of intervals. Since we have a calibration set in conformal prediction, we can also calculate a calibrated confidence of preference. Finally, one can generate a preference directed acyclic graph with six edges for each prompt text in this dataset. On such a preference graph, edges are equipped with the features of pointwise evaluations including intervals and raw response while the link direction is the pairwise preference (A is better than B) while the weight is the calibrated confidence. One might also infer a listwise preference rank with highest accumulated confidence based on this graph.

\subsection{Prompt Used and Responses from Repromt and Regrade}~\label{appendix:prompt}
We adopt G-Eval~\citep{liu2023g} as the main LLM‐as‐a‐judge framework across all tasks and SocREval~\citep{he2024socrevallargelanguagemodels} specifically for reasoning tasks. We make minimal adjuments to the prompt adjustments for each evaluation. Below we provide three representative prompt examples: evaluating relevance on SummEval (Figure~\ref{fig:cot_summeval}), evaluating ROSCOE under both G-Eval (Figure~\ref{fig:geval_roscoe}) and SocREval (Figure~\ref{fig:socreval_roscoe}). It is worth noting that we apply the SummEval prompt template directly to DialSumm. Even though DialSumm is for dialogue summarization rather than news summarization, the resulting intervals on DialSumm still exhibit promising performance.

\newpage
\begin{figure*}
\centering
\begin{tcolorbox}[
    title=\bfseries Prompt on Relevance of SummEval, fonttitle=\large,
    colback=white,
    colframe=black,
    boxsep=1em,
    left=0pt, right=0pt, top=0pt, bottom=0pt,
    arc=0pt,
    enlarge left by=0mm, enlarge right by=0mm
]
\small

You’ll be handed a summary of a news article. \\ [0.5em]

Your challenge is to rate how well the summary captures the essence of the article. \\ [0.5em]

Make sure to thoroughly read and understand these instructions before diving in. Keep this guide handy as you work through the task, so you can refer back to it if needed. \\ [1em]

\textbf{Evaluation Criteria:} \\ [0.25em]
\textbf{Relevance (1--5):} Does the summary hit the mark by including the most important content from the original article? It should focus on the key details without wandering into irrelevant or repetitive information. If the summary strays or over-explains, it should be rated lower. \\ [1em]

\textbf{How to Evaluate:} \\ [0.25em]
\begin{enumerate}[left=1.2em,nosep]
    \item Read both the source article and the summary attentively.
    \item Compare the two, identifying the critical points of the article.
    \item Judge how well the summary captures these important points and avoids unnecessary details.
    \item Give the summary a relevance score between 1 and 5. \\ 
\end{enumerate}

\textbf{Source Article:} \\
\verb|{{Document}}| \\ [1em]

\textbf{Summary:} \\
\verb|{{Summary}}| \\ [1em]

\textbf{Evaluation Form (ENTER A SCORE BETWEEN 1–5):}\\[0.5em]
\textbf{Relevance:}
\end{tcolorbox} 
\caption{Chain-of-thought prompt for evaluating relevance on SummEval.}
\label{fig:cot_summeval}
\end{figure*}

\newpage
\begin{figure*}
\centering
\begin{tcolorbox}[
    title=\bfseries Prompt on ROSCOE by G-Eval, fonttitle=\large,
    colback=white, colframe=black,
    boxsep=1em, left=0pt, right=0pt, top=0pt, bottom=0pt,
    arc=0pt, enlarge left by=0mm, enlarge right by=0mm
]
\small

You will receive a generated response based on the question. \\ [0.75em]

Your mission is to assess whether the generated response answers the question in a well-justified manner. \\ [0.75em]

Please pay close attention to the instructions and keep this guide handy while completing your review. Feel free to refer back to it as needed. \\ [1em]

\textbf{Evaluation Criterion:} \\ [0.25em]
\textbf{Quality (1--5):} 
1=incomprehensible and wrong, \,
5=clear and correct \\ [1em]

\textbf{Steps for Evaluation:} \\ [0.25em]
\begin{enumerate}[left=1.2em,nosep]
\item Carefully read the Situation (Premise) and the Claim (Hypothesis) to understand their key elements.
\item Evaluate the response to determine if it logically connects the Situation to the Claim and sufficiently explains whether the Claim is supported (or not) based on the details provided.
\item Score the response’s quality on a scale of 1 to 5, with 1 being the lowest and 5 the highest, based on clarity, correctness, and the justification provided. \\
\end{enumerate}

\textbf{Situation (Premise):} \\ [0.25em]
\verb|{{Premise}}| \\ [1em]

\textbf{Claim (Hypothesis):} \\ [0.25em]
\verb|{{Hypothesis}}| \\ [1em]

\textbf{Question:} \\ [0.25em]
\verb|{{Question}}| \\ [1em]

\textbf{Generated response:} \\ [0.25em]
\verb|{{Response}}| \\ [1em]

\textbf{Evaluation Form (Only provide the overall quality score from 1–5 without any text):}\\[0.5em]
\textbf{Quality:}
\end{tcolorbox}
\caption{Chain-of-thought prompt on ROSCOE evaluated with G-Eval.}
\label{fig:geval_roscoe}
\end{figure*}

\newpage
\begin{figure*}
\centering
\begin{tcolorbox}[
    title=\bfseries Prompt on ROSCOE by SocREval, fonttitle=\large,
    colback=white, colframe=black,
    boxsep=1em, left=0pt, right=0pt, top=0pt, bottom=0pt,
    arc=0pt, enlarge left by=0mm, enlarge right by=0mm
]
\small

\textbf{Instruction:} \\ [0.5em]
Does the generated response answer the question in a well-justified manner? Please generate your own response for the question first, then conduct a qualitative analysis on the generated response by taking into account your own response, and finally give me an overall quality score in [1, 2, 3, 4, 5] (1=incomprehensible and wrong, 5=clear and correct). Note that you need to consider both the explanation and the answer in the generated response. \\ [1em]

\textbf{Example:} \\ [0.25em]
\begin{itemize}[left=1em,nosep]
    \item \textbf{Situation (Premise):} Two women are embracing while holding to go packages.
    \item \textbf{Claim (Hypothesis):} Two women are holding packages.
    \item \textbf{Question:} Is the Claim supported by the Situation?
    \item \textbf{Generated response:} The two women are most likely embracing because they are either friends or family. If they were just holding packages, there would be no need for them to embrace. The answer is Yes.
    \item \textbf{Example representation:}
    \begin{itemize}
        \item "own response": "Yes, the Claim is supported by the Situation. The Situation states that two women are \"embracing while holding to go packages,\" which means they are holding packages. Therefore, the Claim is accurate.",
        \item "qualitative analysis": "The generated response starts by making an assumption about why the two women might be embracing, suggesting they could be friends or family. While this could be true, it is not directly relevant to the question asked. The core of the question is about the Claim's accuracy given the Situation. The latter part of the generated response does correctly determine that the answer is \"Yes.\" However, the earlier assumption makes the response longer and potentially more confusing.",
        \item "overall quality": 4 \\
    \end{itemize}
\end{itemize}

\textbf{Actual Input:}
\begin{itemize}[left=1em,nosep]
    \item \textbf{Situation (Premise):} \verb|{{Premise}}|
    \item \textbf{Claim (Hypothesis):} \verb|{{Hypothesis}}|
    \item \textbf{Question:} \verb|{{Question}}|
    \item \textbf{Generated response:} \verb|{{Response}}| \\
\end{itemize}

\textbf{Evaluation Request:} \\ [0.25em]
Please generate your own response for the question first, then conduct a qualitative analysis on the generated response by taking into account your own response, and finally give me the overall quality of the given generated response for the question by taking into account both your own response and the qualitative analysis based on the instruction and the format of the example representation. \\ [1em]

\textbf{Evaluation Form (Only provide the overall quality score from 1–5 without any text):}\\[0.5em]
\textbf{Quality:}
\end{tcolorbox}
\caption{Chain-of-thought prompt on ROSCOE evaluated with SocREval.}
\label{fig:socreval_roscoe}
\end{figure*}

\clearpage

\begin{figure*}[ht]
\centering
\begin{tcolorbox}[
    colback=white,
    colframe=black,
    title={Reprompt on ROSCOE by DeepSeek-R1-Distill-Qwen-32B with SocREval},
    fonttitle=\bfseries,
    sharp corners,
    boxrule=0.5pt,
    width=\textwidth
]
\small

Let me show you our evaluation record. Based on all these information, make decision and give me final score. \\ [1em]
\textbf{Initial Prompt:} \\
\verb|{{Prompt in 1st round}}| \\ [1em]
\textbf{Initial Response:} \\
\verb|{{Response in 1st round}}| \\ [1em]
\textbf{Reprompt and Regrade:} \\

Thank you for your initial evaluation! \\ [0.75em]
To help you arrive at a final score that more closely aligns with human expert judgment, we have constructed a 90\% confidence interval for this task using conformal prediction based on your past scoring records. This interval is provided to help you gauge the uncertainty in your recent assessment, which we hope will enhance your evaluation. \\ [0.75em]
\textbf{Interval Information}: The confidence interval we have provided is \verb|{{Interval}}|. Please keep in mind that there is approximately a 90\% probability that the expert’s score lies within this interval, and a 10\% probability that it lies outside. \\ [1em]
— \textbf{Your Objective}: Acting as a human expert, use the interval information along with the recent evaluation task to decide whether and how to adjust the initial score. \\

— Below are some decision-making suggestions for your reference, but we also encourage you to apply your own independent thinking to align as closely as possible with human expert judgment.\\[1em]
\textbf{Decision-Making Suggestions}: \\ [0.5em]
\begin{enumerate}[left=1.2em,nosep]
    \item \textbf{Key Dimensions to Consider}:
    \begin{itemize}[left=1.5em,nosep]
        \item \textbf{Original Score Confidence}: Your confidence level in the score you just assigned;
        \item \textbf{Interval Position}: Whether the original score falls inside or outside the interval;
        \item \textbf{Interval Width}: Whether the interval is narrow (e.g., \(\leq 2.0\)) or wide (e.g., \(\geq 2.0\));
        \item \textbf{Potential Labels}: What specific label options lie within the interval (e.g., \{3.00, 4.00\} for an example interval [3,4]). \\
    \end{itemize}

    \item \textbf{Advice on Decision, Reasoning, and Explanation}:
    \begin{itemize}[left=1.5em,nosep]
        \item \textbf{High Confidence \& Score Within a Narrow Interval}: If the interval is narrow and your score is validated by the interval boundaries, you may confidently retain your original score, provided you believe your evaluation and explanation are seamless. You may also make minor adjustments within the interval where you think the score is most plausible.
        \item \textbf{High Confidence \& Score Outside a Narrow Interval}: Although there is a small probability (<10\%) that you are correct, we encourage you to question your initial judgment, reconsider the evaluation, and consider adjusting the score to the most probable point within the interval, or retain the original score with a brief justification.
        \item \textbf{Low Confidence \& Any Interval}: Use the interval to guide a careful re-examination of the task. For example, consider why an expert rating might take a certain value within the interval and whether that reasoning is sound. After reflection, if you find a value most reasonable, you may choose that score.
        \item \textbf{Challenging the Interval}: You have the right to firmly believe that the true score cannot possibly fall within the provided interval. However, since we guarantee that the interval covers the expert score 90\% of the time, your challenge likely indicates an error in expert judgment. If your explanation convinces us, this would be a valuable discovery. Generally, though, we prefer to treat the expert judgment as ground truth. \\
    \end{itemize}
\end{enumerate}

Please use the suggestions above to produce a \textbf{new final score} through a step-by-step chain of thought:
\begin{enumerate}[left=1.2em,nosep]
    \item Your confidence level in the original score (high/medium/low) and the reason;
    \item How the interval width, potential labels and the position of the original score influence your judgment;
    \item Your adjustment action (retain/minor adjustment/re-examination/other) and the rationale;
    \item The final score you assign. \\
\end{enumerate}

Finally, please first state your final evaluated score (1–5), followed by your explanation: \\ [0.5em]
\textbf{Final Score:}
\end{tcolorbox}
\caption{Prompt used to reprompt and regrade with prediction interval information. Here we provide a template for reprompting and regrading on ROSCOE by DeepSeek-R1-Distill-Qwen-32B with SocREval.}
\label{fig:reprompt_example}
\end{figure*}

\clearpage
\begin{figure*}[ht]
\centering
\begin{tcolorbox}[
    colback=white,
    colframe=black,
    title={Reprompt Response of LLM Judge: Ground truth is 1 in this example (DROP of ROSCOE).},
    fonttitle=\bfseries,
    sharp corners,
    boxrule=0.5pt,
    width=\textwidth
]
\textbf{Final Score}: 2

\medskip
\textbf{Explanation:}
\begin{enumerate}[left=0pt,label=\arabic*.]
    \item \textbf{Confidence Level}: High confidence in the original score of 2. The generated response incorrectly stated the number of losses, which was a clear error.
    \item \textbf{Interval Consideration}: The confidence interval [1.00, 3.00] suggests that the score is likely within this range. The original score of 2 falls within this interval, aligning with the expert judgment.
    \item \textbf{Adjustment Action}: Retain the original score of 2. The reasoning remains sound as the generated response had a factual error, leading to a lower quality score.
    \item \textbf{Rationale}: The generated response's error in interpreting the team's record significantly impacts its quality, justifying the score of 2.
\end{enumerate}
\end{tcolorbox}
\caption{A response of reprompt and regrade with final score and explanations. Initial raw score is 2, initial weighted score is 1.65, and ground truth is 1. Reprompt and regrade strengthens the confidence of LLM judge, since the weighted score changes from 1.65 to 2.04 due to changes in token logits.}
\label{fig:reprompt_example1}
\end{figure*}

\clearpage
\begin{figure*}[ht]
\centering
\begin{tcolorbox}[
    colback=white,
    colframe=black,
    title={Reprompt Response of LLM Judge: Ground truth is 2 in this example (GSM8K of ROSCOE).},
    fonttitle=\bfseries,
    sharp corners,
    boxrule=0.5pt,
    width=\textwidth
]
\textbf{Final Score}: 2

\medskip
\textbf{Explanation:}
After reviewing the generated response and considering the confidence interval [1.00, 2.00], I remain confident that the original score of 2 is appropriate. The generated response incorrectly calculated the water intake rate and the time taken, leading to an overestimation of the water taken on. The correct approach, as detailed in my own response, results in 16 liters, whereas the generated response arrived at 160 liters. The interval suggests that the score is likely within 1 to 2, and since the response is incorrect but shows some understanding, a score of 2 is justified.
\end{tcolorbox}
\caption{A response of reprompt and regrade with final score and explanations. Initial raw score is 2, initial weighted score is 1.81, and ground truth is 2. Reprompt and regrade strengthens the confidence of LLM judge, since the weighted score changes from 1.81 to 2.02 due to changes in token logits.}
\label{fig:reprompt_example2}
\end{figure*}

\clearpage
\begin{figure*}[ht]
\centering
\begin{tcolorbox}[
    colback=white,
    colframe=black,
    title={Reprompt Response of LLM Judge: Ground truth is 3 in this example (CosmosQA of ROSCOE).},
    fonttitle=\bfseries,
    sharp corners,
    boxrule=0.5pt,
    width=\textwidth
]
\textbf{Final Score}: 4

\medskip
\textbf{Explanation:}
1. Confidence in Original Score: High. The initial evaluation correctly identified that Sam's understanding supports the Claim, though it could have elaborated more on the context.

2. Interval Consideration: The interval [1.00, 5.00] is wide, allowing for a range of scores. The original score of 4 falls comfortably within this interval, suggesting it is plausible.

3. Adjustment Action: Retain. The reasoning aligns well with the provided information, and the score accurately reflects the quality of the generated response.

4. Rationale: The generated response correctly answers the question but could have provided more depth. However, it sufficiently supports the Claim, warranting a score of 4.
\end{tcolorbox}
\caption{A response of reprompt and regrade with final score and explanations. Initial raw score is 4, initial weighted score is 3.82, and ground truth is 3. Reprompt and regrade strengthens the confidence of LLM judge, since the weighted score changes from 3.82 to 3.97 due to changes in token logits.}
\label{fig:reprompt_example3}
\end{figure*}

\clearpage
\begin{figure*}[ht]
\centering
\begin{tcolorbox}[
    colback=white,
    colframe=black,
    title={Reprompt Response of LLM Judge: Ground truth is 5 in this example (Consistency of SummEval).},
    fonttitle=\bfseries,
    sharp corners,
    boxrule=0.5pt,
    width=\textwidth
]
\textbf{Final Score}: 4

\medskip
\textbf{Explanation:}
\begin{enumerate}[left=1.2em,nosep]
    \item \textbf{Confidence in Original Score}: High. The initial evaluation correctly assessed the summary’s accuracy without factual errors.
    \item \textbf{Omitted Details}: The summary fails to mention Vanessa’s profession and the police appeal, which are key elements of the source text.
    \item \textbf{Interval Consideration}: The confidence interval [4.67, 5.00] indicates that a higher score is possible under the uncertainty model.
    \item \textbf{Adjustment Action}: Retain. Despite the interval’s suggestion, the omission of critical details justifies keeping the score at 4.
\end{enumerate}
\end{tcolorbox}
\caption{A response of reprompt and regrade with final score and explanations. The LLM judge acknowledges the value of prediction interval but resist changing score.}
\label{fig:reprompt_example4}
\end{figure*}

\clearpage
\begin{figure*}[ht]
\centering
\begin{tcolorbox}[
    colback=white,
    colframe=black,
    title={Reprompt Response of LLM Judge: Ground truth is 5 in this example (Consistency of SummEval).},
    fonttitle=\bfseries,
    sharp corners,
    boxrule=0.5pt,
    width=\textwidth
]
\textbf{Final Score}: 4

\medskip
\textbf{Explanation:}
\begin{enumerate}[left=1.2em,nosep]
    \item \textbf{Confidence in Original Score}: High. The summary accurately reflects the main points without introducing inaccuracies, though it omits some details.
    \item \textbf{Interval Consideration}: The interval [4.67, 5.00] suggests a higher score is expected. However, the summary's omission of Pereira's contract and other clubs' interest justifies a slightly lower score.
    \item \textbf{Adjustment Action}: Retain the original score of 4. The summary's focus on Orlando City's interest makes the omissions acceptable, maintaining consistency without misleading information.
    \item \textbf{Final Score}: 4. The summary is truthful and consistent, though less comprehensive, warranting a score within the lower end of the interval.
\end{enumerate}
\end{tcolorbox}
\caption{A response of reprompt and regrade with final score and explanations. The LLM judge acknowledges the value of prediction interval but resist changing score.}
\label{fig:reprompt_example5}
\end{figure*}

\clearpage
\begin{figure*}[ht]
\centering
\begin{tcolorbox}[
    colback=white,
    colframe=black,
    title={Reprompt Response of LLM Judge: Ground truth is 5 in this example (Consistency of SummEval).},
    fonttitle=\bfseries,
    sharp corners,
    boxrule=0.5pt,
    width=\textwidth
]
\textbf{Final Score}: 4

\medskip

\textbf{Explanation:} The initial score of 4 is within the provided confidence interval [4.67, 5.00], but it's on the lower end. However, the summary accurately reflects the main point without errors, though it lacks some context. Given the interval's guidance, a score of 4 remains appropriate as it aligns with the evaluation criteria.
\end{tcolorbox}
\caption{A response of reprompt and regrade with final score and explanations after moving interval information to the front to avoid position bias. The LLM judge thinks that ``4 is within the provided confidence interval [4.67, 5.00]''.}
\label{fig:reprompt_example6}
\end{figure*}

\clearpage
\begin{figure*}[ht]
\centering
\begin{tcolorbox}[
    colback=white,
    colframe=black,
    title={Reprompt Response of LLM Judge: Ground truth is 5 in this example (Consistency of SummEval).},
    fonttitle=\bfseries,
    sharp corners,
    boxrule=0.5pt,
    width=\textwidth
]
\textbf{Final Score}: 4

\medskip

\textbf{Explanation:} The initial score of 4 is appropriate as the summary accurately reflects the key points without introducing inaccuracies. The confidence interval [4.33, 5.00] suggests a higher score is possible, but the redundancy and omission of details justify maintaining the score at 4.
\end{tcolorbox}
\caption{A response of reprompt and regrade with final score and explanations after moving interval information to the front to avoid position bias. The LLM judge resist changing the score even though it thinks that ``[4.33, 5.00] suggests a higher score is possible''.}
\label{fig:reprompt_example7}
\end{figure*}

\clearpage
\begin{figure*}[ht]
\centering
\begin{tcolorbox}[
    colback=white,
    colframe=black,
    title={Reprompt Response of LLM Judge: Ground truth is 4.67 in this example (Consistency of SummEval).},
    fonttitle=\bfseries,
    sharp corners,
    boxrule=0.5pt,
    width=\textwidth
]
\textbf{Final Score}: 4.67

\medskip

\textbf{Explanation:} The initial score of 4 was based on the summary's accurate reflection of key facts without introducing errors. However, considering the confidence interval [4.67, 5.00], which suggests a high likelihood that the expert score falls within this range, and recognizing that the summary effectively captures the essential details without significant omissions that affect consistency, it is reasonable to adjust the score upward. The summary's completeness, while not perfect, is sufficiently accurate to warrant a higher score within the interval. Therefore, the final score is 4.67.
\end{tcolorbox}
\caption{A response of reprompt and regrade with final score and explanations when the judge can provide intermediate score (not just integer). The judge is willing to raise initial lower score to the lower bound of the interval.}
\label{fig:reprompt_example8}
\end{figure*}

\clearpage
\begin{figure*}[ht]
\centering
\begin{tcolorbox}[
    colback=white,
    colframe=black,
    title={Reprompt Response of LLM Judge: Ground truth is 5 in this example (Consistency of SummEval).},
    fonttitle=\bfseries,
    sharp corners,
    boxrule=0.5pt,
    width=\textwidth
]
\textbf{Final Score}: 4.33

\medskip

\textbf{Explanation:}

\medskip

\begin{enumerate}
    \item \textbf{Confidence in Original Score}: My confidence in the original score of 4 is medium. While the summary is accurate, I recognize that it omits some key details, which might slightly affect its comprehensiveness.
    \item \textbf{Interval Analysis}: The provided confidence interval is [4.67, 5.00], which is narrow. The original score of 4 falls outside this interval, suggesting that the summary might be more consistent than initially assessed.
    \item \textbf{Adjustment Consideration}: Given the interval's narrowness and the fact that the original score is outside, I reconsidered the evaluation. The summary's accuracy and adherence to facts are strong, and the omissions, while present, do not introduce inaccuracies. Therefore, a slight adjustment within the interval is warranted.
    \item \textbf{Final Adjustment}: I adjusted the score to 4.33, reflecting a balance between the summary's accuracy and the minor omissions, while aligning more closely with the confidence interval.
\end{enumerate}
\end{tcolorbox}
\caption{A response of reprompt and regrade with final score and explanations when the judge can provide intermediate score (not just integer). The judge is willing to raise initial lower score but still falls outside of the prediction interval.}
\label{fig:reprompt_example9}
\end{figure*}

\clearpage
\begin{figure*}[ht]
\centering
\begin{tcolorbox}[
    colback=white,
    colframe=black,
    title={Reprompt Response of LLM Judge: Ground truth is 5 in this example (Consistency of SummEval).},
    fonttitle=\bfseries,
    sharp corners,
    boxrule=0.5pt,
    width=\textwidth
]
\textbf{Final Score}: 4

\medskip

\textbf{Explanation:}

\medskip

\begin{enumerate}
    \item \textbf{Confidence Level}: High confidence in the original score of 4. The summary accurately captures the main points without introducing inaccuracies, though it omits some details.
    \item \textbf{Interval Consideration}: The confidence interval is [4.67, 5.00], which is narrow. The original score of 4 falls outside this interval, suggesting a potential adjustment.
    \item \textbf{Adjustment Rationale}: Despite the interval suggesting a higher score, the summary's omission of key details justifies a score of 4. The interval may not fully account for the summary's incompleteness, so retaining the original score is appropriate.
    \item \textbf{Final Decision}: Retain the score of 4, as it aligns with the evaluation criteria and accurately reflects the summary's consistency.
\end{enumerate}
\end{tcolorbox}
\caption{A response of reprompt and regrade with final score and explanations when the judge can provide intermediate score (not just integer). The judge is willing to raise initial lower score but resist changing the score.}
\label{fig:reprompt_example10}
\end{figure*}

\newpage

\begin{figure*}[h!]
\centering
\begin{tcolorbox}[
    colback=white,
    colframe=black,
    title={Standard Prompt for GenAI-Bench: Text-to-Image Quality},
    fonttitle=\bfseries,
    sharp corners,
    boxrule=0.5pt,
    width=\textwidth
]
Evaluate the quality between the following text prompt and the corresponding visual generation on a scale of 1 to 5, where 5 is perfect and 1 is poor. 
\end{tcolorbox}
\caption{Standard prompt for GenAI-Bench evaluated by Qwen2.5-VL-32B-Instruct.}
\label{fig:Std_prompt}
\end{figure*}

\newpage
\begin{figure*}[ht]
\centering
\begin{tcolorbox}[
colback=white,
colframe=black,
title={Chain-of-thought Prompt for GenAI-Bench: Text-to-Image Quality},
fonttitle=\bfseries,
sharp corners,
boxrule=0.5pt,
width=\textwidth
]
\textbf{You will be given:} \\
1) a single generated image $I$ \\
2) the corresponding prompt $P$ \{prompt\}  
    
\medskip
\textbf{Your task:} Output \texttt{Quality Score: (1--5)} followed by a concise explanation.  
Use the five-step chain-of-thought below \textbf{silently}, then distill the insight into one short sentence.

\medskip
\textbf{Internal CoT Steps (do not output):}
\begin{enumerate}[left=0pt,label=\arabic*.]
    \item \textbf{Skill Inventory}: Scan P for the typical generation skills it implicitly tests, e.g.
    \begin{itemize}
        \item Counting (how many objects) 
        \item Spatial layout (relative positions) 
        \item Attribute binding (color, material, style) 
        \item Negation \& absence (what must not appear) 
        \item Action or interaction realism 
        \item Scene coherence \& lighting plausibility
    \end{itemize}
    \item \textbf{Global Coherence Check}: Does the entire scene feel like one unified photograph / artwork that matches the prompt’s mood and setting?  
    \item \textbf{Critical Object Pass}: Identify the 1--3 most semantically heavy objects or relations. Are they present, correctly shaped, and believably integrated?  
    \item \textbf{Fine-Grain Glance}: Spot any subtle oddities (extra limbs, miscounts, impossible lighting, texture smears). Even tiny flaws here can reveal skill gaps.
    \item \textbf{Holistic Judgment}: Weigh the combined effect: if the key skills are satisfied and the image looks natural → 5; mild off-notes → 4; clear but non-fatal misses → 3; multiple visible failures → 2; fundamentally wrong → 1.
\end{enumerate}

\medskip
\textbf{Output Format (public):} \\
Quality Score: X \\
Explanation: One short sentence summarizing the overall perception ($\leq$ 25 words).
\end{tcolorbox}
\caption{Chain-of-thought prompt on GenAI-Bench evaluated by Qwen2.5-VL-32B-Instruct.}
\label{fig:CoT_prompt}
\end{figure*}

\end{document}